%% file: main.tex
\documentclass[]{fairmeta}

\usepackage{microtype}
\usepackage{graphicx}
\usepackage{graphicx}
\usepackage{subcaption}
\usepackage{booktabs} 
\usepackage[table]{xcolor}

\usepackage{hyperref}

\usepackage{amsmath}
\usepackage{amssymb}
\usepackage{mathtools}
\usepackage{amsthm}
\usepackage{enumitem}

\usepackage{tikz}
\usetikzlibrary{arrows.meta, positioning, calc}
\usepackage{pgfplots}
\pgfplotsset{compat=1.18}

\definecolor{maroon}{HTML}{F26035}
\definecolor{yellow}{HTML}{FDBC42}
\definecolor{darkblue}{RGB}{31, 90, 153}
\definecolor{forestgreen}{rgb}{0.13, 0.55, 0.13}
\definecolor{brickred}{rgb}{0.8, 0.25, 0.33}

\colorlet{metagray}{metafg!45}
\colorlet{metalight}{metafg!12}

\newcommand{\compr}{compression rate}
\newcommand{\bpp}{bytes per parameter}
\newcommand{\pari}{parity}

\usepackage{tcolorbox}
\tcbuselibrary{skins,breakable}

\newtcolorbox{keyfinding}[1][]{
  colback=metabg,
  colframe=metablue,
  fonttitle=\bfseries,
  title={Key Finding},
  arc=2mm,
  boxrule=0.5pt,
  left=4pt, right=4pt, top=4pt, bottom=4pt,
  #1
}

\pgfplotsset{
  metaAxis/.style={
    width=0.235\textwidth,
    height=0.22\textwidth,
    axis line style={metafg, line width=0.6pt},
    tick style={metafg, line width=0.6pt},
    ticklabel style={metafg, font=\small},
    label style={metafg, font=\small},
    title style={metafg, font=\small},
    grid=both,
    grid style={metalight},
    major grid style={metalight},
    minor tick num=1,
    legend style={draw=none, fill=none, font=\small, text=metafg},
  },
  metaBlueLine/.style={metablue, line width=0.9pt},
  metaGrayLine/.style={metagray, line width=0.7pt},
  metaMin/.style={only marks, mark=triangle*, mark size=2.6pt,
                  draw=metafg, fill=metablue},
  metaOpen/.style={only marks, mark=o, mark size=2.2pt,
                  draw=metablue, fill=white, line width=0.8pt},
}

\theoremstyle{plain}

\theoremstyle{definition}

\theoremstyle{remark}

\usepackage[textsize=tiny]{todonotes}
\usepackage{wrapfig}
\title{Compute Optimal Tokenization}

\author[1]{Tomasz Limisiewicz}
\author[1]{Artidoro Pagnoni}
\author[1]{Srini Iyer}
\author[1]{Mike Lewis}
\author[1]{Sachin Mehta}
\author[2]{Alisa Liu}
\author[2]{Margaret Li}
\author[1]{Gargi Ghosh}
\author[1]{Luke Zettlemoyer}

\affiliation[1]{FAIR at Meta}
\affiliation[2]{University of Washington}

\abstract{
Scaling laws enable the optimal selection of data amount and language model size, yet the impact of the data unit, \emph{the token}, on this relationship remains underexplored.
In this work, we systematically investigate how the information granularity of tokens, controlled by the \compr{} (i.e., average bytes of text per token), affects scaling trends.
We train 988 latent tokenized models (BLT) ranging from 50M to 7B parameters that enable setting the desired \compr{}. This flexibility allows us to study the role of \compr{} well beyond 4.57 bytes per token obtained with a popular BPE tokenizer.
Our experiments reveal that in compute-optimal configurations, model parameter counts scale proportionally to data size measured in \emph{bytes}, not in \emph{tokens} as commonly perceived \citep{kaplan2020scaling, hoffmann2022training}. Furthermore, we discover that the optimal \compr{} differs from the one obtained with BPE and decreases with compute.
These findings generalize to both latent and subword tokenization, as well as to languages other than English, guiding language model developers on tokenization scheme selection for maximal compute efficiency.
}

\date{May 4, 2026}
\correspondence{\email{tomlim@meta.com}}

\metadata[Code]{\url{https://github.com/facebookresearch/compute-optimal-tokenization}}
\metadata[Blogpost]{\url{https://co-tok.github.io}}
\metadata[Accepted]{Transactions on Machine Learning Research}
\begin{document}

\maketitle



\vspace{20pt}
\begin{figure*}[!htb]
    \centering
    \begin{subfigure}{0.32\textwidth}
        \includegraphics[width=\linewidth]{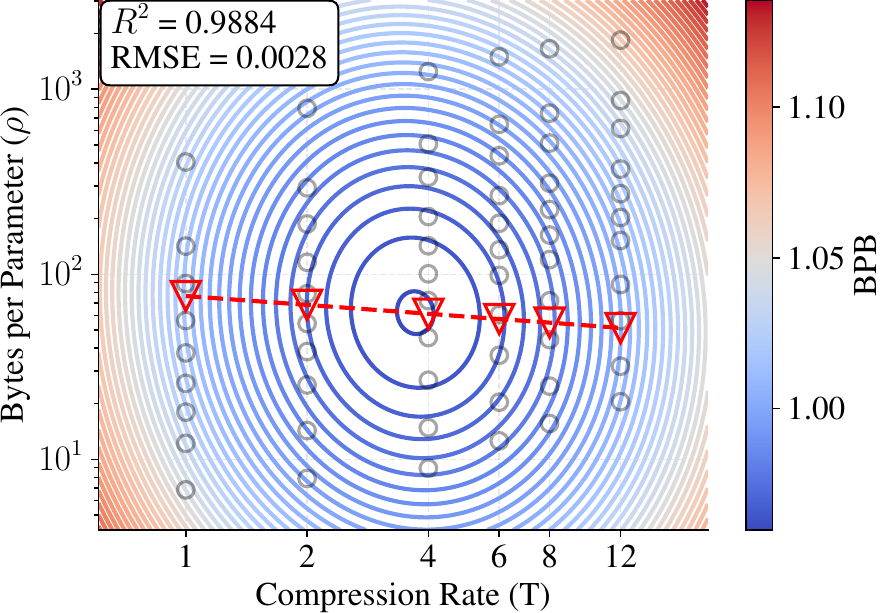}
        \caption{Optimal \bpp{} ratio across compression rates. Fixed training budget $10^{20}$ FLOPs.}
        \label{fig:main-bpp}
    \end{subfigure}
    \hfill
    \begin{subfigure}{0.32\textwidth}
        \includegraphics[width=\linewidth]{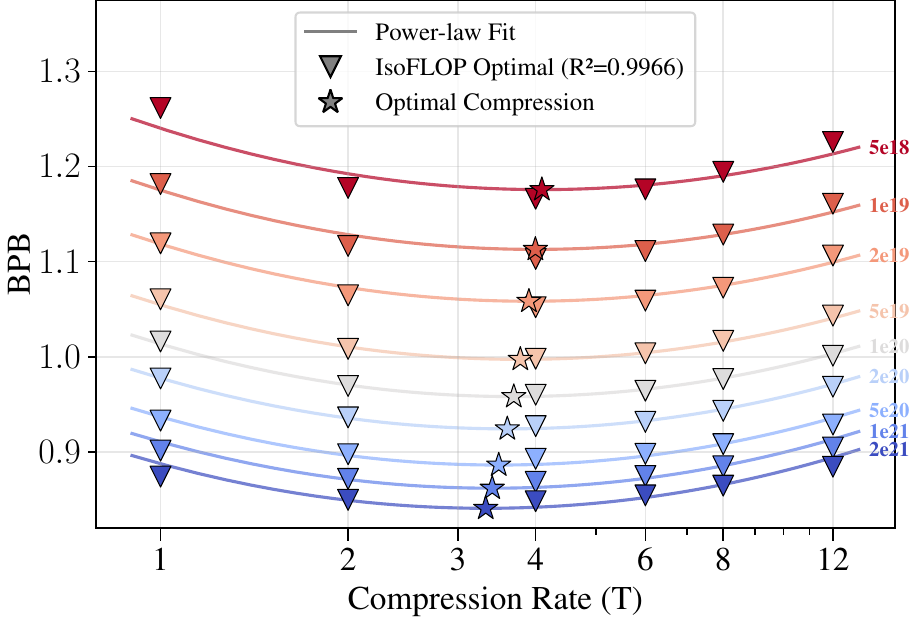}         
        \caption{Optimal \compr{} based on a scaling law fit. Training budget marked by color.}
        \label{fig:main-cr}
    \end{subfigure}
    \hfill
    \begin{subfigure}{0.32\textwidth}
        \includegraphics[width=\linewidth]{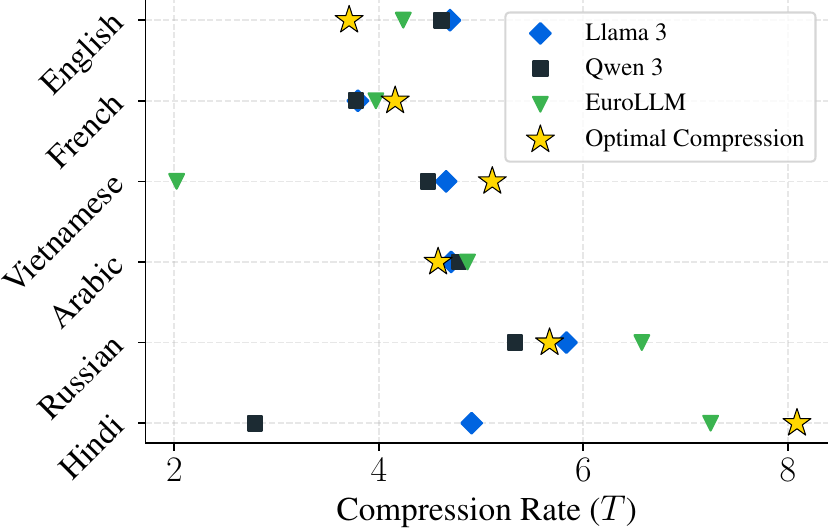}
        \caption{Optimal \compr{} differs from compression of subword tokenizers}
        \label{fig:main-tokenizers}
    \end{subfigure}
    \caption{Key findings of this work: \ref{fig:main-bpp}
    in compute optimal scaling: bytes (not tokens) of data  increase proportionally to parameter count; \ref{fig:main-cr} for each training budget, we find optimal \compr{}, its value decreases with scale;
    \ref{fig:main-tokenizers} the optimal \compr{} varies across languages and differs from compression of popular BPE tokenizers.}
    \label{fig:main_figure_idea}
\end{figure*}

\clearpage
\begingroup
\small
\tableofcontents
\endgroup
\clearpage

\input{sections/01_introduction}
\input{sections/02_methodology}

\input{sections/03_compression_scaling}

\input{sections/04_isotropic}

\input{sections/05_across_domains}
\input{sections/06_related_work}
\input{sections/07_discussion}
\input{sections/08_conclusion}

\clearpage
\newpage
\bibliographystyle{assets/plainnat}
\bibliography{custom}

\clearpage
\newpage
\beginappendix
\input{sections/appendices}

\end{document}

%% file: sections/01_introduction.tex
\section{Introduction}
\label{sec:introduction}


Scaling laws have informed the efficient design of language models, prescribing the optimal balance between model size and training data \citep{kaplan2020scaling, hoffmann2022training}.
Standard approaches estimate the optimal amount of data in tokens for a given compute budget (and model size).
However, expressing data volume in tokens overlooks a critical aspect: the information density that each token represents.
Consequently, scaling findings inherently depend on specific tokenizers and their key property: the \compr{}.

To fill the research gap, we introduce laws that are aware of the \compr{} $T$, defined as the average number of bytes per token in a given dataset.
For that purpose, we need to vary the \compr{} without changing the vocabulary size (and thus the number of parameters).
Therefore, in our experiments we rely on Byte Latent Transformer (BLT, \citealp{pagnoni-etal-2025-byte}), a recent architecture that segments byte-level input in a latent space.
BLT's latent tokenization is a robust tool for this purpose, as it allows us to precisely adjust the \compr{} by setting an average segment size.\footnote{Recent latent tokenized models allow achieving a wide range of \compr{}, similarly to BLT.
However, in other approaches \compr{} cannot be precisely controlled due to reliance on a segment boundary predictor \citep{hwang2025dynamicchunkingendtoendhierarchical,nawrot-etal-2023-efficient} or whitespace supervision \citep{neitemeier2025hierarchical, slagle2024spacebyte,videau2025bytesideaslanguagemodeling}.}
Additionally, compression plays a significant role in subword tokenization.
We can order popular subword methods by their \compr{}: from pure byte or character-level segmentation ($T\approx1$) \citep{xue-etal-2022-byt5,wang2024mambabytetokenfreeselectivestate}, through widely used BPE ($T\approx4.57$) \citep{sennrich-etal-2015-neural}, to SuperBPE ($T\approx6.16$) \citep{liu2025superbpespacetravellanguage}, which achieves high compression by allowing multi-word tokens.\footnote{Estimates of \compr{} are computed for DCLM corpus \citep{li2024datacomp} consisting of ``plain English'' texts.}

In the context of scaling, \compr{} impacts model efficiency in both training and inference.
Increasing compression allows the same data to be represented with fewer tokens, directly reducing the computational cost of processing.
The unlocked savings in FLOPs (unit of computation) can be used to increase training data, model size, or both, without increasing the total computation budget.

To the best of our knowledge, this is the first thorough study of
the effect of \compr{} on the compute efficiency of language models.
We pose the following research questions:

\paragraph{\textbf{[R1]}: How does compression rate impact the compute-optimal ratio between parameters and data?}
This question concerns the unit of data we should use in model scaling.
We investigate whether the compute-optimal ratio is best expressed in tokens or bytes (which are the underlying unit of text encoding).
For example, given the Chinchilla rule of thumb of training on $\approx 20$ tokens per parameter \citep{hoffmann2022training}, does this ratio hold as we increase compression, or should the ratio of bytes to parameters remain constant given a dataset of English texts?

\paragraph{\textbf{[R2]}: Is there an optimal compression rate for specific datasets?}
We investigate whether there exists a \compr{} that yields the lowest loss for a fixed compute budget, assuming the optimal data to parameter ratio.
Furthermore, we examine whether this optimal \compr{} shifts with the compute budget or dataset domain. 

\paragraph{\textbf{[R3]}: Is the impact of compression rate on scaling trends similar for latent and subword tokenized models?}
Does the answer to the previous questions depend on the tokenization method?
We conduct experiments on subword-tokenized models to validate if the scaling trends match those observed for BLT.

\paragraph{\textbf{[R4]}: Is optimal compression rate language specific?}
We extend our experiments to languages other than English to test whether optimal data to parameter ratio and compression rate change depending on language.
We hypothesize that both will grow proportionally to \emph{parity}, defined as the ratio of byte length of parallel sentences expressed in two languages \citep{petrov2023token_unfairness, ahia-etal-2023-languages}.

The structure of the paper is as follows. In Section~\ref{sec:methodology}, we describe background work and  our experimental setting, including details of the datasets, models, and methods for deriving power laws.
In Section~\ref{sec:compression_scaling}, we present experiments scaling BLT across a wide range of \compr{}s to answer \textbf{[R1]} and \textbf{[R2]}.
In Section~\ref{sec:subword_scaling}, we examine subword-tokenized models to compare with the findings from the previous section and address \textbf{[R3]}.
Finally, in Section~\ref{sec:beyond_english}, we extend our scaling experiments to languages other than English to answer \textbf{[R4]}.

%% file: sections/02_methodology.tex
\section{Methodology}
\label{sec:methodology}

\begin{wrapfigure}{r}{0.45\textwidth}
  \begin{minipage}{0.44\textwidth}
  \includegraphics[width=\linewidth]{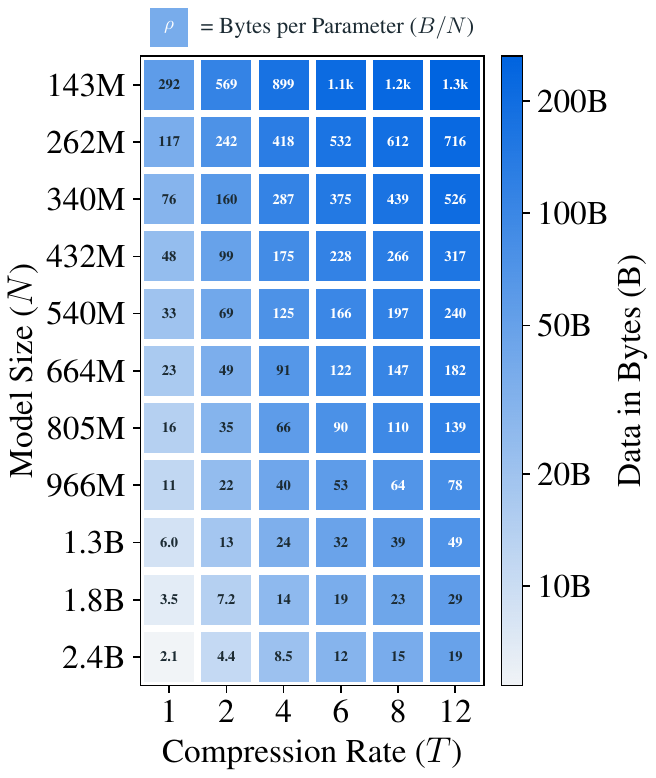}
  \caption{The grid of experiments for the budget of $C=10^{20}$ FLOPs. Value of \compr{} $T$ (x-axis) and model size $N$ (y-axis) determine the amount of training data $B$ (color) and the \bpp{} ratio $\rho$ (values in squares).}
  \label{fig:experimental_grid}
  \end{minipage}
\end{wrapfigure}

In this section, we describe model architectures, training and evaluation procedures.
We define variables used to describe the compute optimal scaling trends and outline methods for fitting power laws to model these trends.

\subsection{Background: Subword Tokenization}

Byte Pair Encoding (BPE) \citep{sennrich-etal-2015-neural} is the dominant subword tokenization
method in modern language models. 
BPE iteratively merges the most frequent pair of adjacent symbols in a corpus, starting from a base alphabet of bytes or characters, until a target vocabulary size $V$
is reached. 
Each merge produces a new token representing a longer byte sequence.
As the core BPE implementation, we use Llama3~\citep{llama3herd2024} tokenizer with vocabulary of $V = 128{,}000$,
yielding \compr{} of approximately $T\approx 4.57$ bytes per token on English texts.
A key property of BPE for our study is that its compression rate is fixed once the vocabulary is
trained. 

To vary the effective compression
within the BPE framework, we employ \textbf{vocabulary masking}: we disable a fraction of the
learned merge rules (masking 75\% or 90\% of the vocabulary), forcing the tokenizer to produce shorter, more frequent tokens. 
The tokenizers differ from base BPE mainly by achieving lower \compr{s} ($T =4.16$ and $T= 3.71$, respectively) while segmentation algorithm and vocabulary construction are the same (except the masking).
We also consider a \textbf{character-level} representation as a special case of subword tokenization, which obtains low \compr{} ($T=1.01$).

At the other extreme, \textbf{SuperBPE}~\citep{liu2025superbpespacetravellanguage} relaxes the standard assumption that tokens are restricted within word boundaries. 
By allowing merges to cross whitespace, SuperBPE produces multi-word tokens that encode text with fewer tokens than a standard BPE vocabulary of equivalent size. 
In our experiments, SuperBPE ($V = 200{,}000$) achieves
$T \approx 6.16$, representing the highest compression rate among subword methods we
consider.

\subsection{Background: Latent Tokenization}

In contrast to subword tokenizers that commit to a fixed segmentation, \textbf{latent tokenization} learns to group input units (bytes or characters) into variable-length segments within the model. We describe this method of tokenization as ``latent'', because latent tokens (or segments or patches) appear only in model's internal representation, while the input and output are byte-level.

Recent works have introduced hierarchical model architecture to implement latent tokenization \citep{nawrot-etal-2023-efficient, hwang2025dynamicchunkingendtoendhierarchical, neitemeier2025hierarchical}.
Such an architecture comprises three modules: (1) an encoder that aggregates byte-level representations into latent tokens; (2) a global module operating on these latent tokens (i.e., Transformer operating on latent tokens); and (3) a decoder that maps latent representations back to the byte level for next-byte prediction.
In contrast to hierarchical architecture, throughout the paper we refer to language models with subword tokenization as ``isotropic'', i.e., models whose modules operate on sequences of the same granularity, unlike in hierarchical models.

\paragraph{Byte Latent Transformer (BLT)}
In our experiments, we use Byte Latent Transformer (BLT) architecture~\citep{pagnoni-etal-2025-byte}.
BLT utilizes entropy spikes to segment byte sequences into latent tokens, allowing us to control the \compr{} by adjusting the entropy threshold.
Crucially for our purposes, BLT segments byte sequences based on the entropy of the next byte: high-entropy positions indicate segment boundaries, allocating more compute where data complexity is greater. 
By adjusting the entropy threshold, we can precisely control the average patch size, and hence the \compr{}, without changing the model's parameter count. 
This makes BLT an ideal testbed for studying the effect \compr{} on scaling behavior across wide range: 
($T \in \{1, 2, 4, 6, 8, 12\}$).
A key deviation from the original BLT implementation is the omission of hash embeddings for byte n-grams.
We omit these because n-grams can span more bytes than the latent tokens themselves, potentially interfering with the target \compr{}.

\subsection{Scaling Variables Definitions}

Scaling laws allow us to predict optimal training regimes of language models.
Given a pre-defined computation budget, our task is to select specific model size in parameters and training data amount to minimize the loss of the model.
Throughout this paper, we use the following notation for the key variables governing compute-optimal
scaling:

\begin{itemize}[leftmargin=*]
    \item \textbf{Compute budget ($C$)} expressed by floating point operations (FLOPs) quantify how many operations are performed during model training.
    \item \textbf{Model size ($N$)} in parameters. Unless stated otherwise, we consider global module parameters, excluding encoder, decoder, and embedding layers.
    \item \textbf{Training data size ($D$)} is the number of tokens used for training a language model. 
    In this paper we prefer to quantify data as number of bytes of text used for training ($B$), because this value is tokenizer independent.
    \item \textbf{Byte-per-parameter ($\rho$)} is defined as $\rho = B / N$. This quantity captures the data-to-model proportion and is crucial for scaling analysis.
    \item \textbf{Compression rate ($T$)}, defined as the average number of bytes per token on reference corpus $T=B/D$.
    \item \textbf{Validation loss ($L$)} measured in bits-per-byte (BPB) to enable fair comparison across different tokenizers.
\end{itemize}

\subsection{Model Architectures}

For all experiments, we train Transformer models \citep{vaswani2017attention} of varying parameter sizes, adhering to Llama 3 architectural choices \citep{llama3herd2024}.
We follow a standard scaling recipe: increasing models' width and depth in a 1:1 ratio, meaning the number of heads equals the number of layers.
The latent dimension size is set to 128 times the number of heads, and the feed-forward network uses 4$\times$ upscaling.

The exact architecture specifications for all models used in our study are presented in Appendix~\ref{sec:model-scaling-details}.

\subsection{Training and Evaluation}

We train models under compute budgets ($C$) expressed in FLOPs, ranging from $5\times10^{18}$ to $2\times10^{21}$ FLOPs. 
If not stated otherwise, we use exact computation of training FLOPs, instead of an approximation.
In total, we train 988 latently and 320 subword tokenized models with sizes from 50M to 6.7B parameters on training data of sizes from 4B to 1.1T bytes.

For each budget $C$, we vary the parameter size ($N$) and compression rate ($T$).
The parameters $N$ and \compr{} $T$ uniquely determine the training data amount in bytes ($B$).
Consequently, for each compute budget, we obtain a grid of models corresponding to the cartesian product of $T$ and $N$.
For each of the configurations, we compute the \bpp{} ratio ($\rho$), as shown in Figure~\ref{fig:experimental_grid}.
For BLT, we test six \compr{} values $T \in \{1,2,4,6,8,12\}$, while for subword models, \compr{} is determined by the tokenizer $T \in \{1.01,3.71,4.16,4.57,6.16\}$.
For all training runs, we fix the batch size at 2 million bytes and the learning rate at $4\times10^{-4}$.
We use the AdamW optimizer \citep{loshchilov2019decoupled} with a warmup-stable-decay learning rate schedule.

Unless stated otherwise, we train on DCLM \citep{li2024datacomp}, a dataset of plain English texts selected to limit data mixing across domains and languages.
Data mixing could cause non-uniform granularity of information and thus confound our analysis.
We evaluate models on the C4 validation split \citep{raffel2020exploring}.

To compare loss across different models with various tokenization methods, we evaluate models using bits-per-byte (BPB), which is loss divided by the number of bytes in the evaluation texts.
In each training and evaluation example, we fix context to contain the same number of 8192 bytes (e.g., with \compr{} $T=4$ we evaluate on 2048 tokens per example, then with compression rate $T=8$ we evaluate on 1024 tokens).

\subsection{Fitting Power Laws}

We fit the parameters for power laws presented in the next section using the BFGS optimizer \citep{liu1989bfgs,zhu1997bfgsb} minimizing sum of squares loss.
To ensure reliability, we initiate optimization from multiple random seeds and compute confidence intervals using a numerical approximation of the Hessian.
Further details on the fitting procedure can be found in Appendix~\ref{sec:power-law-details}.

%% file: sections/03_compression_scaling.tex
\section{Scaling Laws and Data Compression}
\label{sec:compression_scaling}

\begin{figure*}[!tb]
    \centering
    \begin{subfigure}{0.49\textwidth}
        \includegraphics[width=\linewidth]{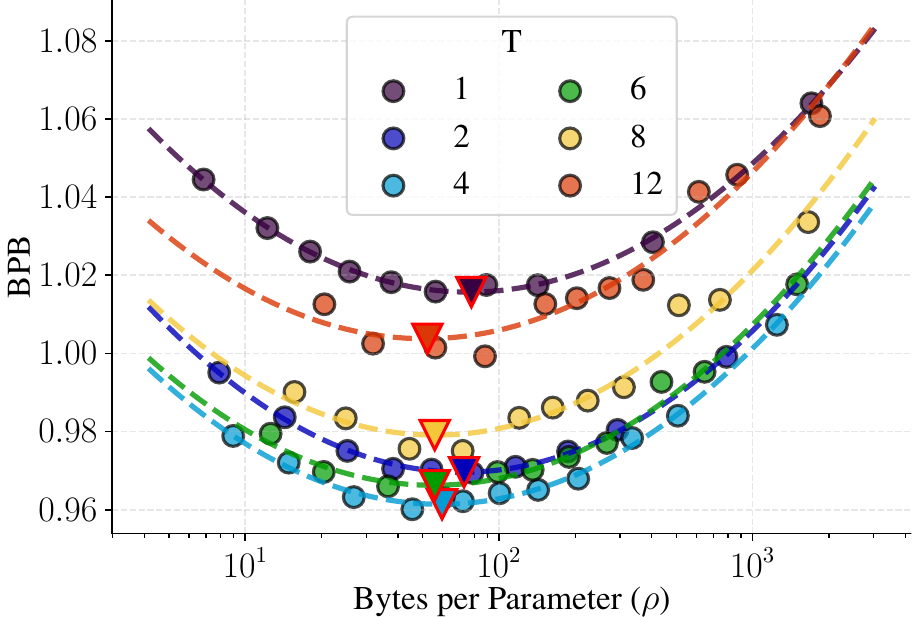}
        \caption{2D IsoFLOP}
    \end{subfigure}
    \hfill
    \begin{subfigure}{0.49\textwidth}
        \includegraphics[width=\linewidth]{figures/blt_entropy_1e20_results_compression_contour_global_parameters.pdf}
        \caption{3D IsoFLOP (heatmap)}
    \end{subfigure}
    \caption{
    Evaluation scores of latent tokenized models on C4 test set with fixed FLOPs budget ($C=10^{20}$), compared against \bpp{} ratio.
    2-dimensional IsoFLOP (parabola) were fitted for each compression rate, while 3-dimensional IsoFLOP jointly for all compression rates (on x-axis). 
    Minima of both fits show that minimal loss is obtained at almost constant value of \bpp{} ratio $\rho\approx60$.
    For IsoFLOPs as function of data, parameters, and for other compute budgets, refer to Appendix~\ref{sec:isoflop-across-budgets} 
    }
    \label{fig:stage_1_blt_1e20}
\end{figure*}

In this section, we present scaling results for BLT models revealing the role of data \compr{}.
We fit scaling laws in two stages, as such an approach shows more faithful approximations \citep{li2025misfitting}.
In the first stage, we estimate the optimal training data size in bytes $B^{\star}$ and model size  $N^{\star}$  as a power law function of compute budget $C$ and compression rate $T$, addressing research question \textbf{R1}.
Subsequently, in the second stage, we model the dynamics of the optimal loss $L^\star$ obtained for the found $B^{\star}$ and $N^{\star}$ configuration.
We examine the effect of \compr{} $T$ on $L^\star$  to answer research question \textbf{R2}.

\subsection{Scaling Law I: Optimal Data and Parameters}

\begin{figure}[!htb]
\centering
  \begin{subfigure}[t]{0.49\textwidth}
    \centering
    \includegraphics[width=\linewidth]{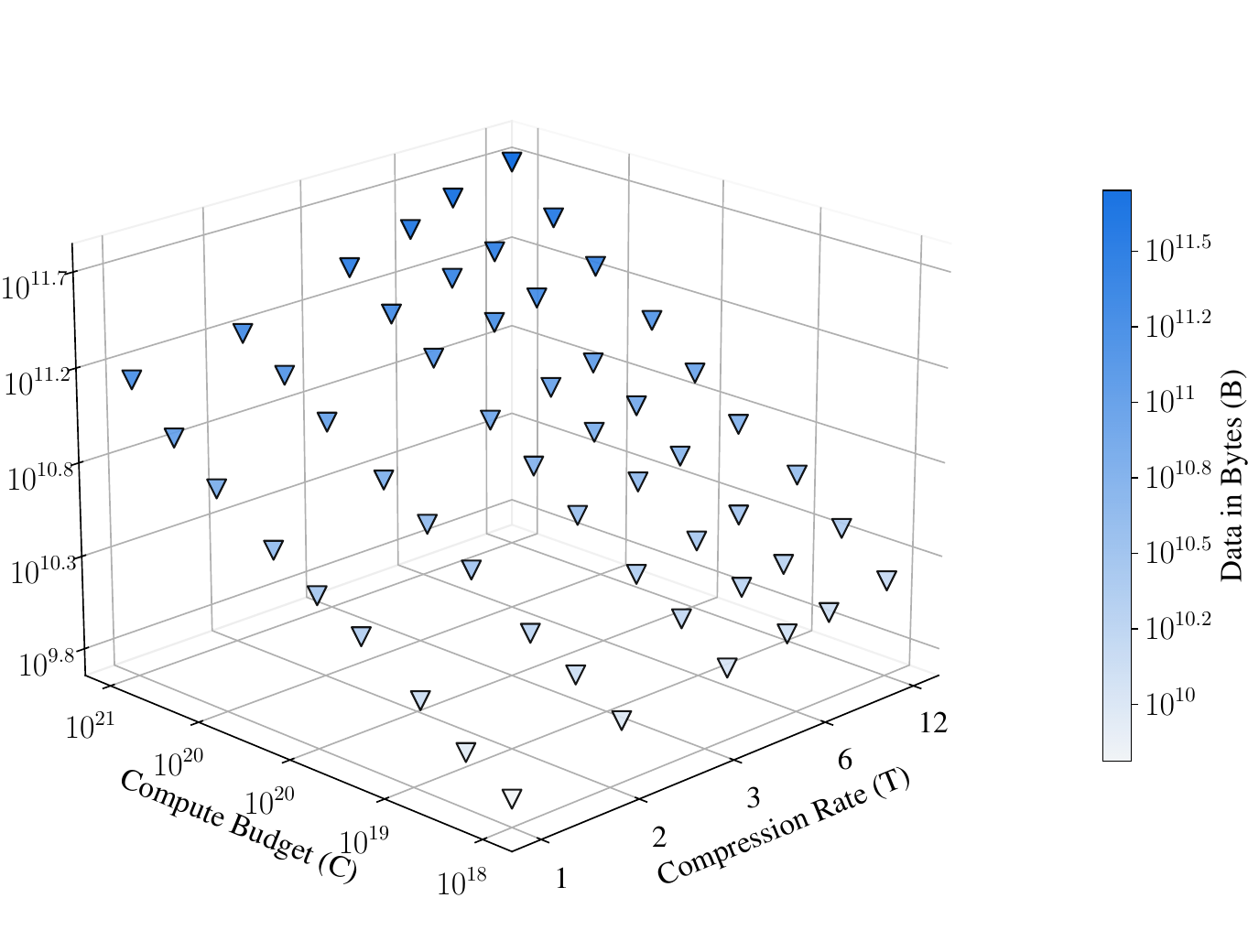}
    \caption{Amount of data}
    \label{fig:optimal_data_blt}
    \end{subfigure}
    \hfill
    \begin{subfigure}[t]{0.49\textwidth}
    \centering
    \includegraphics[width=\linewidth]{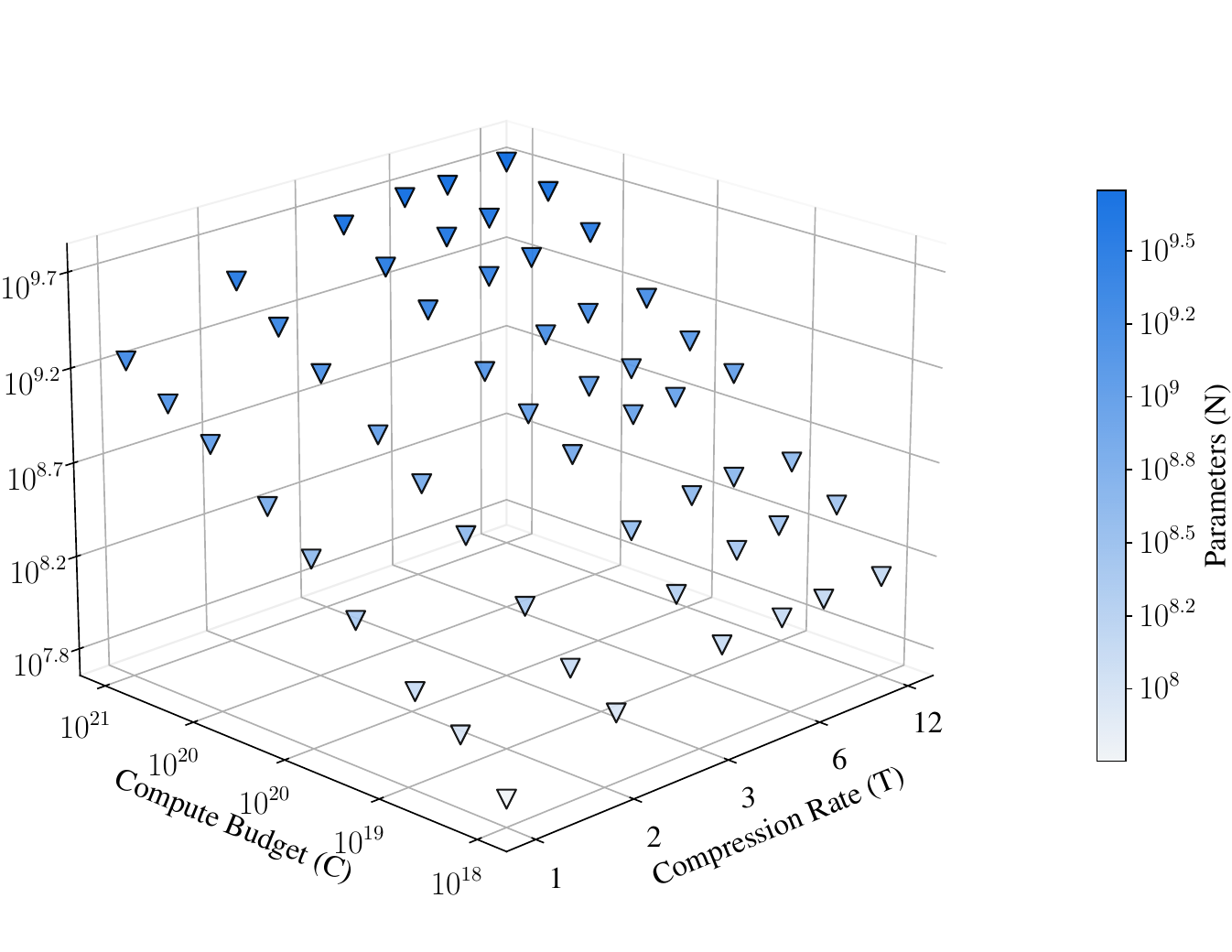}
    \caption{Model size}
    \label{fig:optimal_params_blt}
    \end{subfigure}
    \caption{Optimal data and model size configurations for each compute budget and compression rate (latent tokenized models).}
    \label{fig:optimal_data_params_blt}
\end{figure}

For each compute budget $C$ and compression rate $T$, we identify the optimal training data size by fitting a second-degree polynomial (i.e. IsoFLOP) to the relationship between log-data $\log(B)$ and validation loss $L$.
The optimal data size $B^{\star}$ corresponds to the minimum of this parabola.
We determine the corresponding optimal parameter count $N^{\star}$ via log-linear interpolation.

In the first power law we estimate the optimal training data size $B^{\star}$ as a power law function of compute budget $C$ and compression rate $T$:

\begin{equation}
\label{eq:optimal_data}
B^{\star}(C,T) \cong  B_0C^\alpha T^\beta
\end{equation}

This equation involves three parameters: $B_0$ (initial optimal data), $\alpha$ (scaling with compute), and $\beta$ (scaling with compression).
In this fit, for simplicity and better generalization across tokenizers, we consider only the parameters of the latent module (i.e., excluding encoder/decoder parameters for BLT and embedding parameters for subword models).
Importantly, given our fixed scaling recipe, the number of the model's ``latent'' parameters determines the ``total'' parameter count.
We can approximate the latent module's compute $C$ as:

\begin{equation}
\label{eq:compute_approximation}
C \approx 6N\frac{B}{T}
\end{equation}

Where $\frac{B}{T}$ is the amount of data expressed in tokens, typically denoted as $D$ in other scaling laws works.
Solving Approximation~\ref{eq:optimal_data} allows us to obtain a power trend for optimal global parameter count:

\begin{equation}
\label{eq:optimal_param}
N^{\star}(C,T) \cong \frac{1}{6B_0}C^{1-\alpha}T^{1-\beta} = N_0C^{1-\alpha}T^{1-\beta}
\end{equation}

We also define the optimal Byte-per-Parameter ratio, $\rho^\star = B^\star / N^{\star}$.
Based on the derived power laws, this ratio has the following form:

\begin{equation}
\label{eq:optimal_bpp}
\rho^\star(C,T) \cong \frac{B_0}{N_0}C^{2\alpha-1}T^{2\beta -1}
\end{equation}

\newpage
Before observing the actual fit, we can describe the meaning of specific hypothetical values of $\alpha$ and $\beta$.

\begin{itemize}
    \item When $\alpha\approx 0.5$, $\rho^\star$ would remain constant for varying values of compute budget $C$. This would mean that data and parameters should be scaled in 1:1 proportion. Similar equivalence was observed in \citet{hoffmann2022training}.
    \item Analogously $\beta\approx 0.5$, would indicate that compute unlocked with higher compression should be allocated equally in increase of parameters and training data. Hence, the optimal \bpp{} $\rho^\star$ would remain constant across varying compression rates $T$. 
    \item $\beta\approx 1$  would indicate that we can omit the notion of compression from scaling laws and replace $B$ (amount of data in bytes) with used $D = \frac{B}{T}$ (amount of data in tokens). Such observation would suggest that we should simplify the scaling law to consider data amount in tokens $D^\star$ and neglect the impact of compression (as done in previous scaling studies).
\end{itemize}

\subsection{Scaling Law I: Results}



The IsoFLOPs analysis shows that for a set compute budget $C$, a second degree fit faithfully describes the relationship between logarithm of data size $\log(B)$ and validation loss $L$ (see Figure~\ref{fig:1d_isoflops_blt} in Appendix).
Therefore, we can easily identify the optimal data size $B^\star$ by finding the minimum of the parabola (or paraboloid in the three-dimensional case).

Moreover, the results empirically confirm that the optimal data and parameter count gradually increase with increasing \compr{} $T$, thanks to a decrease in compute cost per byte.
Figure~\ref{fig:stage_1_blt_1e20} indicates that across compression rates the optimal byte-per-parameter ratio $\rho^\star$ is close to constant.
This implies that modifying tokenization (and thus \compr{}) changes the compute optimal relation between tokens and parameters, whereas the relationship between bytes and parameters remains constant.
Therefore, the latter is a more robust way to express the optimal data-to-model-size ratio, and we recommend considering it when designing language models with different tokenizers or vocabularies.


Plotting the values of $B^\star$ and $N^\star$  in Figure~\ref{fig:optimal_data_params_blt}, across $C$ and $T$, we observe a log-log linear relationship proving the adequacy of the power law form in Equation~\ref{eq:optimal_data}.
The fit reveals the following values of parameters: $B_0=11.1$, $N_0=1.5\times10^{-2}$, $\alpha=0.475$, $\beta=0.479$.
Crucially, both the values of $\alpha$ and $\beta$ are close to $0.5$, indicating that the optimal byte-per-parameter ratio is close to constant across varying compute budget and compression rate (see Table~\ref{tab:power_law_fits}).
This allows us to answer the first research question \textbf{R1}:

\begin{keyfinding}[title={Finding 1}]
The optimal ratio between bytes of data and model parameters ($\rho^\star$) remains close to constant across variable compute budget and compression rates.
Therefore, when generalizing a scaling recipe to a model with a different tokenizer, we advise matching the ratio of training bytes (not tokens) to model parameters.
\end{keyfinding}

With the data-to-parameter ratio pinned down, a natural next step is to consider what is optimal value of \compr{} that we should target at scale.

\subsection{Scaling Law II: Optimal Loss Dynamics}
\label{sec:scaling_law_2}

\begin{figure}[!htb]
    \begin{minipage}[t]{0.49\textwidth}
    \centering
    \includegraphics[width=\linewidth]{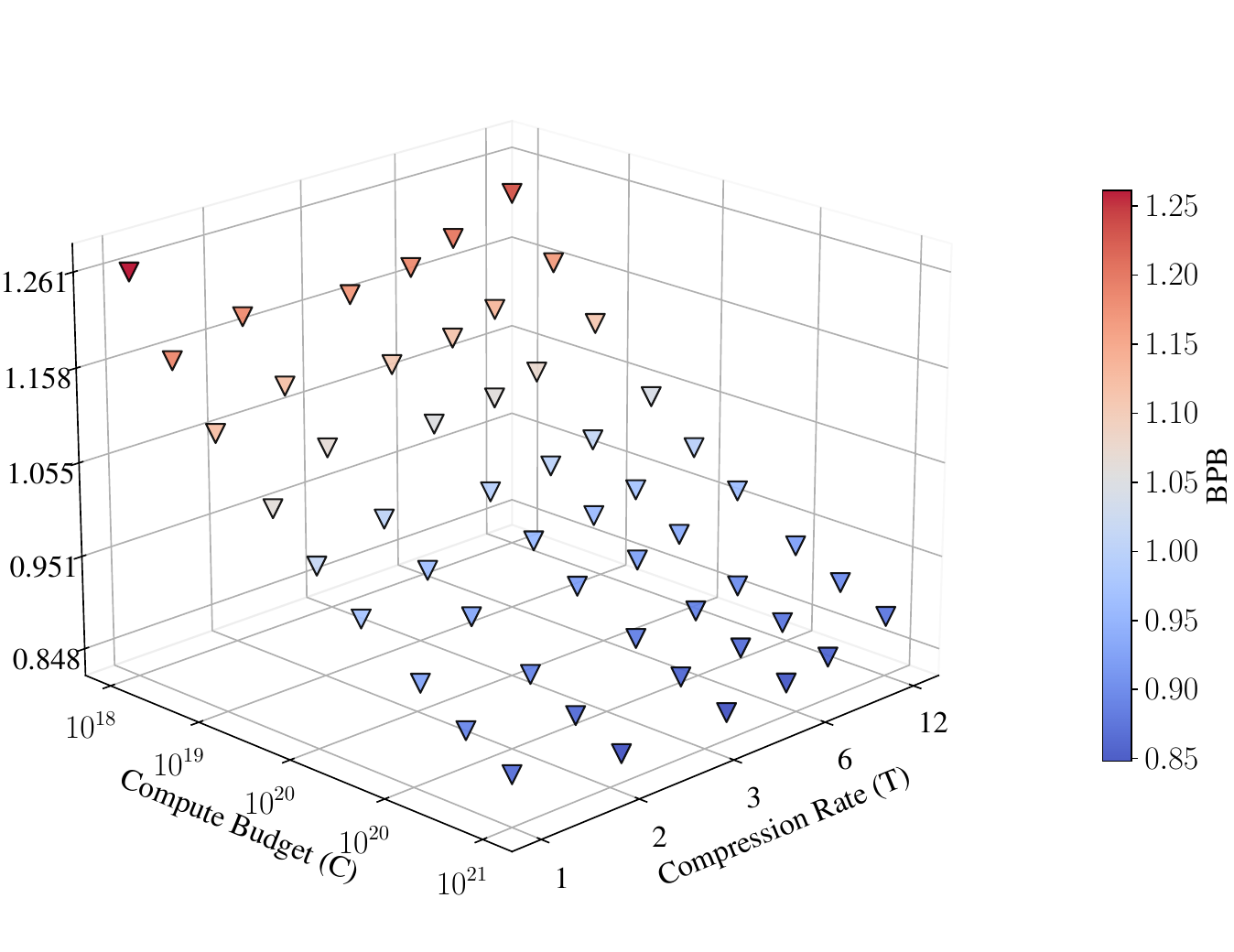}
    \caption{Optimal loss obtained for each compute budget and compression rate with latently tokenized models. Points at {$C=10^{20}$} correspond to the red triangles from Figure~\ref{fig:stage_1_blt_1e20}.}
    \label{fig:optimal_loss_blt}
    \end{minipage}
\hfill
    \begin{minipage}[t]{0.49\textwidth}
    \centering
    \includegraphics[width=\linewidth]{figures/blt_entropy_loss_profile.pdf}
    \caption{Power law fit for loss prediction based on compute budget and compression rate for BLT models. The slices of the fitted manifold for each compute budget (lines) are compared with the optimal loss values (triangles).}
    \label{fig:cr_vs_loss_blt}
    \end{minipage}
\end{figure}


In the next stage, we model the optimal loss $L^\star$, defined as the loss obtained with the optimal data $B^\star$ and parameter count $N^\star$ for a given compute budget and compression rate:

\begin{equation}
\label{eq:optimal_loss_def}
L^\star(C,T) \overset{\mathrm{def}}{=} L(B^\star(C,T), N^\star(C,T))
\end{equation}

We posit that the optimal loss can be approximated by a power law of the form:

\begin{equation}
    \label{eq:optimal_loss}
    L^\star(C,T) \cong  L_0 \times C^\gamma + f(C,T)
\end{equation}

This stage involves fitting three variables: $L_0$ (initial loss), $\gamma < 0$ (scaling with compute), and $f(T)$ (a function representing compression-specific residuals, including irreducible loss).
We do not make a priori assumptions about the form of $f(C,T)$; instead, we fit it empirically based on the results obtained for each compression rate separately.


\subsection{Scaling Law II: Results}


We plot the optimal loss $L^\star$ as a function of compute budget $C$ and compression rate $T$ in Figure~\ref{fig:optimal_loss_blt}.
While expectedly, the loss decreases with increasing compute budget, we observe that the relation between \compr{} $T$ and $L^\star$ is non-monotonic.
Specifically, the loss obtains a minimum for $T^{\star}\approx4$ and rises for both higher and lower \compr{}s. 
We observe a slow decrease of optimal \compr{}
with increase of compute budget.

The power law fit gives us the following values of parameters: $L_0=3342$, $\gamma=-0.206$.
We further examine the distribution of compression-specific offsets $f(C,T)$ in Figures~\ref{fig:optimal_loss_blt}~and~\ref{fig:cr_vs_loss_blt}.
Based on the polynomial profile for $f(\cdot)$ we can estimate with high confidence its form as:\footnote{We discuss empirical derivation of this formula in Appendix~\ref{sec:scaling_law_2_derivation}.}


\begin{equation}
\label{eq:optimal_compression_rate}
    f(C,T) = F \times \log^2\left(\frac{C^\delta T}{T_0} \right) + E
\end{equation}

The best fit was obtained with $F=0.032$, $\delta=0.035$, $T_0=18.2$, and $E=0.70$.
Both visual and power law evidence support the claim that the optimal compression rate $T^\star=\frac{T_0}{C^\delta}$ slowly decreases with training budget, e.g. $T^\star= 3.69$ for $C=10^{20}$ and $T^\star= 3.33 $ for $C=2\times10^{21}$.
This allows us to answer the second research question \textbf{R2}: 

\begin{keyfinding}[title={Finding 2}]
At each training compute budget, there is an optimal \compr{} $T^\star$.
Diverging from its value in either direction increases loss.
We observe decreasing optimal \compr{} for higher training budgets.
\end{keyfinding}

The described scaling laws allow us to identify the values of \bpp{} and \compr{} for compute optimal training.
Another practical consideration is how compute optimal \compr{} impact computation cost of processing inference queries.
We study it in the following subsection.


\subsection{Optimal Tokenization during Inference}
\label{sec:inference-flops}



\begin{figure}[!htb]
    \centering
    \begin{subfigure}[t]{0.49\textwidth}
    \includegraphics[width=\linewidth]{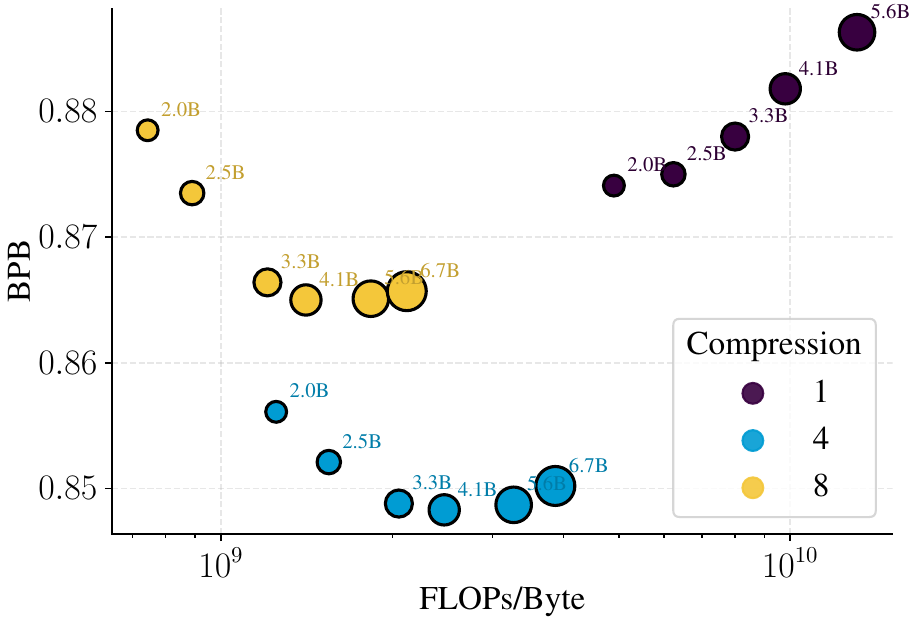}
    \caption{BPB on C4 test-set}
    \label{fig:endtask_c4}
    \end{subfigure}
    \hfill
    \begin{subfigure}[t]{0.49\textwidth}
    \centering
    \includegraphics[width=\linewidth]{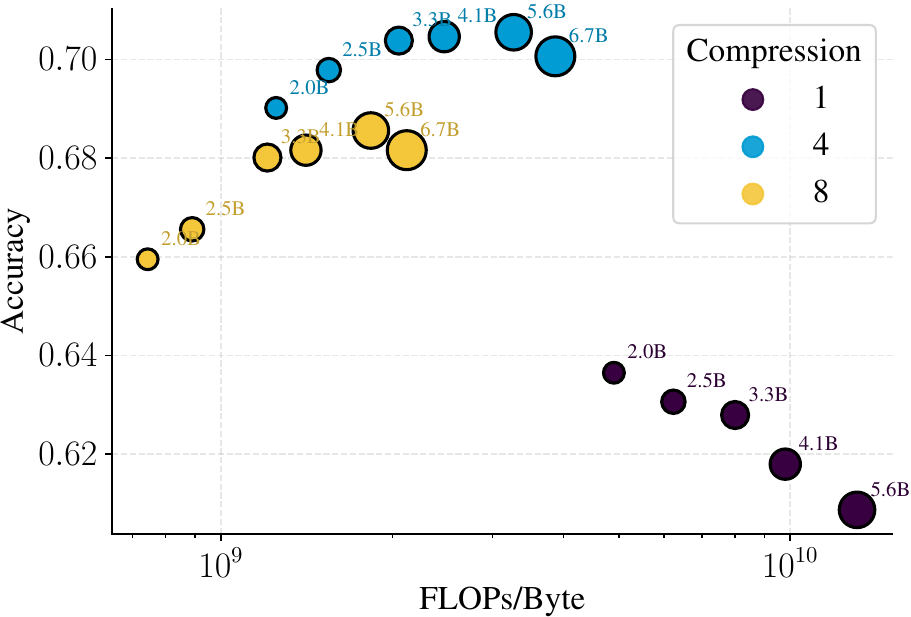}
    \caption{0-shot Accuracy on HellaSwag}
    \label{fig:endtask_hellswag}
    \end{subfigure}
    \caption{Evaluation of the BLT models trained for $C=2\times10^{21}$ FLOPs. Size of the point corresponds to model parameter count in the model. The results are plotted against inference compute cost per byte, which depends on model size $N$ and compression rate $T$.}
\end{figure}

To further study the role of optimal tokenization in inference, we compare the performance of models trained under $C=2\times10^{21}$ budget with different compression rates against their inference cost.
Specifically, we consider the results on language modeling and 0-shot accuracy on HellaSwag generative benchmark \citep{zellers2019hellaswag}. 
In Figure~\ref{fig:endtask_hellswag}, we observe that a higher compression rate decreases the inference compute cost for models of the same size (e.g. 3.3B parameter model with $T=8$ is cheaper to run than a model of the same size $T=4$).
However, we also observe that \compr{} closer to the optimal value improves the results of the inference-compute-matched setting.
For instance, the 3.3B model with \compr{} $T=4$ has a similar inference cost of $2.1\times10^{9}$ FLOPs/Byte as the 6.7B model with \compr{} $T=8$, while the former achieves higher score on the endtask accuracy (70.6\% vs. 68.6\%).
We present further results for AI2 Reasoning Challenge \citep{clark2018arc} in Appendix~\ref{sec:arc-results}.

Till now all the experiments were conducted for latent tokenized models. 
In the following section, we analyze whether the same trends hold for their counterparts with subword tokenization, e.g., BPE.

%% file: sections/04_isotropic.tex
\section{Compute Optimal Subword Tokenization}
\label{sec:subword_scaling}


\input{tables/power_law_fits_new}

In this section, we validate the observations from the previous section for subword tokenized models.
We train models with different subword tokenization algorithms: character-level tokenization, BPE, BPE with vocabulary masking, and SuperBPE to differentiate the values of compression rate $T$.
Then we repeat the analysis of optimal data and parameters configurations and compare the fits of Scaling Laws I and II between latent and subword tokenized models, in order to answer the research question \textbf{R3}.

\subsection{Results}

\begin{figure*}[!htb]
    \centering
    \begin{subfigure}{0.48\textwidth}
        \includegraphics[width=\linewidth]{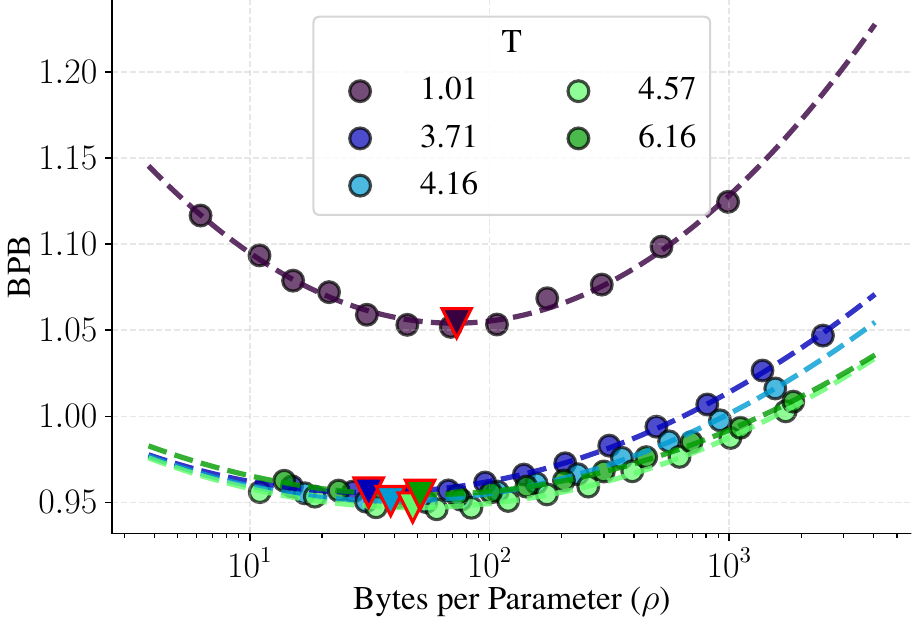}
        \caption{2D IsoFLOP}
    \end{subfigure}
    \hfill
    \begin{subfigure}{0.48\textwidth}
        \includegraphics[width=\linewidth]{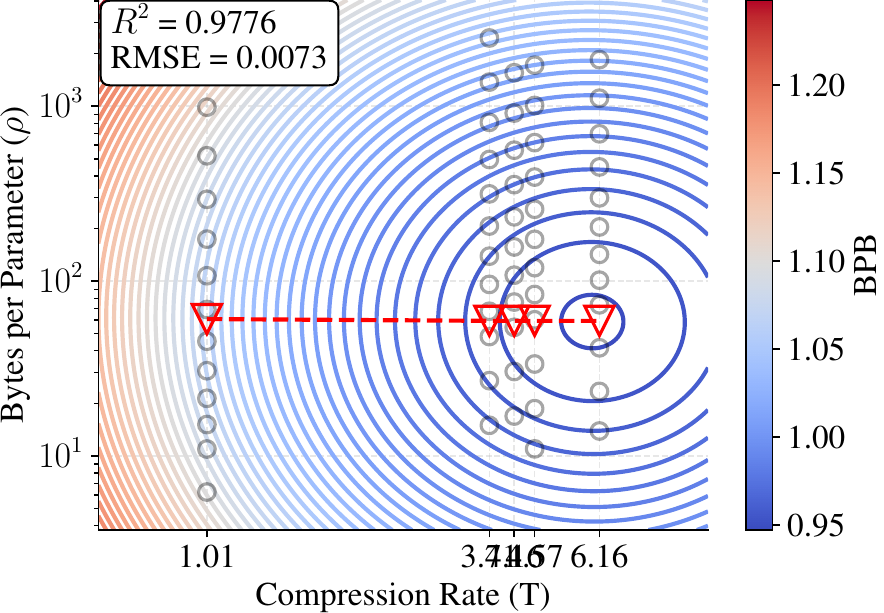}
        \caption{3D IsoFLOP (heatmap)}
    \end{subfigure}
    \caption{
    Evaluation scores of subword tokenized models on C4 test set with fixed FLOPs budget ($C=10^{20}$), compared against \bpp{} ratio.
    Different subword tokenization algorithms are obtain varying compression rates: 1.01 for character-level tokenization, 3.71, 4.16, and 4.57 for BPE, 6.16 for SuperBPE
    2-dimensional IsoFLOP (parabola) were fitted for each compression rate, while 3-dimensional IsoFLOP jointly for all compression rates (on x-axis). 
    Similar to latent tokenized models, minima of both fits show that minimal loss is obtained at almost constant value.
    For IsoFLOPs as function of data, parameters, and for other compute budgets, refer to Appendix~\ref{sec:isoflop-across-budgets}}
    \label{fig:stage_1_iso_1e20}
\end{figure*}

\begin{figure}[!htb]
    \centering
    \begin{minipage}[t]{0.49\textwidth}
    \centering
    \includegraphics[width=\linewidth]{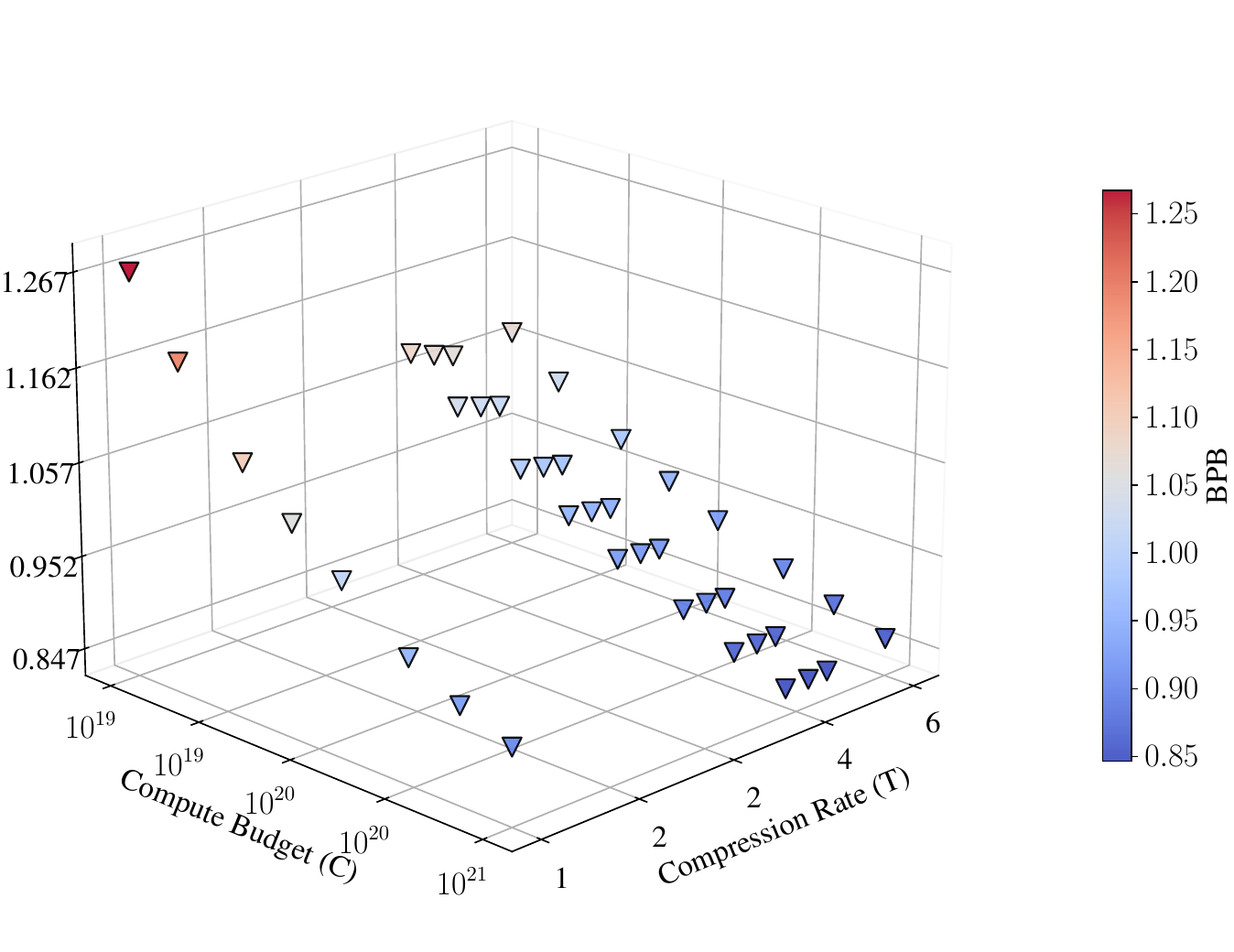}
    \caption{Optimal loss obtained for each compute budget and compression rate. Points at {$C=10^{20}$} correspond to data red triangles from Figure~\ref{fig:stage_1_iso_1e20}.}
    \label{fig:optimal_loss_iso}
    \end{minipage}
\hfill
    \begin{minipage}[t]{0.49\textwidth}
    \centering
    \includegraphics[width=\linewidth]{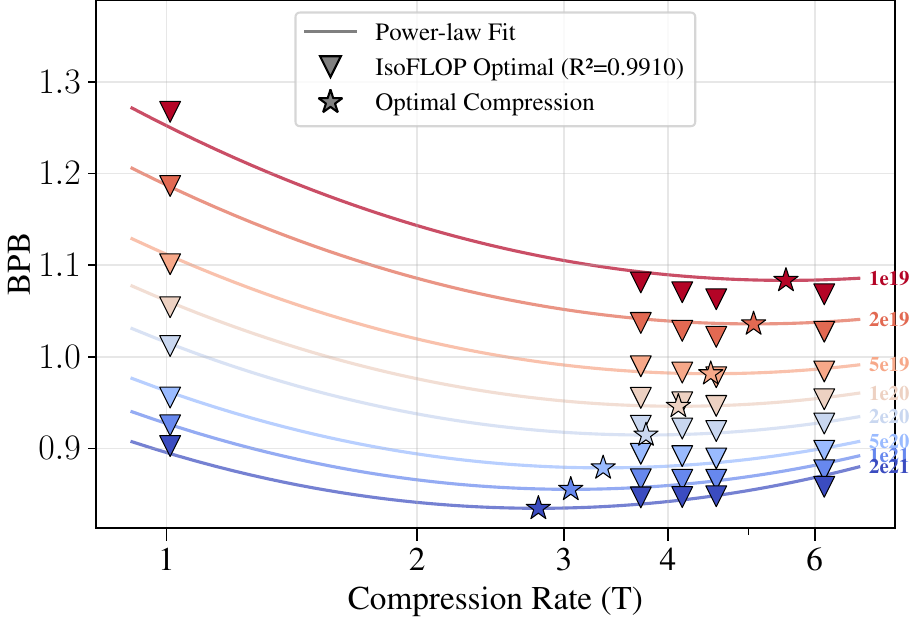}
    \caption{Power law fit for loss prediction based on compute budget and compression rate for isotropic models. The slices of the fitted manifold for each compute budget (lines) are compared with the optimal loss values (triangles).}
    \label{fig:cr_vs_loss_iso}
    \end{minipage}
\end{figure}

Similar to the latent tokenization case, the IsoFLOPs curves allow us to identify the optimal data amount in bytes $B^\star$ for a fixed compute budget $C$.
Figure~\ref{fig:stage_1_iso_1e20} shows the optimal values for a specific compute budget.
We observe that the optimal byte-per-parameter ratio $\rho^\star$ is similar across tokenizers.

The Scaling Law I fit shows results close to the latent tokenization case: $B_0=2.8$, $N_0=59\times10^{-3}$, $\alpha=0.501$, $\beta=0.446$.
Also, loss dynamics presented in Figure~\ref{fig:optimal_loss_iso} and the Scaling Law II fit show similar results as the latent tokenization case: $L_0=1087$, $\gamma=-0.181$, $F=0.0575$, $\delta=0.129$, $T_0=1577$, and $E=0.680$.
The fit values are compared with latent tokenized models, as shown in~Table~\ref{tab:power_law_fits}.

\input{tables/isotropic_budget_bpb_comparison}

Similar to the latent case, the Scaling Law II fit yields a positive compression–compute coupling $\delta = 0.129$, with a 95\% confidence interval of $[0.108, 0.150]$ that excludes zero. This is our core evidence that the compute-optimal compression rate $T^\star$ decreases as the training budget grows, i.e. lower compression becomes preferable at larger scale  (Figure~\ref{fig:cr_vs_loss_iso}).
Consistent with this trend, at our highest compute budget the models with 75\% and 90\% of the BPE vocabulary masked reach a marginally lower BPB than the original BPE tokenizer, despite spending the same compute in the embedding and de-embedding layers (Table~\ref{tab:isotropic_budget_bpb_comparison}).
We note that these differences are small and our claim that lower compression benefits larger-scale training rests on the statistically confirmed positive $\delta$ rather than on these pointwise comparisons.

These observations allow us to answer the question \textbf{R3}:

\begin{keyfinding}[title={Finding 3}]
Discovered scaling trends for models with latent tokenization (BLT) hold for models with subword tokenization (BPE, SuperBPE).
\end{keyfinding}

We have thoroughly described the trends for compute optimal \bpp{} and \compr{} for training and evaluation on English datasets.
In the next section, we are asking whether and how these findings generalize to other languages.

%% file: tables/power_law_fits_new.tex
\begin{table}[tbp]
\centering
    \centering
    \rowcolors{3}{metabg}{white}
    \begin{tabular}{l cc cc}
    \toprule
    & \multicolumn{2}{c}{\textbf{Latent}} & \multicolumn{2}{c}{\textbf{Subword}} \\
    \cmidrule(lr){2-3}\cmidrule(lr){4-5}
    \textbf{Param} & Value & 95\% CI & Value & 95\% CI \\
    \midrule
    $\alpha$ & 0.475 & [0.467,0.483] & 0.501 & [0.471,0.532] \\
    $\beta$  & 0.479 & [0.458,0.500] & 0.446 & [0.387,0.506] \\
    $B_0$    & 11.1  & [7.7,16.0]    & 2.8   & [0.7,11.0] \\
    $N_0$    & 0.015 & [0.0104,0.0217] & 0.059 & [0.015,0.229] \\ \midrule
    $\gamma$ & -0.206 & [-0.217,-0.195] & -0.181 & [-0.226,-0.135] \\
    $L_0$    & 3342  & [2114,5286]   & 1087 & [171,6896] \\ \midrule
    $F$      & 0.032 & [0.030,0.035] & 0.058 & [0.051,0.064] \\
    $E$      & 0.703 & [0.688,0.718] & 0.680 & [0.618,0.741] \\
    $\delta$ & 0.035 & [0.022,0.047] & 0.129 & [0.108,0.150] \\
    $T_0$    & 18.2  & [10.2,32.4]   & 1577 & [569,4369] \\
    \bottomrule
    \end{tabular}
    \captionof{table}{Fitted power law parameters for the families of latent and subword tokenized models.
    The 95\% confidence intervals were computed with a numeric Hessian.
    }
    \label{tab:power_law_fits}
\end{table}

%% file: tables/isotropic_budget_bpb_comparison.tex
\begin{table}[tbp]
\centering
\rowcolors{4}{metabg}{white}
\setlength{\tabcolsep}{4pt}
\renewcommand{\arraystretch}{1.2}
\begin{tabular}{lccccc}
\toprule
\textbf{Compute} & \textbf{Character} & \multicolumn{3}{c}{\textbf{BPE}} & \textbf{SuperBPE} \\ 
\cmidrule(lr){3-5} 
\textbf{(FLOPs)} &  & V. mask=90\% & V. mask=75\% & Original &  \\ \midrule
$1 \times 10^{19}$ & 1.2678 & 1.0819 & 1.0709 & \textbf{1.0635} & 1.0682 \\
$2 \times 10^{19}$ & 1.1812 & 1.0381 & 1.0281 & \textbf{1.0214} & 1.0273 \\
$5 \times 10^{19}$ & 1.0989 & 0.9887 & 0.9819 & \textbf{0.9769} & 0.9840 \\
$1 \times 10^{20}$ & 1.0519 & 0.9554 & 0.9502 & \textbf{0.9461} & 0.9532 \\
$2 \times 10^{20}$ & 1.0126 & 0.9254 & 0.9220 & \textbf{0.9186} & 0.9272 \\
$5 \times 10^{20}$ & 0.9556 & 0.8942 & 0.8916 & \textbf{0.8891} & 0.8976 \\
$1 \times 10^{21}$ & 0.9253 & 0.8665 & \textbf{0.8658} & 0.8659 & 0.8763 \\
$2 \times 10^{21}$ & 0.9027 & \textbf{0.8466} & 0.8469 & 0.8479 & 0.8582 \\ \midrule
\textbf{Compression:} & 1.01 & 3.71 & 4.16 & 4.57 & 6.16 \\ \bottomrule
\end{tabular}
\caption{Comparison of the lowest BPB obtained by subword tokenized models for specific compute budgets.
}
\label{tab:isotropic_budget_bpb_comparison}
\end{table}

%% file: sections/05_across_domains.tex
\section{Compute Optimal Tokenization Beyond English}
\label{sec:beyond_english}

\begin{figure*}[!htb]
    \centering
    \begin{subfigure}{0.48\textwidth}
        \includegraphics[width=\linewidth]{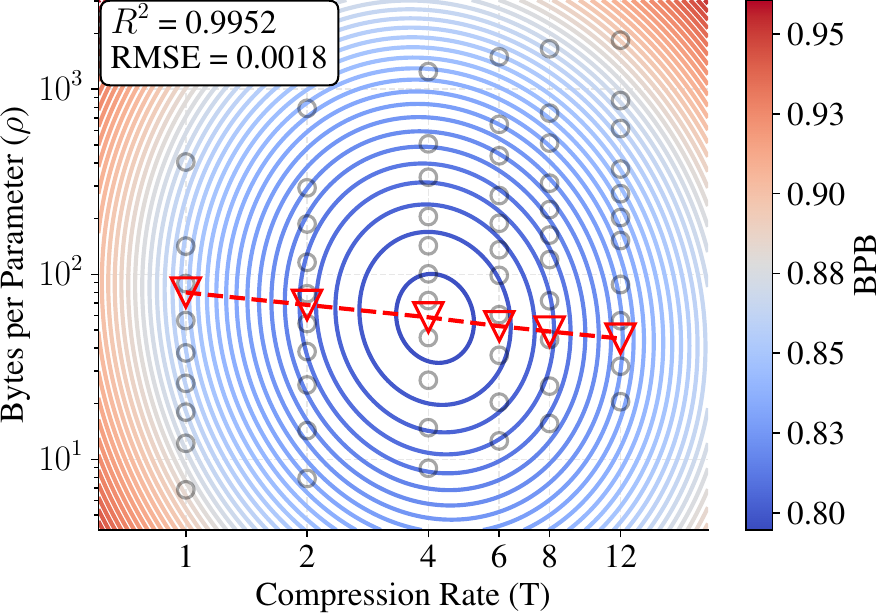}
        \caption{French (Latin)}
    \end{subfigure}
    \begin{subfigure}{0.48\textwidth}
        \includegraphics[width=\linewidth]{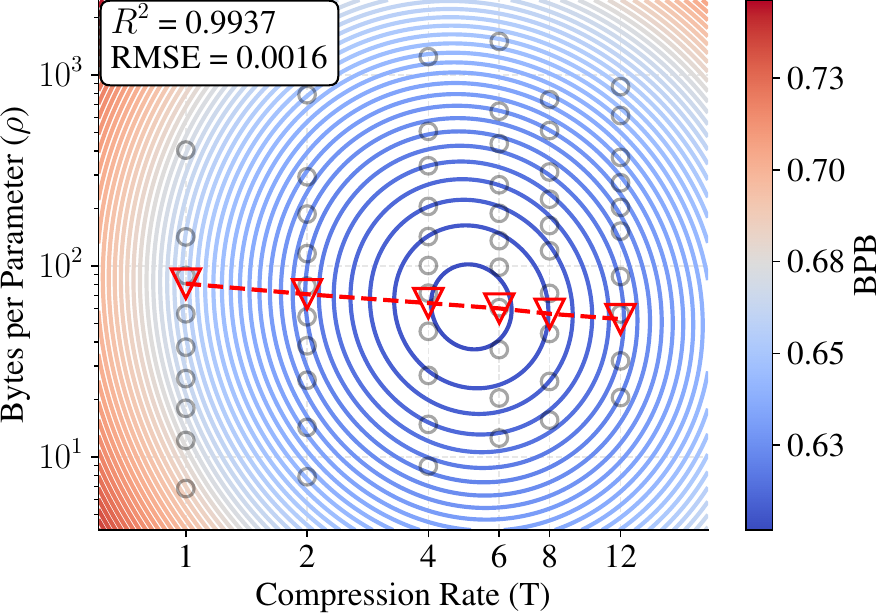}
        \caption{Vietnamese (Latin)}
    \end{subfigure}
    \begin{subfigure}{0.48\textwidth}
        \includegraphics[width=\linewidth]{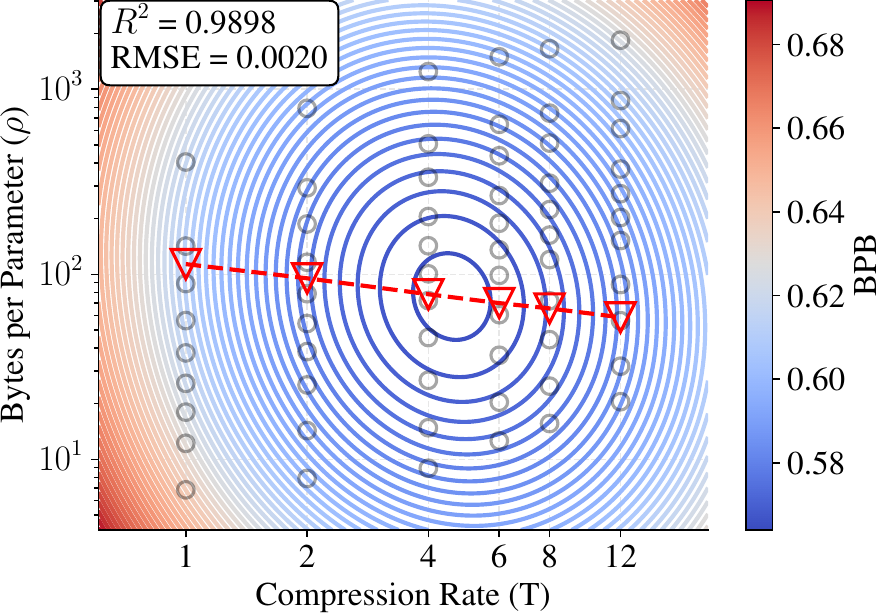}
        \caption{Arabic (Arabic)}
    \end{subfigure}
    \begin{subfigure}{0.48\textwidth}
        \includegraphics[width=\linewidth]{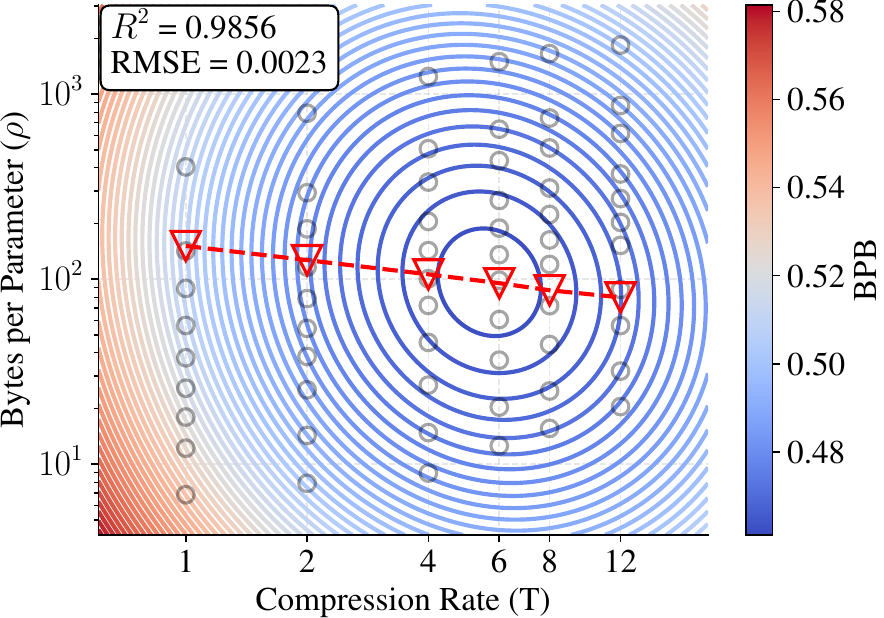}
        \caption{Russian (Cyrillic)}
    \end{subfigure}
        \begin{subfigure}{0.48\textwidth}
        \includegraphics[width=\linewidth]{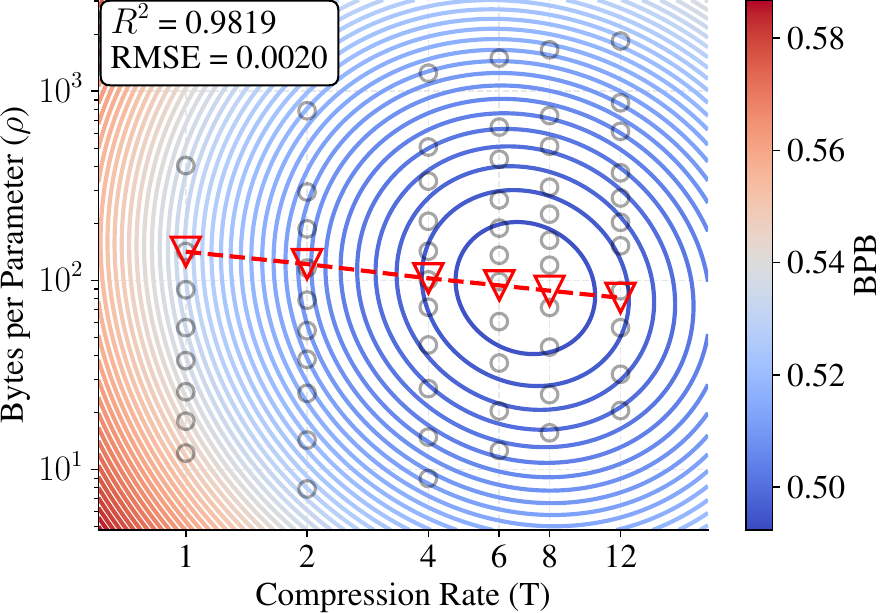}
        \caption{English $\times2$ (Latin)}
    \end{subfigure}
    \begin{subfigure}{0.48\textwidth}
        \includegraphics[width=\linewidth]{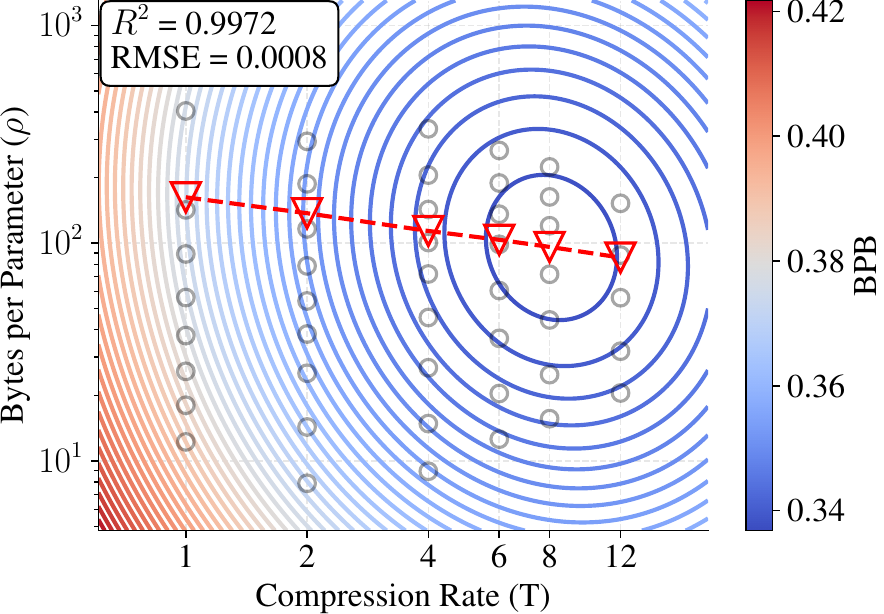}
        \caption{Hindi (Devanagari)}
    \end{subfigure}
    \caption{3D IsoFLOP (heatmap) fits across languages ($C=10^{20}$) as function of \bpp{} and \compr{} for six languages. All models use latent tokenization to achieve the set compression. IsoFLOPs are fitted jointly for all compression rates. 
    }
    \label{fig:2d_xlingual_1e20}
\end{figure*}

To test how the language choice affects the compute-optimal compression rate and \bpp{} ratio, we extend our experiments to five languages with diverse writing scripts: French (Latin), Vietnamese (Latin), Russian (Cyrillic), Arabic (Arabic), Hindi (Devanagari).
We also create an artificially inflated version of English data by adding a dummy byte between pairs of original UTF-8 bytes. Such English $\times2$ data represent the same information at half the density.

For this purpose, we train latent-tokenized models (BLT) on monolingual data from FineWeb-2 \citep{penedo2025fineweb}, training a separate set of models for each language.\footnote{The results for jointly trained multilingual models are in Appendix~\ref{sec:beyond_english_mix}).}
We evaluate each model on the corresponding test split from the same source.
For English $\times2$, we use an inflated version of the C4 test set used in the previous experiments.

Our training setup is analogous to the one described in Section~\ref{sec:compression_scaling} for English.
For each language $l$, we fix the training budget to  $C=10^{20}$ FLOPs.
The IsoFLOPs analysis (similar to that presented in Figure~\ref{fig:stage_1_blt_1e20}) allows us to identify the compute optimal \bpp{} ratio $\rho_l^\star$ and \compr{} $T_l^\star$, for each language $l$ at fixed compute budget.

Further, we compare these values to cross-lingual \pari{}, defined as the proportion between the amount of bytes required to express the same information in different languages \citep{petrov2023token_unfairness}.
We estimate \pari{} by dividing the byte length of sentences in each language by the byte length of their English translations.
We use translations from FLORES-200 multi-parallel corpus \citep{goyal2021flores,nllb2022}, which test split contains 1000 English sentences and their translations in a wide range of languages.

These experiments address our last research question \textbf{R4}.

\subsection{Results}




\begin{figure}[t]
\centering
    \begin{minipage}[t]{0.49\textwidth}
    \centering
    \includegraphics[width=\linewidth]{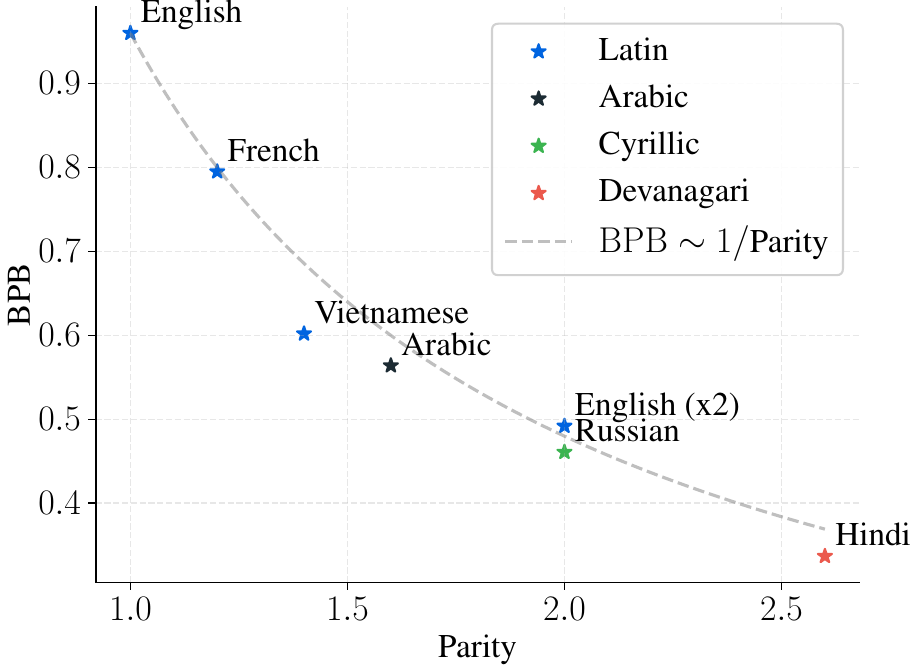}
    \caption{Language specific minimal loss compared against \pari{}.
    We observe that BPB is inversely proportional to \pari{}.}
    \label{fig:xlingual_bpbs_1e20}
    \end{minipage}
\hfill
    \begin{minipage}[t]{0.49\textwidth}
    \centering
    \includegraphics[width=\linewidth]{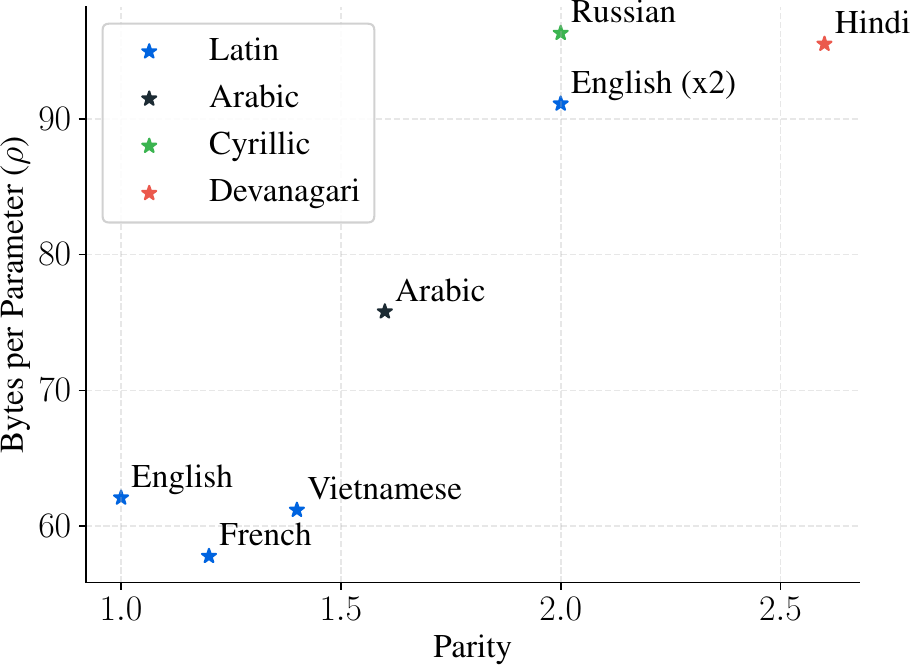}
    \caption{Language specific optimal \bpp{} ratio ($\rho^\star_l$).
    Lower information density (high \pari{}) correlates with preference of large training data size over model size (high $\rho^\star_l$).}
    \label{fig:xlingual_bpp_1e20}
    \end{minipage}
\end{figure}

\input{tables/xlingual}

Figure~\ref{fig:2d_xlingual_1e20} present the results of the IsoFLOPs analysis across all analyzed languages.
Similarly to English, we observe that the minimal loss is achieved by models with close to constant \bpp{} ($\rho^\star_l$).
From the polynomial fit, we estimate the compute-optimal  \bpp{} ratio and \compr{} by analytically finding the coordinates of the global minimum (i.e. lowest loss).

\begin{figure}[t]
\centering
    \begin{minipage}[t]{0.49\textwidth}
    \centering
    \includegraphics[width=\linewidth]{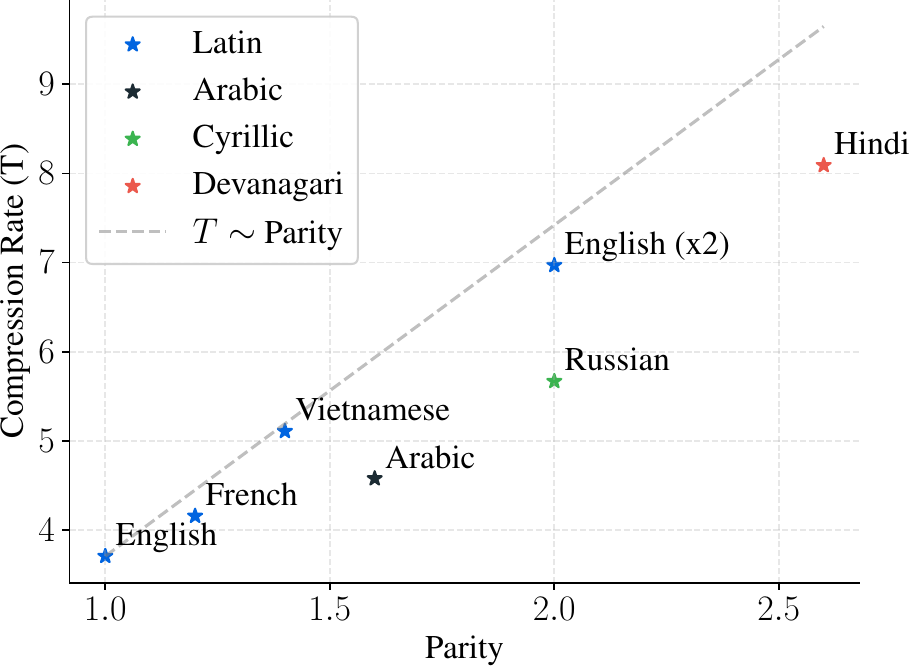}
    \caption{Language specific compression rate ($T^\star_l$) compared against \pari{}.
    We observe that languages with higher \pari{} prefer tokenization with higher compression.}
    \label{fig:xlingual_crs_1e20}
    \end{minipage}
\hfill
    \begin{minipage}[t]{0.49\textwidth}
    \includegraphics[width=\linewidth]{figures/tokenization_compression_vs_optimal.pdf}
    \caption{Compute optimal compression differs from the data compression obtained by popular language models. We observe that  subword tokenizers tend to over-compress high-resource languages, e.g.: English, Arabic, while significantly undercompressing the less resourced ones, e.g.: Vietnamese, Hindi.}
    \label{fig:tokenizers_crs}
    \end{minipage}
\end{figure}

Notably, the $\rho^\star_l$ is language dependent (e.g. $\rho^\star_\text{AR}\approx75.8$; $\rho^\star_\text{RU}\approx96.3$).
We also observe language-dependent differences in the compute-optimal compression rate  (e.g. $T^\star_\text{AR}\approx4.58$; $T^\star_\text{RU}\approx5.67$).
In Table~\ref{tab:xlingual_optimals_1e20}, we compare these compute-optimal values to cross-lingual \pari{}.
We observe that the optimal values depend on language and its \pari{}.
Figure~\ref{fig:xlingual_bpbs_1e20} shows that language-specific BPB scales inversely with parity.
This confirms the observation from \citet{limisiewicz-etal-2024-myte}: under optimal tokenization, similar information expressed across languages has similar likelihood.
We further observe that \pari{} correlates with optimal \bpp{}, which is explained by the fact that more coarsely encoded languages tend to benefit more from additional training data than from larger models (Figure~\ref{fig:xlingual_bpp_1e20}).
While, in joint multilingual training the optimal \bpp{} ratios converge to the same value across languages (see Appendix~\ref{sec:beyond_english_mix}).

Lastly, we observe that higher \pari{} translates to a higher optimal \compr{}.
For Latin-script languages, this relationship is close to a 1:1 increase (Figure~\ref{fig:xlingual_crs_1e20}).
Importantly, the compression achieved by popular multilingual tokenizers: Llama 3 \citep{llama3herd2024}, Qwen 3 \citep{qwen3technicalreport}, and EuroLLM \citep{martins2024eurollm}, differs from the optimal value, as seen in Figure~\ref{fig:tokenizers_crs}.
These tokenizers tend to over-compress high-resource languages while under-compressing lower-resource ones.

These results bring an answer to the last research question \textbf{[R4]}:

\begin{keyfinding}[title={Finding 4}]
The  optimal byte of data to parameter ratio ($\rho^\star$) and \compr{} ($T^\star$) vary across different languages.
Both are correlated with average information value of bytes in a given language (measured by \pari{}).
\end{keyfinding}

%% file: tables/xlingual.tex
\begin{table}[tbp]
\centering
\rowcolors{3}{metabg}{white}
\setlength{\tabcolsep}{4pt}
\renewcommand{\arraystretch}{1.2}
\begin{tabular}{ccccccc}
\toprule
\textbf{Language} & \textbf{Parity} & \multicolumn{2}{c}{\textbf{$\rho^\star_l$}} & \multicolumn{2}{c}{\textbf{$T^\star_l$}} & \textbf{BPB} \\ \cmidrule(lr){3-4} \cmidrule(lr){5-6}
 &  & \textbf{Value} & \textbf{Ratio} & \textbf{Value} & \textbf{Ratio} & \\ \midrule
English & 1.0 & 62.1 & 1.0 & 3.71 & 1.0 & 0.960 \\
French & 1.2 & 57.8 & 0.93 & 4.16 & 1.12 & 0.795 \\
Vietnamese & 1.4 & 61.2 & 0.99 & 5.11 & 1.38 & 0.602 \\
Arabic & 1.6 & 75.8 & 1.22 & 4.58 & 1.23 &  0.564 \\
Russian & 2.0 & 96.3 & 1.55 & 5.67 & 1.52 & 0.461 \\
English (x2) & 2.0 &  91.1 & 1.47 & 6.97 & 1.87 & 0.492 \\
Hindi & 2.6  &  95.5 & 1.54  & 8.09 & 2.18 & 0.337 \\
\bottomrule
\end{tabular}
\caption{Compute-optimal byte-per-parameter ($\rho^\star_l$), \compr{} ($T^\star_l$) compared to cross-lingual \pari{}.
Results for monolingual models, with $C=10^{20}$ FLOPs budget.
The \pari{} and compute-optimal ratios are proportions between each language and English baseline.}
\label{tab:xlingual_optimals_1e20}
\end{table}

%% file: sections/06_related_work.tex
\section{Related Work}
\subsection{Data Compression in Scaling Laws}

Foundational studies on neural scaling laws, such as those by \citet{kaplan2020scaling} and \citet{hoffmann2022training}, have primarily focused on the relationship between model size, dataset size (in tokens), and compute.
Subsequent works \citep{Pearce2024reconciling,porian2024resolving} have pointed out that the decision of whether to include vocabulary embeddings in the analysis was one of the causes of divergence between scaling laws derived in these studies.
\citet{hoffmann2022training} propose a compute-optimal training ratio of approximately $20$ tokens per parameter.
However, they assume a fixed tokenization scheme, overlooking the information content of the tokens themselves.
We generalize this scaling rule across tokenizers and express it as a comprehensive byte-per-parameter ratio: $\rho^\star\approx60$ (for English data).

\citet{tao2024scaling} derived scaling laws for vocabulary size in BPE-tokenized models. Their study explores how varying vocabulary size impacts computational cost and performance.
They also consider the importance of \compr{} in model scaling, which is indirectly controlled by the vocabulary size.
By considering a broad scope of compression values and compute budgets, we show that the benefits of scaling up vocabulary diminish at larger scales.
We further discuss differences between experimental settings in Appendix~\ref{sec:comparison_tao_et_al}.

Multiple recent works discussed language model scaling trends across domains and languages.
\citet{yang2025scalinglawscode} derived scaling laws across programming languages, showing that language-specific data composition significantly affects scaling behavior.
In the multilingual space, \citet{he-etal-2025-scaling} established per-language scaling laws, while \citet{longpre2026atlas} studied the dynamics of cross-lingual transfer at scale.
Overall, these works demonstrate that scaling laws differ across domains and languages, as we have also observed in our multilingual experiments.

\subsection{Search for Optimal Tokenization}

The research community has long sought to identify tokenizer properties that correlate with language model performance.
The compression rate, or its proxies such as \emph{fertility}, have been identified as a significant factor, especially in the multilingual setting.
\citet{rust-etal-2021-good} observed that in multilingual language models, monolingual tokenizers with higher in-language compression outperform multilingual ones.
Similarly, \citet{limisiewicz-etal-2023-tokenization, goldman-etal-2024-unpacking} noted the benefits of higher \compr{} for certain downstream tasks in multilingual models.
\citet{galle-2019-investigating} show that higher compression is also beneficial for machine translation.
However, in the subword tokenizers considered in these works, language-specific compression depends on the representation of the language in the training corpora.
Thus, compression could be a proxy for the root cause of the performance differences, namely language frequency in the data mix.

In monolingual (English) models, \citet{schmidt-etal-2024-tokenization} argued that higher compression is not inherently beneficial.
However, \citet{liu2025superbpespacetravellanguage} observed an upward trend in downstream task performance with higher compression, even when perplexity (measured in bits per byte) degraded.
This discrepancy underscores the importance of evaluating downstream performance alongside language modeling metrics.

In contrast to prior work, our extensive search reveals that the impact of \compr{} on performance is non-monotonic: there exists an optimal \compr{} beyond which performance degrades.
We also observe a preference for lower compression in longer training.
The recurring prior-work assumption linking higher \compr{} with performance improvement in the multilingual setting may stem from the fact that subword tokenizers typically result in a lower-than-optimal \compr{} for low-resource languages, as shown in Figure~\ref{fig:tokenizers_crs}.


\subsection{Scaling Latent Tokenized Models}

We employ BLT \citep{pagnoni-etal-2025-byte} as our primary framework for studying scaling laws with variable compression.
Techniques such as entropy-based or static patching allow precise control of the compression rate across a wide range. While promising, the data efficiency of training and inference for dynamically tokenized models has not yet been comprehensively studied in the context of scaling laws.

Recent results in latent tokenization suggest that this approach yields greater gains at scale.
The works of \citet{pagnoni-etal-2025-byte, hwang2025dynamicchunkingendtoendhierarchical, neitemeier2025hierarchical, nawrot-etal-2023-efficient} demonstrate that, given sufficient training compute, hierarchical models can surpass their subword-tokenized counterparts.
Furthermore, latent tokenization allows adjusting compression for specific languages \citep{ahia-etal-2024-magnet,owodunni2025flexitokens}.
 Based on our findings, we expect such approaches to be particularly beneficial for multilingual language modeling.

%% file: sections/07_discussion.tex
\section{Discussion}

The relationship between scaling laws and data compression highlights the importance of considering tokenizer \compr{} in the optimal design of large language models.
Our observations overlap with the \citet{hoffmann2022training} (Chinchilla) recipe, suggesting that data and model parameters should be scaled proportionally.
Generalizing the Chinchilla rule, we show that the appropriate unit for data quantity is bytes, not tokens.
Therefore, the widely accepted rule of using approximately 
$20$ tokens per parameter for compute-optimal training holds only under \compr{} specific to a BPE tokenizer.
We generalize this rule (Scaling Law I) by empirically showing that compute-optimal architectures for English text should use approximately $60$ \bpp{}, regardless of data compression.
This generalization makes it easy to transfer efficient training settings across different tokenization schemes, spanning from byte-level to superword-level tokens.

Furthermore, Scaling Law II reveals the existence of an optimal \compr{} that depends on the training domain and the compute budget.
Interestingly, when training on English data with a small FLOP budget, the optimal \compr{} is close to that of a BPE tokenizer.
However, we observe a slow decrease as training compute increases.
This observation also holds for subword-tokenized models.
Strikingly, a model with $90\%$ of the BPE vocabulary masked performs slightly better than standard BPE in our largest runs (even though both spend the same compute in the embedding and de-embedding layers).
This surprising result suggests that, for compute-efficient training of large models, it could be beneficial to decrease vocabulary size or apply techniques such as BPE-dropout \citep{provilkov-etal-2020-bpe}.
Why do we observe such a counterintuitive result?
Our hypothesis is that less-compressed tokenizers allow the model to use more compute at inference time by dividing each evaluation sample into more tokens that are processed by the model.
It is important to keep in mind that lower compression naturally increases the cost of model usage (as shown in Section~\ref{sec:inference-flops}).
Therefore, when controlling for \compr{}, we should consider the trade-off between performance and inference cost.
Specifically, it is advisable to use higher \compr{} to decrease model usage cost, similarly to how model developers opt for over-training language models to boost the performance of relatively smaller (and thus cheaper) models compared to their training-compute-optimal counterparts.

The search for compute-optimal \compr{} is especially important for languages other than English, where the compression obtained by subword tokenizers tends to diverge from the optimal value to a more extreme extent.
Previously, it was thought that multilingual performance of language models is affected by over-segmentation, and many studies focused on increasing compression for better multilingual performance \citep{rust-etal-2021-good, limisiewicz-etal-2023-tokenization}.
We observe, across all considered languages, that overly high compression deteriorates results.
Furthermore, for each language we find a specific optimal \compr{}, the value of which is correlated with the relative information density of the text, i.e., \emph{parity}.
This observation highlights the importance of identifying and achieving an optimal \compr{} for each of the modeled languages.
For statistics-based subword tokenizers (such as BPE), \compr{} is heavily impacted by the amount of in-language data in the training corpus \citep{ahia-etal-2023-languages} and encoding efficiency~\citep{limisiewicz-etal-2024-myte}, and thus cannot be easily controlled in a massively multilingual setting.
This limitation provides a strong argument for latent tokenizers in multilingual language modeling, whose compression can be adapted for specific languages \citep{ahia-etal-2024-magnet, owodunni2025flexitokens}.






\subsection{Future Work}

While our study is the most comprehensive investigation of the impact of tokenization on scaling laws to date, there are multiple directions for future work in this area:

\paragraph{Optimal Compression for Other Modalities.}
In this work, we focused on text data.
We expect the impact of data compression to be equally relevant for other modalities, such as vision, speech, and code.
Currently, each modality utilizes a different set of tokenization techniques, such as variational autoencoders~\citep{oord2017vqvae} or vision transformers~\citep{yu2024image} for images.
Therefore, the scaling analysis requires considering the impact of modality-specific tokenization artifacts.

\paragraph{Sparse Architectures.}
Another direction is considering architectures other than dense (hierarchical) transformer models, such as Mixture of Experts (MoE) models.
Studying the role of data compression could answer the question of how it interacts with parameter sparsity, and thus could be an important contribution to MoE scaling laws~\citep{ludziejewski2024moe}.


\subsection{Limitations}

To keep the study tractable, we fixed several training hyperparameters across all runs.
In particular, we did not tune the learning rate for specific training budgets or adapt any hyperparameters other than those named.


While we have examined a wide range of tokenization methods, spanning both latent and subword families, there could be other design choices that affect the results.
Examples of such aspects include pre-tokenization rules, other subword algorithms (e.g., Unigram; \citealp{kudo-2018-subword}), or token boundary prediction for latent tokenizers.
We expect that such changes would have a minor effect on the main findings of this work.
In Appendix~\ref{sec:tokenization_method_comparison}, we provide a further comparison across tokenization methods.

%% file: sections/08_conclusion.tex
\section{Conclusion}

In this work, we have systematically studied the role of data compression on the scaling trend for large language models.
We have shown that the optimal ratio between training data bytes and model parameters, denoted as $\rho^\star$, remains approximately constant across varying compute budgets and compression rates.
Consequently, when generalizing scaling recipes to models with different tokenizers, we advise matching the ratio of training bytes (not tokens) to model parameters.
Additionally, we find the optimal compression rate $T^\star$ that is specific to the training domain and slowly decreases with the training budget.
Finally, we show that these scaling trends with \compr{} hold consistently for both latent and subword-tokenized models.

\section*{Acknowledgments}

We thank Jonathan Hayase, Julie Kallini, Pedro Rodriguez, and Rylan Schaeffer for insightful discussions that helped shape and improve this work.
We are grateful to Artyom Kozhevnikov, David Dale, and Marta R. Costa-jussà for their advice and practical assistance with multilingual experiments.
Finally, we express special gratitude to Cody Ohlsen for going to great lengths to resolve technical obstacles at one of the project’s most critical moments.

%% file: sections/appendices.tex
\section{Model Scaling: Technical Details}
\label{sec:model-scaling-details}

In this section, we describe the model architectures in detail.

The core experiments were conducted with BLT models \citep{pagnoni-etal-2025-byte}.
We followed the original implementation with a few notable exceptions.
We find that the local modules should be wide (high number of heads) and shallow (low number of layers).
To achieve such a shape, we set the number of layers in each local module to the ceiling of one-fourth of the number of global layers, and the local head count to the ceiling of one-fourth of the number of global heads, plus 8.
The cross-attention key-query duplication factor $k$ is set to the ceiling of one-eighth of the global module's head count.
The hidden dimension of the local modules is set to 64 times the number of heads.
This scaling recipe ensures that the compute overhead introduced by the local modules is comparable to the embedding layers found in isotropic models of similar scale.

An important divergence from the original BLT architecture of \citet{pagnoni-etal-2025-byte} is the omission of hash embeddings.
To compensate for the reduced capacity for encoding the input, we increase the number of layers in the encoder to match the decoder (originally, the encoder has only one layer).
Table~\ref{tab:blt_scales} presents the scales and architecture hyperparameters of all BLT models used in this work.

Similarly, Table~\ref{tab:isotropic_scales} outlines the scaling recipe for subword tokenized models.

We compare the compute spend in the latent module as a percentage of the total inference compute for both families of models in Table~\ref{tab:scales_flops_share}.
We observe that with our scaling recipes, the global module takes up a similar share of compute in the BLT architecture as in the isotropic model when the model scale and \compr{} are matched.
We observe that decreasing \compr{} or increasing model size correlates with a higher relative utilization of the global model.

\input{tables/BLT_scales}

\input{tables/isotropic_scales}

\input{tables/scales_flops_share}

\section{Scaling Laws: Technical Details}
\label{sec:power-law-details}

We characterize compute-optimal scaling through a two-stage fitting procedure.

\subsection{Scaling Law I}

We fit the scaling laws to find optimal data and parameters as described in Equations~\ref{eq:optimal_data}~and~\ref{eq:optimal_param}.
As noted in the methodology, we restrict this fit to the parameters of the global latent model (excluding encoder/decoder and embeddings) to ensure consistency across tokenization methods.

We perform the fit using the L-BFGS-B  \citep{zhu1997bfgsb} algorithm with a gradient tolerance of $10^{-10}$.
To ensure robust convergence, we employ a grid search for initialization:
\begin{enumerate}
    \item We first compute an Ordinary Least Squares (OLS) solution ($\alpha_\text{OLS}, \beta_\text{OLS}, B_\text{OLS}, N_\text{OLS}$) on the log-transformed data to serve as a prior.
    \item We define a search grid by perturbing the OLS solution. We test 13 values for each parameter, resulting in $13^4$ total initialization points (though we fix $\alpha$ and $\beta$ ranges tighter than $B_0$ and $N_0$).
\end{enumerate}

The grid is constructed as follows:
\begin{itemize}
    \centering
    \item $\log(B_\text{init}) \in \{\log(B_\text{OLS}) + \epsilon : \epsilon \in [-3, 3]\}$
    \item $\log(N_\text{init}) \in \{\log(N_\text{OLS}) + \epsilon : \epsilon \in [-3, 3]\}$
    \item $\alpha_\text{init} \in \{\alpha_\text{OLS} + \epsilon : \epsilon \in [-0.3, 0.3]\}$
    \item $\beta_\text{init} \in \{\beta_\text{OLS} + \epsilon : \epsilon \in [-0.3, 0.3]\}$
\end{itemize}
We select the solution that minimizes the sum of squares loss objective.
The BFGS algorithm obtained the parameter values similar to OLS regardless of its starting point.

\subsection{Scaling Law II}

In the second stage, we fit the power law for optimal loss $L^\star(C,T) \simeq L_0 C^\gamma + f(T)$.
Unlike Stage I, we use the total compute budget, for this fit, including the cost of the encoder, decoder, and embeddings.

We fit the parameters $L_0$, $\gamma$, and the compression-specific offsets $f(T)$ simultaneously.
We again use BFGS with a grid search for initialization.
The grid spans 13 values for $L_0$ and $\gamma$ (169 combinations):
\begin{itemize}
    \centering
    \item $\log(L_\text{init}) \in [-3, 3]$
    \item $\gamma_\text{init} \in [-0.6, 0.0]$
\end{itemize}
The initial value for $f(T)$ is set to the mean loss observed at compression rate $T$.
During optimization, we bound the parameters to physically plausible ranges: $\log(L_0) \in [-30, 30]$, $\gamma \in [-2, 0]$, and $f(T) \in [-5, 5]$.

\subsection{Derivation and Validation of Scaling Law II}
\label{sec:scaling_law_2_derivation}

\input{tables/scaling_law_validation}

As described in Section~\ref{sec:scaling_law_2}, we begin the search for the scaling law equation by assuming the classical form from \citet{kaplan2020scaling}, disregarding the role of compression.
It is presented by Equation~\ref{eq:optimal_loss}:
\begin{equation*}
    L^\star(C,T) \simeq L_0 \times C^\gamma + f(C,T)
\end{equation*}

First, we fit the first part of the scaling law, and then we examine the functions that would give the best approximation of $f(C,T)$ (residuals of the fit) with minimal complexity.
We consider the following candidates for $f(C,T)$:

\paragraph{Mean of the residuals} is equivalent to the ``irreducible loss'' term or intercept used in many scaling fits.
It is the simplest form of $f(C,T)$, yet it still completely disregards the role of compression on loss.
We consider the following form of irreducible loss:
\begin{equation}
    \label{eq:optimal_loss_irreducible}
    f(C,T) = E
\end{equation}

\paragraph{Constant optimal compression ($T^\star$)} is an assumption that the loss is always minimal for one compression rate, regardless of compute budget $C$. By an inspection of $f(C,T)$ residuals in Figure~\ref{fig:cr_vs_loss_blt}, we observe that they are distributed along a non-monotonic convex function of $T$, with a minimum at some point $T_0$.
We assume that a quadratic function fits this relation well.
Considering that $T$ is on a logarithmic scale, we propose the following equation for residuals (including also irreducible loss):
\begin{equation}
    \label{eq:optimal_const_compression}
    f(C,T) = F \times \left(\log(T) - \log(T_0)\right)^2 +E = F \times \log^2\left(\frac{T}{T_0}\right) + E
\end{equation}

\paragraph{Compute-dependent optimal compression ($T^\star$)} is based on a hypothesis that the optimal compression depends on compute budget.
We observe that the minimum of the quadratic function modeling $f(T,C)$ described in the last paragraph shifts to a lower value with an increase of the training budget.
To account for that, we include the effect of the compute budget in the log-quadratic function, arriving at the following formulation of Equation~\ref{eq:optimal_compression_rate}:
\begin{equation*}
    f(C,T) = F \times \log^2\left(\frac{C^\delta T}{T_0} \right) + E
\end{equation*}

To validate the extrapolation accuracy of the three candidate formulas, we fitted scaling laws for results of models trained for 8 computation budgets from $5 \times 10^{18}$ to $1 \times 10^{21}$, and 6 compression rates.
For each compression rate and budget, we use the optimal model size (and training data) estimated by the Scaling Law I.
Then we validate the obtained scaling laws by comparing expected vs.\ obtained loss for models trained with a higher compute budget: $2 \times 10^{21}$.
In Table~\ref{tab:scaling_law_validation}, we observe that the last formulation, making an assumption that the optimal \compr{} is compute-dependent, obtains significantly lower mean square error in extrapolation than other candidate formulations.
Moreover, the fit using this formula obtains the highest goodness-of-fit coefficient, both standard ($R^2$) and adjusted for the number of fitted variables ($\bar{R}^2$).
Therefore, we decided to choose this formulation for the final version of the scaling law.


\subsection{Loss Sensitivity to Compression Rate}
\label{sec:loss_sensitivity}

\begin{figure*}[!htb]
    \centering
    \begin{subfigure}{0.49\textwidth}
        \includegraphics[width=\linewidth]{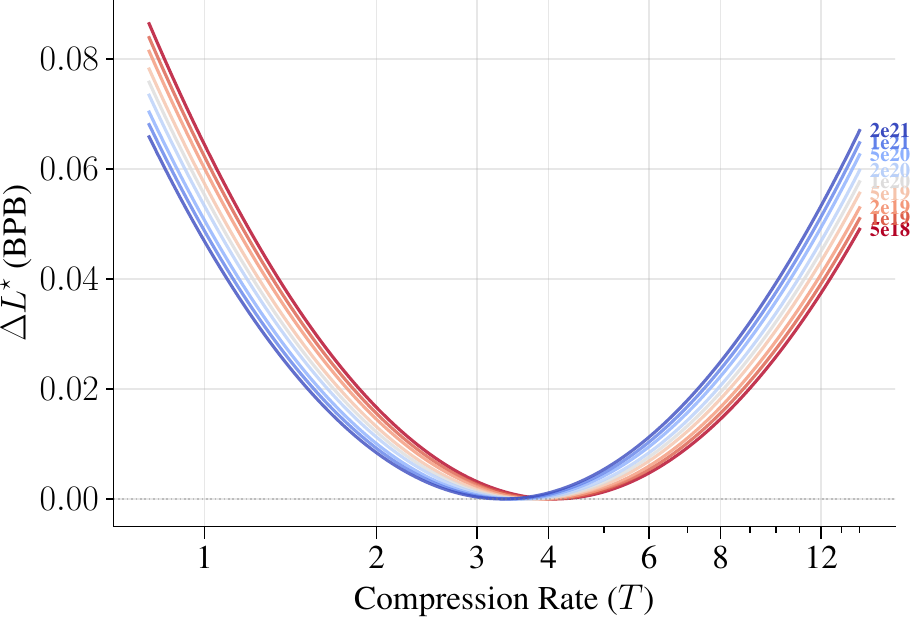}
        \caption{Latent Tokenization }
    \end{subfigure}
    \begin{subfigure}{0.49\textwidth}
        \includegraphics[width=\linewidth]{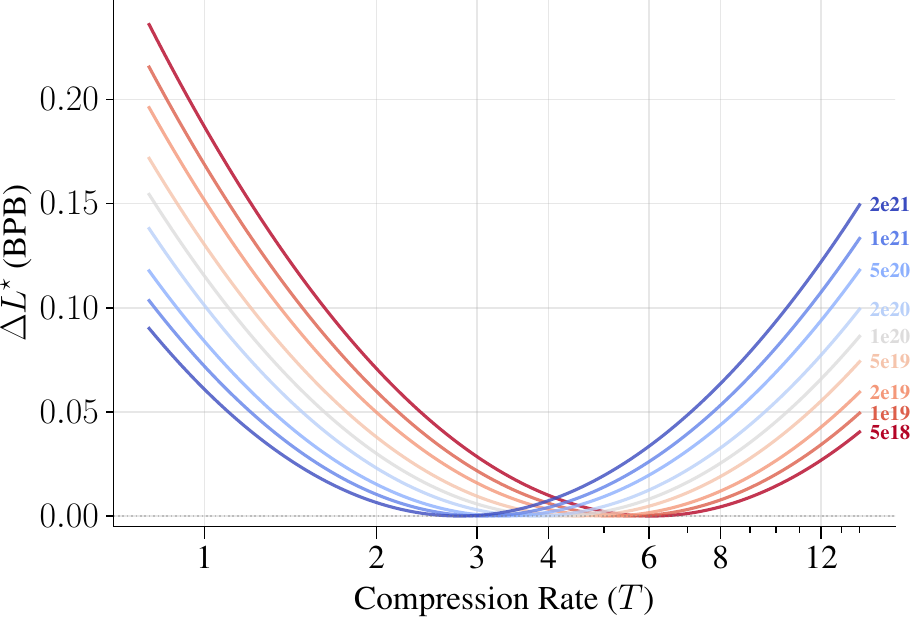}
        \caption{Subword Tokenization}
    \end{subfigure}
    \caption{The BPB deterioration across compression compared to the value at optimal \compr{}.
    $\Delta L ^\star$ function was predicted based on Scaling Law II fit.
    }
    \label{fig:loss_sensitivity}
\end{figure*}

Figure~\ref{fig:loss_sensitivity} shows marginal sensitivity of loss to the choice of compression rate.
We observe that \compr{} close to optimal has minimal impact on loss, yet diverging further from the optimum can cause up to $0.2$ and $0.1$ deterioration in test BPB for subword and latent tokenized models respectively. 

\subsection{Confidence Intervals}


We compute $95\%$ confidence intervals for the fitted parameters
$\hat{\boldsymbol{\theta}} \in \mathbb{R}^p$ from $n$ data points,
where $p$ is the number of parameters.
$\mathcal{L}(\boldsymbol{\theta})$ denotes the sum of squares loss
evaluated at $\boldsymbol{\theta}$, and  $\mathbf{e}_k$ be the $k$-th
standard basis vector in $\mathbb{R}^p$.

The Hessian $H \in \mathbb{R}^{p \times p}$ of $\mathcal{L}$ is estimated via central
finite differences with step size $\epsilon = 10^{-5}$:
\begin{equation}
  H_{ij}
  =
  \frac{
    \mathcal{L}_{ij}^{++} - \mathcal{L}_{ij}^{+-} - \mathcal{L}_{ij}^{-+} + \mathcal{L}_{ij}^{--}
  }{4\,\epsilon^{2}}
\end{equation}
where
\begin{equation}
  \mathcal{L}_{ij}^{s_1 s_2}
  =
  \mathcal{L}\!\bigl(\boldsymbol{\theta}
    + s_1\,\epsilon\,\mathbf{e}_i
    + s_2\,\epsilon\,\mathbf{e}_j\bigr)
  \qquad s_1, s_2 \in \{+, -\}
\end{equation}
The residual variance is estimated as
\begin{equation}
  \hat{\sigma}^{2}
  =
  \frac{\displaystyle\sum_{i=1}^{n} r_i^{2}}{n - p}
\end{equation}
where $r_i = y_i - \hat{y}_i$ is the $i$-th residual
(observed minus predicted value).
The parameter covariance matrix is
\begin{equation}
  \hat{\Sigma} = 2 \hat{\sigma}^{2}\,H^{-1}
\end{equation}
The $95\%$ confidence interval for each parameter $\hat{\boldsymbol{\theta}}_{k}$ is:
\begin{equation}
  \hat{\boldsymbol{\theta}}_{k} \pm t \cdot \sqrt{\hat{\Sigma}_{kk}}
\end{equation}
where $\sqrt{\hat{\Sigma}_{kk}}$ is the standard error for estimation of $\hat{\boldsymbol{\theta}}_{k}$ and $t$ is the two-sided $95\%$ critical value of Student's
$t$-distribution with $n - p$ degrees of freedom.

\clearpage

\section{Impact of Tokenization Method}
\label{sec:tokenization_method_comparison}

\begin{figure*}[!htb]
    \centering
    \begin{subfigure}{0.32\textwidth}
        \includegraphics[width=\linewidth]{figures/blt_entropy_1e20_results_compression_contour_global_parameters.pdf}
        \caption{Latent Tokenization (Entropy)}
    \end{subfigure}
    \begin{subfigure}{0.32\textwidth}
        \includegraphics[width=\linewidth]{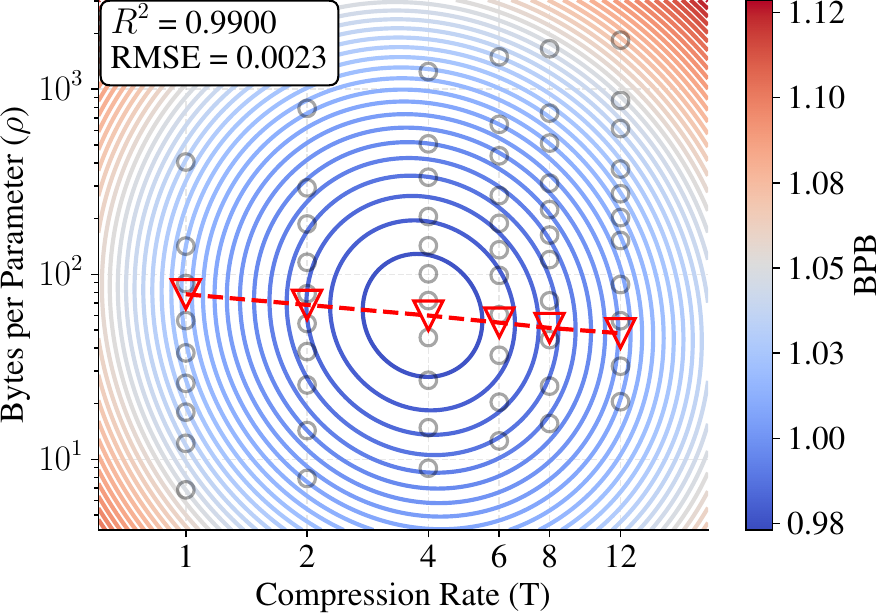}
        \caption{Latent Tokenization (Fixed Size)}
    \end{subfigure}
    \begin{subfigure}{0.32\textwidth}
        \includegraphics[width=\linewidth]{figures/isotropic_1e20_results_compression_contour_global_parameters.pdf}
        \caption{Subword Tokenization }
    \end{subfigure}
    \caption{Comparison of three-dimensional IsoFLOPs 
    ($C=10^{20}$) for three methods of tokenization: Latent Entropy, Latent Fixed Size (each latent token has fixed size of $T$ bytes), and Subword.
    The loss profile is visibly similar across the methods, with optimal loss achieved along constant \bpp{}.}
    \label{fig:stage_1_comparison_1e20}
\end{figure*}

\input{tables/tokenization_method_comparison}

We compare our results for different methods of tokenization: latent (with entropy supervision) and subword, as described in the main text.
Moreover, we compare the results with another method of latent tokenization, where all latent tokens are of the same fixed size in bytes equal to \compr{}.
In Figure~\ref{fig:stage_1_comparison_1e20}, we see similar loss profiles across different methods.
For all the methods and \compr{}s, the optimal configurations fall at $\approx 60$ bytes per parameter ratio $(\rho)$.
In Table~\ref{tab:tokenization_method_comparison}, we further observe that for two latent tokenization methods the optimal \compr{} is similar, while in subword tokenization it is higher.
This is due to an imperfect IsoFLOP paraboloid fit caused by poor performance of character-level models ($T=1.01$) under the considered budget, skewing optimal $T$ to be higher than in reality.
Based on the more reliable Scaling Law II estimation (see Section~\ref{sec:subword_scaling}) we expect to observe lower optimal \compr{} for this budget: $T^\star=4.11$.
For comparison, optimal \compr{} for latent models based on Scaling Law is $T^\star=3.67$.

\clearpage

\section{Impact of Mixing Languages}
\label{sec:beyond_english_mix}

\input{tables/xlingual_mix}

\begin{figure}[ht]
\centering
    \begin{subfigure}[t]{0.32\textwidth}
    \centering
    \includegraphics[width=\linewidth]{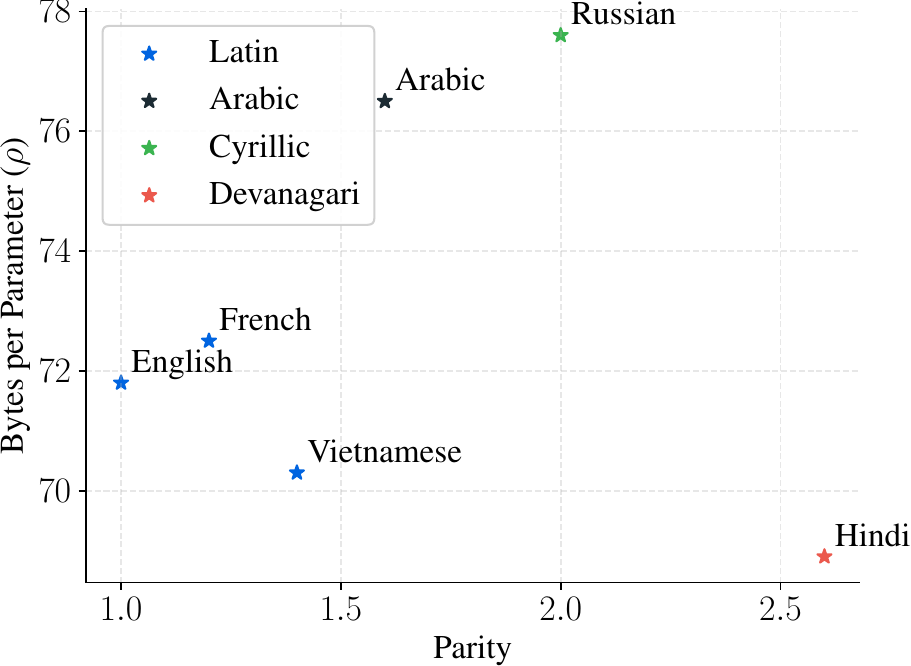}
    \caption{\bpp{}}
    \label{fig:mix_xlingual_bpp_1e20}
    \end{subfigure}
\hfill
    \begin{subfigure}[t]{0.32\textwidth}
    \includegraphics[width=\linewidth]{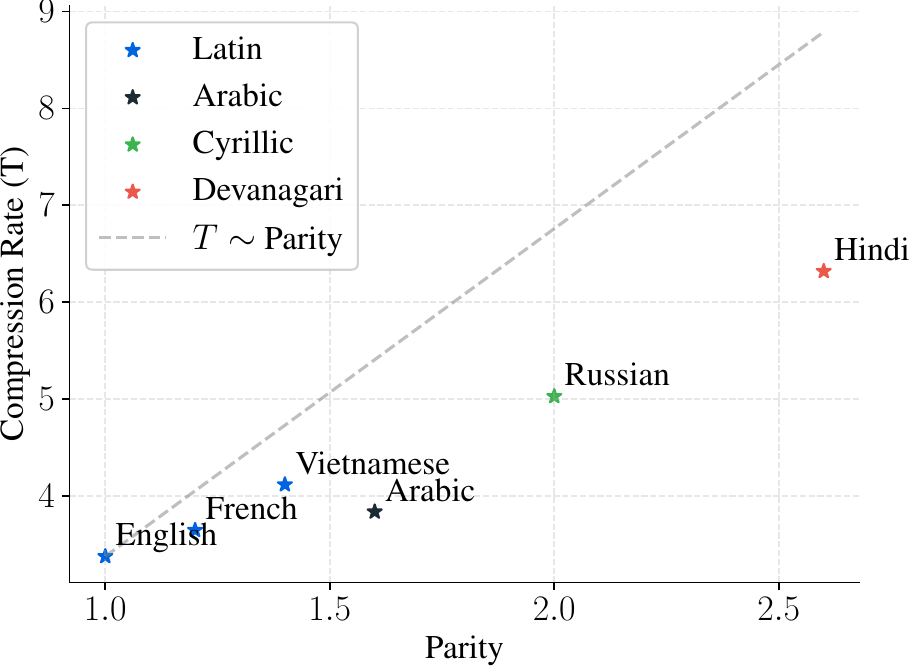}
    \caption{\compr{}}
    \label{fig:mix_xlingual_crs_1e20}
    \end{subfigure}
\hfill
    \begin{subfigure}[t]{0.32\textwidth}
    \centering
    \includegraphics[width=\linewidth]{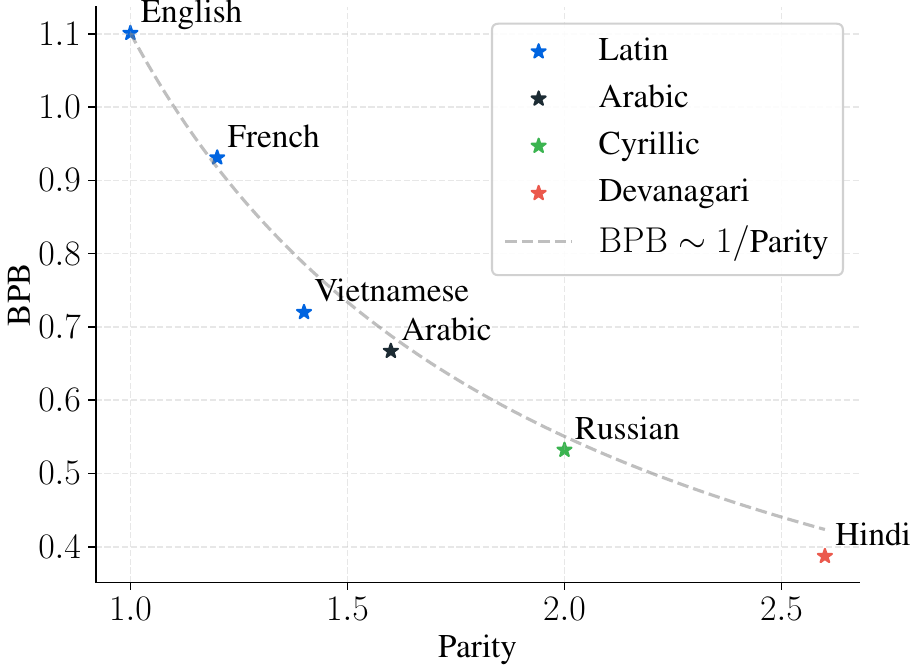}
    \caption{BPB}
    \label{fig:mix_xlingual_bpbs_1e20}
    \end{subfigure}
\caption{The optimal \bpp{}, \compr{}, and BPB for multilingual models trained on all languages jointly for $C=10^{20}$ FLOPs. The optimal values are compared against \pari{} on x-axis.}    
\label{fig:mix_xlingual_1e20}
\end{figure}

To examine the impact of mixing languages during training, we train a set of models jointly on multilingual data in six languages (including English), described in Section~\ref{sec:beyond_english}.
To enforce an equitable training signal across languages, we sample languages with weights equal to their \pari{}.
For instance, we train on $2.6$ more bytes in Hindi than in English, but we expect the two samples to be matched in information value.
All training runs are constrained to a fixed budget of $C = 10^{20}$ FLOPs; thus, multilingual models see less in-language data per language than their monolingual counterparts.

The optimal values of \bpp{} and \compr{} for each language are computed based on fits to the in-language test set, the results are gathered in Table~\ref{tab:xlingual_mix_optimals_1e20}.
Figure~\ref{fig:mix_xlingual_bpp_1e20} shows that the optimal \bpp{} is similar across languages. 
This contrasts with the findings in Section~\ref{sec:beyond_english}, where the optimal \bpp{} was language-dependent and correlated with \pari{}.
Notably, the multilingual optimal \bpp{} ($\rho^\star \approx 70$) is close to the median of the language-specific optimal values, $\rho^\star_l$.
As in the monolingual experiments, we observe that the optimal \compr{} (Figure~\ref{fig:mix_xlingual_crs_1e20}) is correlated with \pari{}. 
The multilingual optimal values are lower than the corresponding monolingual ones.
Test BPB (Figure~\ref{fig:mix_xlingual_bpbs_1e20}) is inversely correlated with \pari{}, in line with the results of Section~\ref{sec:beyond_english}.
As expected, multilingual models perform worse than monolingual ones due to the smaller amount of in-language data.

\clearpage

\section{Comparison with ``Scaling Laws with Vocabulary''}
\label{sec:comparison_tao_et_al}

\citet{tao2024scaling} posited similar research questions to ours regarding the role of tokenization in scaling laws, yet reached significantly different conclusions, showing that vocabularies (and thus \compr{}) should increase with model scale.
Meanwhile, we observe that the compute optimal \compr{} does not increase with model scale.
We identify the following methodological differences that explain discrepancies:

\paragraph{Approach to embedding-layer compute and vocabulary size}
The main difference is how compression is connected to the size of the embedding layers.
\citet{tao2024scaling} control \compr{} by changing the vocabulary size, which affects the size of the embedding layer.
This leads to a preference for smaller vocabularies at low compute and parameter budgets, so the FLOPs saved in embedding layers can be used for significantly longer training.
In our experiments, vocabulary cost is (almost) the same regardless of compression, thanks to the use of BLT \citep{pagnoni-etal-2025-byte} or alternative subword methods such as SuperBPE \citep{liu2025superbpespacetravellanguage}.
Therefore, our results extrapolate better to larger scales, where the cost of the embedding layer is negligible, as seen in Table~\ref{tab:scales_flops_share}.

\paragraph{Considered compression range}
BPE achieves a narrow \compr{} range (by our estimates, $T\in[3,4.5]$ bytes per token).
Considering only compressions attainable by BPE allows us to observe only a portion of the loss profile, one that falls below the optimal compression value.

\paragraph{Evaluation}
Both works use normalized negative log likelihood enabling a fair comparison across tokenizers.
\citet{tao2024scaling} match validation context length in tokens, so the number of bytes in an evaluation example varies with vocabulary.
We match the number of bytes across compression levels (e.g., if with \compr{} $T=4$ we evaluate on 2048 tokens per example, then with \compr{} $T=8$ we evaluate on 1024 tokens).
Because early bytes are harder than later bytes, matching validation context in tokens can favor higher-compression tokenizers (more “late” bytes in an example).
This could explain why we do not see the same preference for high compression (large vocabulary) at larger scales.
For further reference, SuperBPE \citep{liu2025superbpespacetravellanguage} also matched evaluation context in bytes.
Similarly to our results, they observed worse BPB scores for highly compressed SuperBPE compared to regular BPE.

\clearpage

\section{Supplementary Results}

\subsection{IsoFLOP Analysis across Compute Budgets}
\label{sec:isoflop-across-budgets}

\begin{figure}[!htb]
    \centering
    \begin{subfigure}{0.95\textwidth}
        \includegraphics[width=\linewidth]{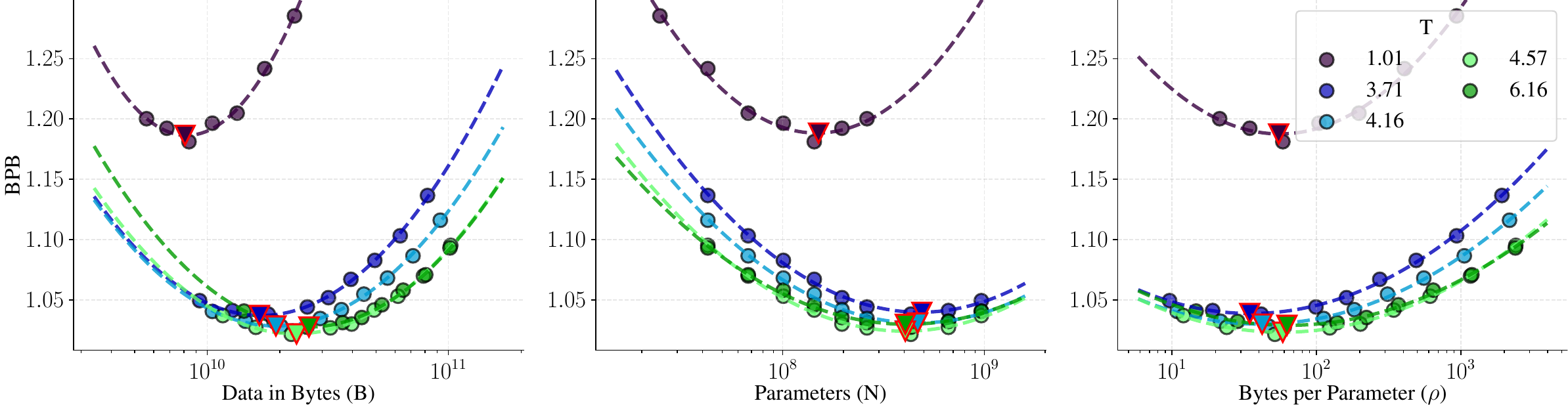}
        \caption{$C=5\times10^{19}$ Subword Tokenization}
    \end{subfigure}
    \begin{subfigure}{0.95\textwidth}
        \includegraphics[width=\linewidth]{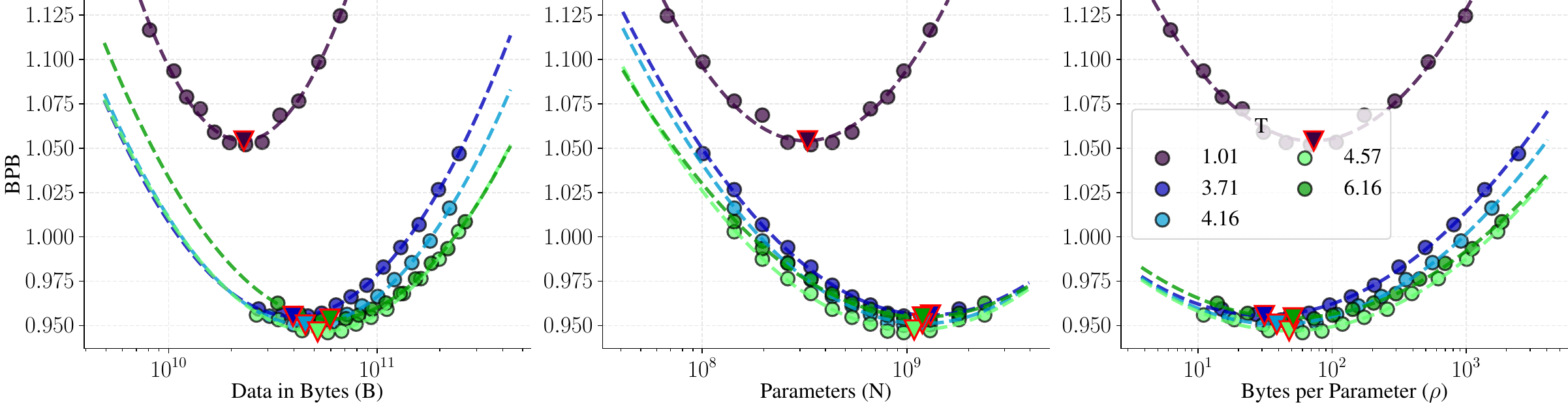}
        \caption{$C=10^{20}$ Subword Tokenization}
    \end{subfigure}
    \begin{subfigure}{0.95\textwidth}
        \includegraphics[width=\linewidth]{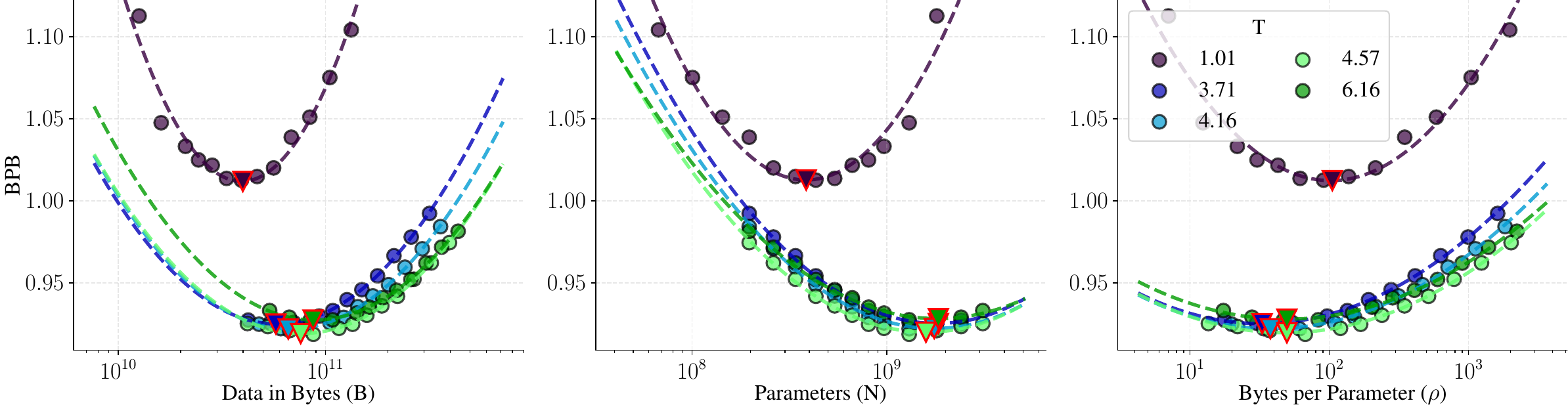}
        \caption{$C=2\times10^{20}$ Subword Tokenization}
    \end{subfigure}
    \caption{2-dimensional IsoFLOPs for subword tokenized models, as a function of data ($B$), parameters ($N$), or \bpp{} ratio ($\rho$).
    Training budgets are indicated in each panel's caption.
    IsoFLOPs (parabolas) are fitted for each compression line to interpolate values of the loss.
    }
    \label{fig:1d_isoflops_iso}
\end{figure}

We present the IsoFLOPs across multiple compute budgets and compression rates for latent tokenized models in Figures~\ref{fig:1d_isoflops_blt}.
And for subword tokenized models in Figures~\ref{fig:1d_isoflops_iso}.
We observe that the optimal byte-per-parameter ratio $\rho^\star$ remains constant for most of the considered configurations, this trend is more visible for $C > 10^{20}$, where the compute of the global module becomes dominant, thus it is expected to hold also at larger scales.

\begin{figure}[!htb]
    \centering
    \begin{subfigure}{0.95\textwidth}
        \includegraphics[width=\linewidth]{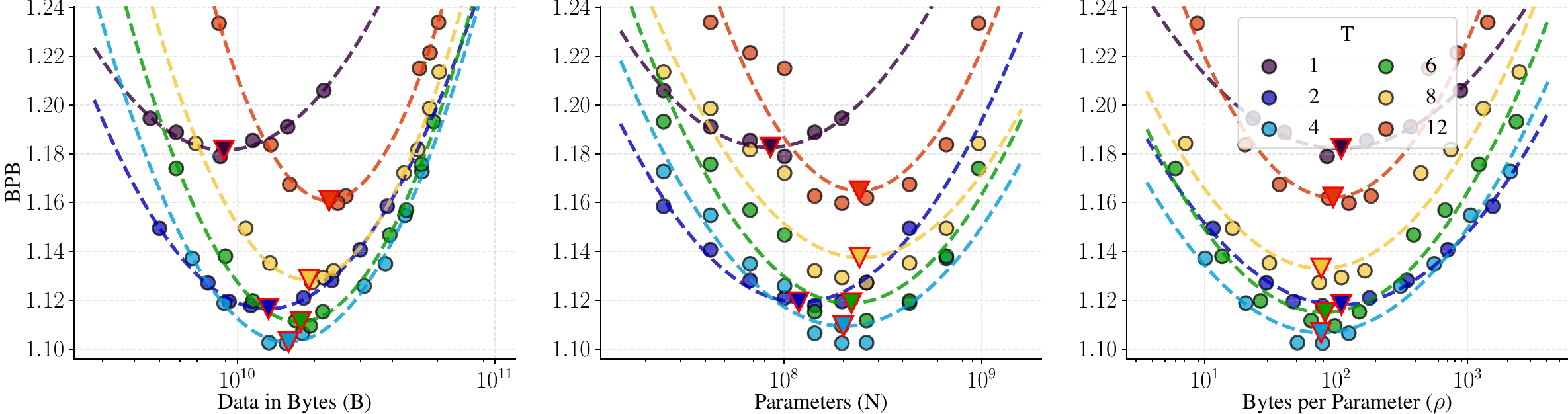}
        \caption{$C=10^{19}$ Latent Tokenization}
    \end{subfigure}
    \begin{subfigure}{0.95\textwidth}
        \includegraphics[width=\linewidth]{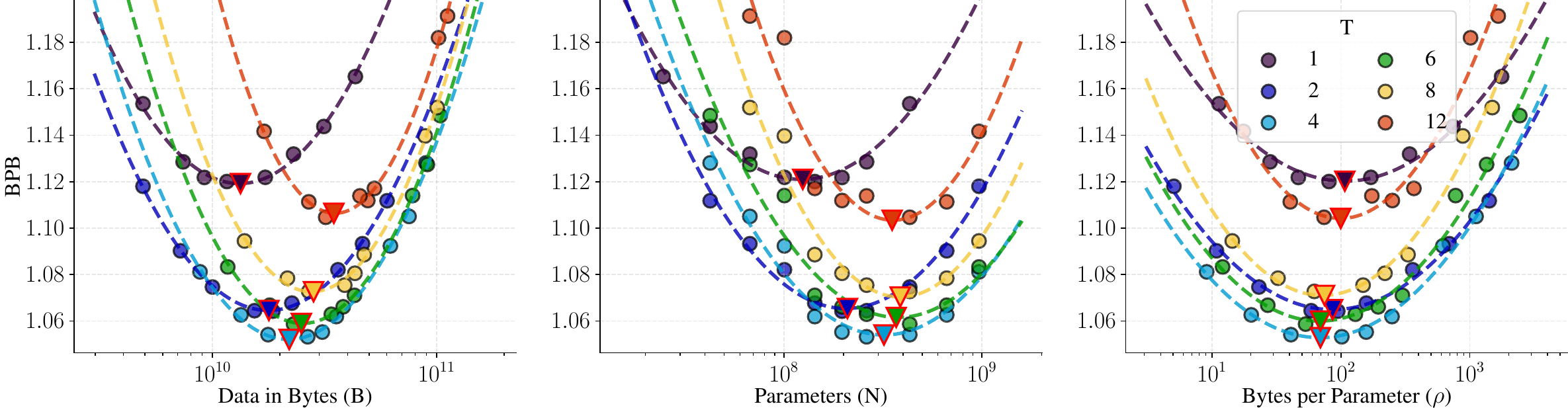}
        \caption{$C=2\times10^{19}$ Latent Tokenization}
    \end{subfigure}
    \begin{subfigure}{0.95\textwidth}
        \includegraphics[width=\linewidth]{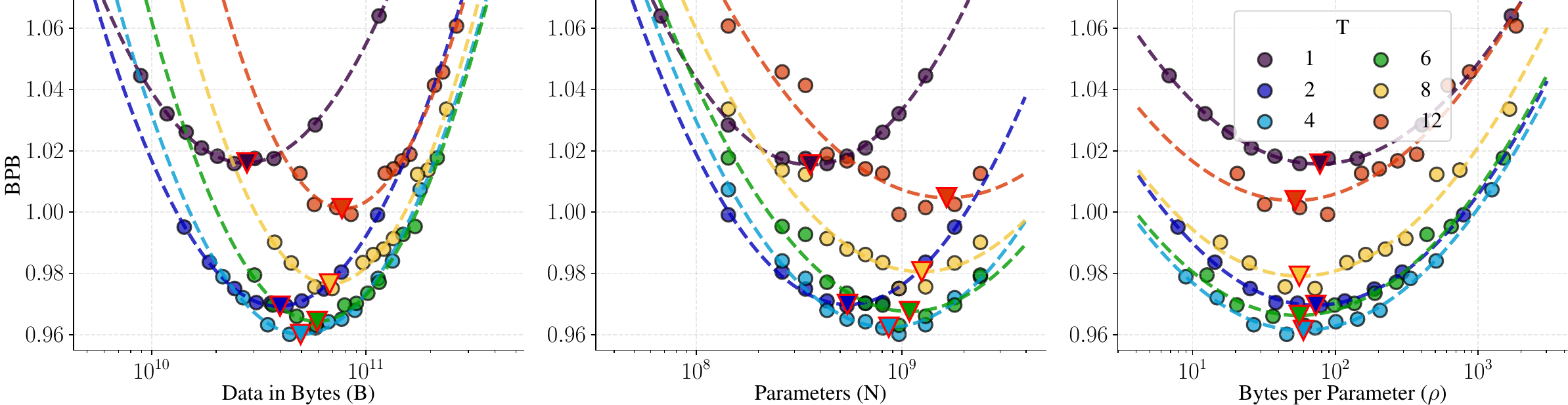}
        \caption{$C=10^{20}$ Latent Tokenization}
    \end{subfigure}
\end{figure}
\begin{figure}[!htb]
    \ContinuedFloat
    \begin{subfigure}{0.95\textwidth}
        \includegraphics[width=\linewidth]{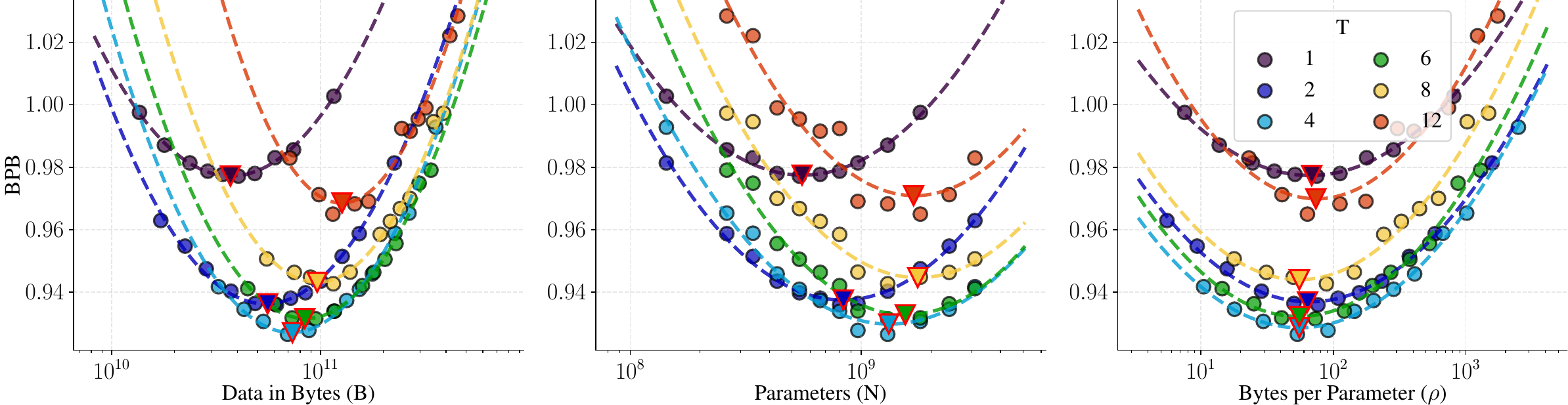}
        \caption{$C=2\times10^{20}$ Latent Tokenization}
    \end{subfigure}
    \begin{subfigure}{0.95\textwidth}
        \includegraphics[width=\linewidth]{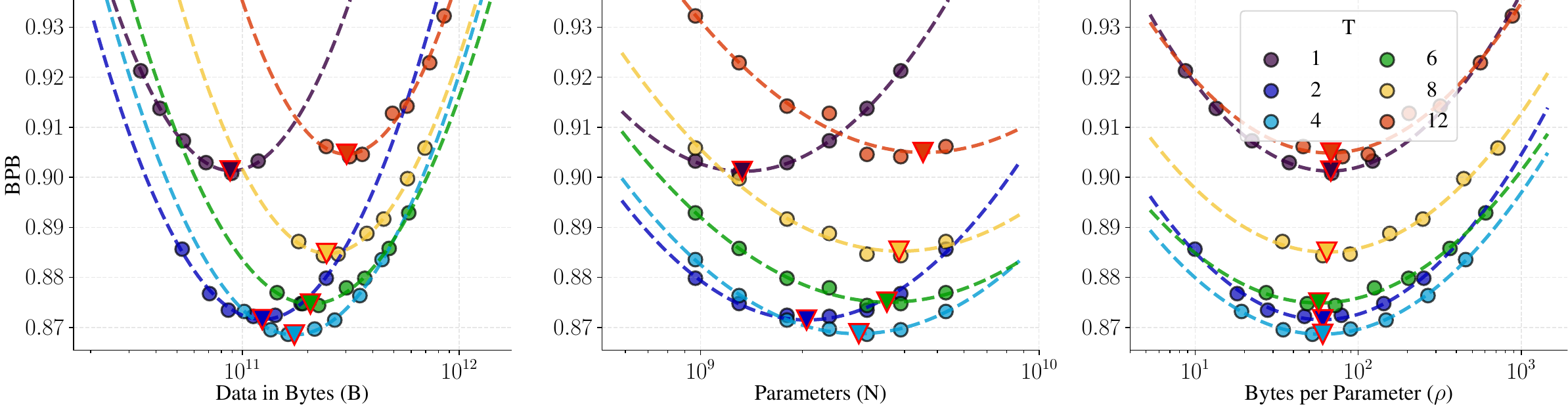}
        \caption{$C=10^{21}$ Latent Tokenization}
    \end{subfigure}
    \begin{subfigure}{0.95\textwidth}
        \includegraphics[width=\linewidth]{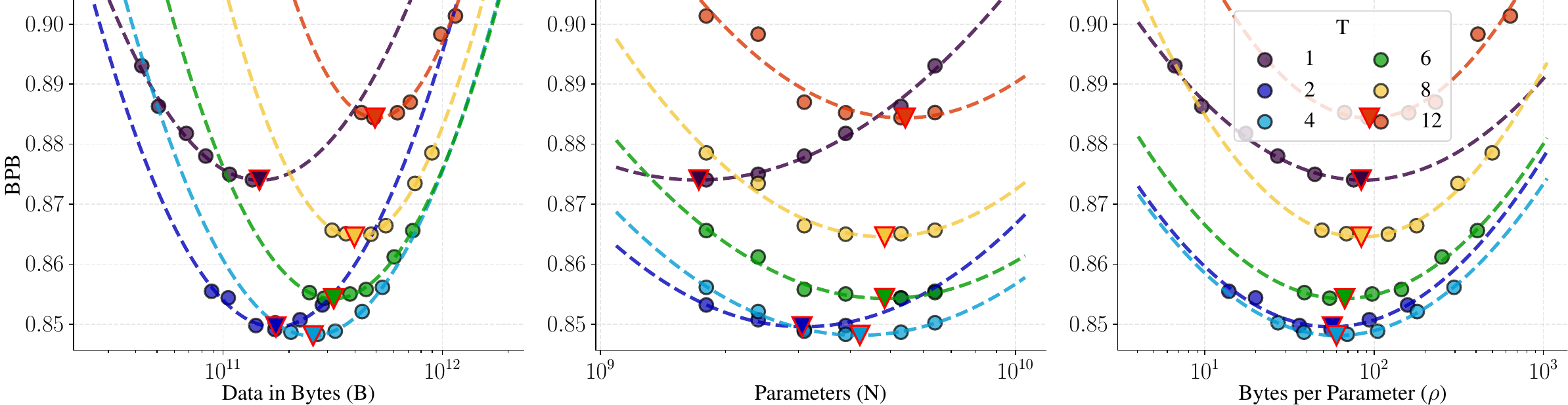}
        \caption{$C=2\times10^{21}$ Latent Tokenization}
    \end{subfigure}
    \caption{2-dimensional IsoFLOPs for latent tokenized models, as a function of data ($B$), parameters ($N$), or \bpp{} ratio ($\rho$).
    Training budgets are indicated in each panel's caption.
    IsoFLOPs (parabolas) are fitted for each compression line to interpolate values of the loss.
    }
    \label{fig:1d_isoflops_blt}
\end{figure}

\begin{figure}[!htb]
    \centering
    \begin{subfigure}{0.95\textwidth}
        \includegraphics[width=\linewidth]{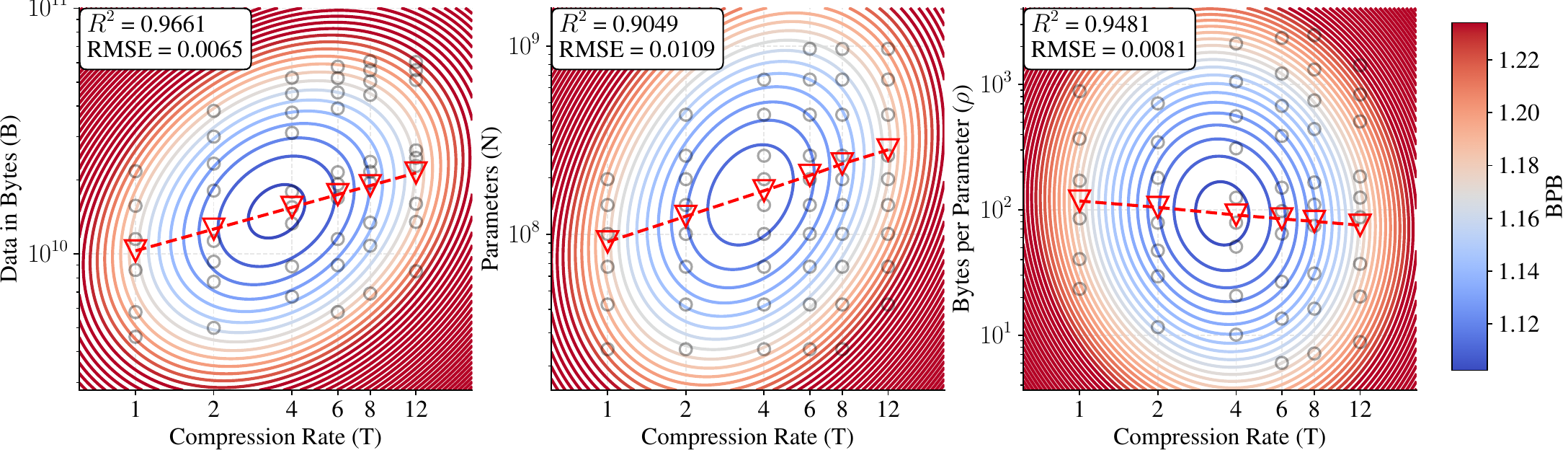}
        \caption{$C=10^{19}$ Latent Tokenization}
    \end{subfigure}
    \begin{subfigure}{0.95\textwidth}
        \includegraphics[width=\linewidth]{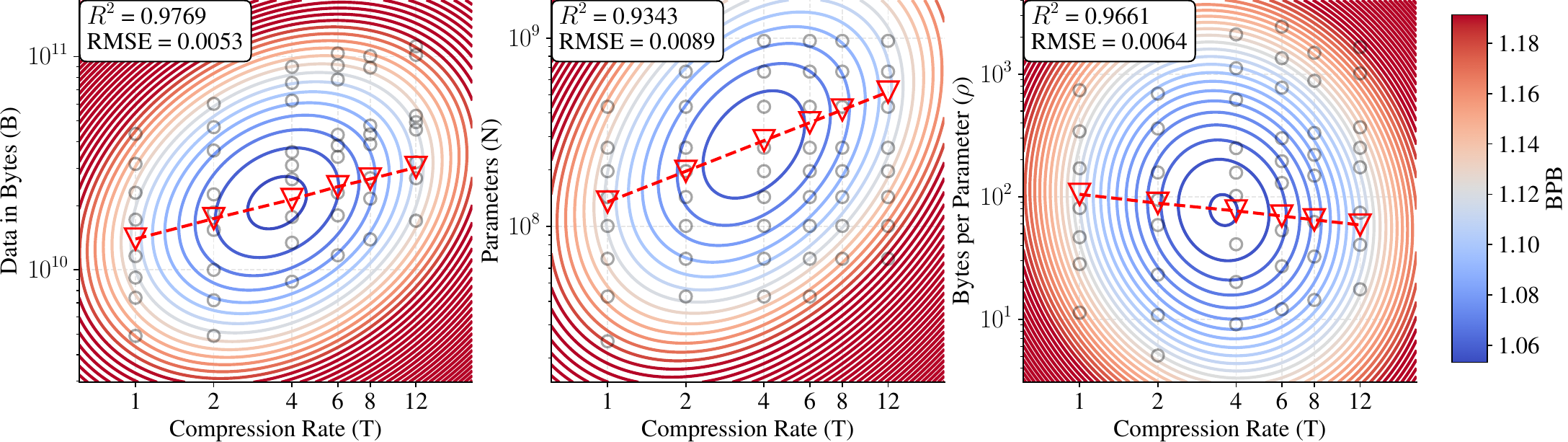}
        \caption{$C=2\times10^{19}$ Latent Tokenization}
    \end{subfigure}
        \begin{subfigure}{0.95\textwidth}
        \includegraphics[width=\linewidth]{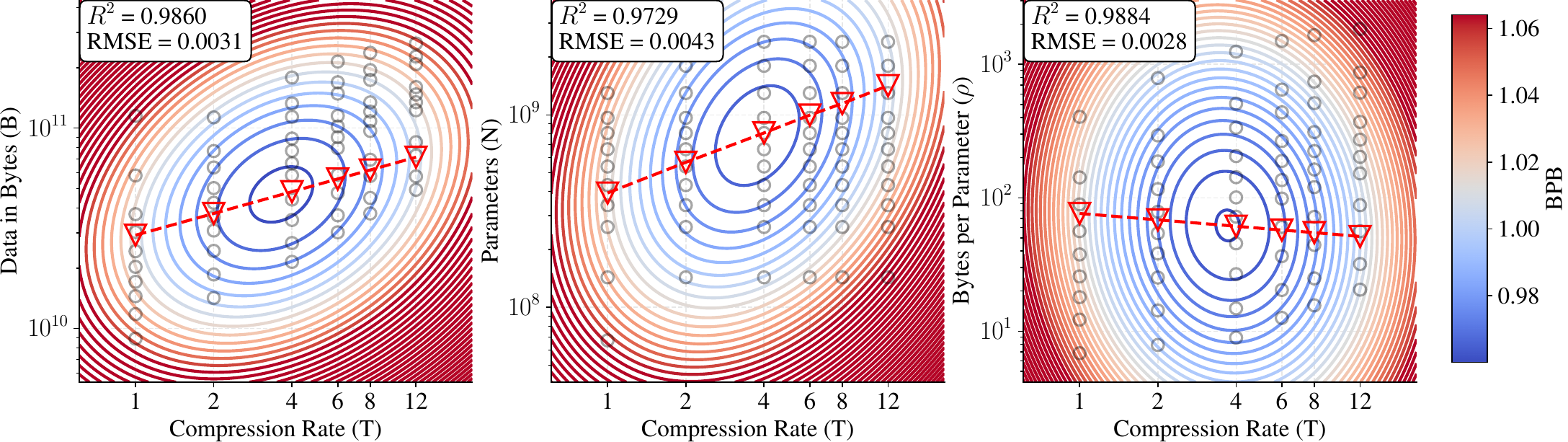}
        \caption{$C=10^{20}$ Latent Tokenization}
    \end{subfigure}
\end{figure}
\begin{figure}[!htb]
    \ContinuedFloat
    \begin{subfigure}{0.95\textwidth}
        \includegraphics[width=\linewidth]{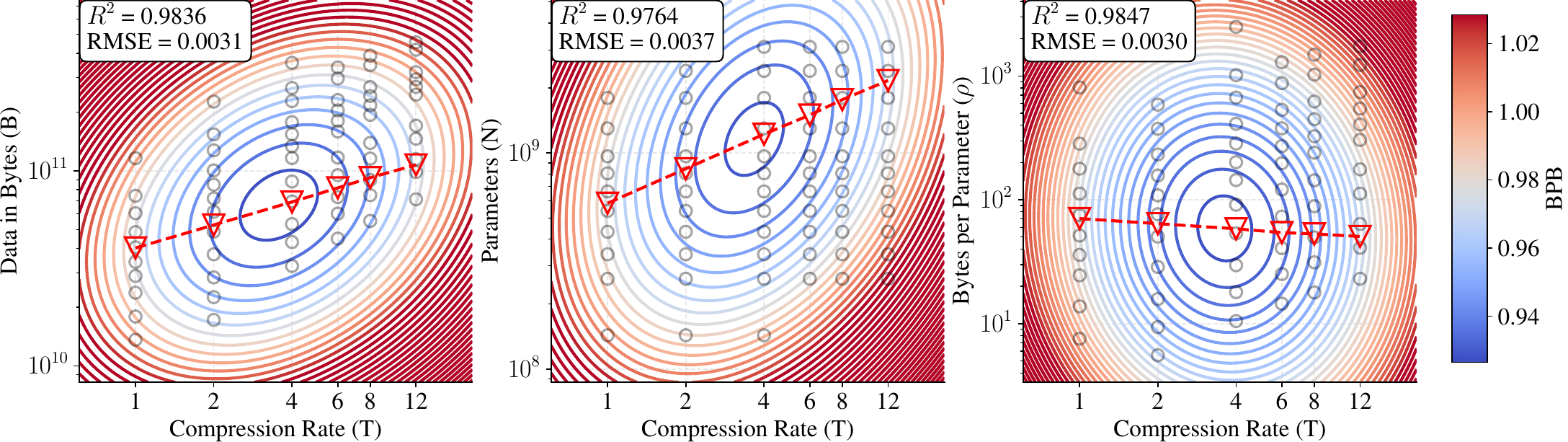}
        \caption{$C=2\times10^{20}$ Latent Tokenization}
    \end{subfigure}
    \begin{subfigure}{0.95\textwidth}
        \includegraphics[width=\linewidth]{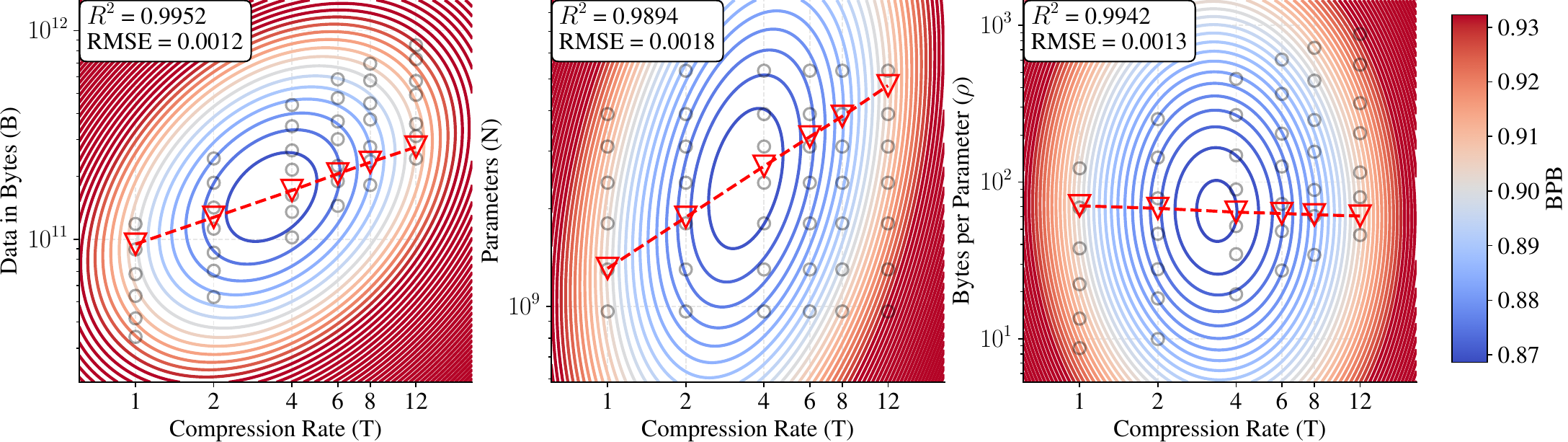}
        \caption{$C=10^{21}$ Latent Tokenization}
    \end{subfigure}
    \begin{subfigure}{0.95\textwidth}
        \includegraphics[width=\linewidth]{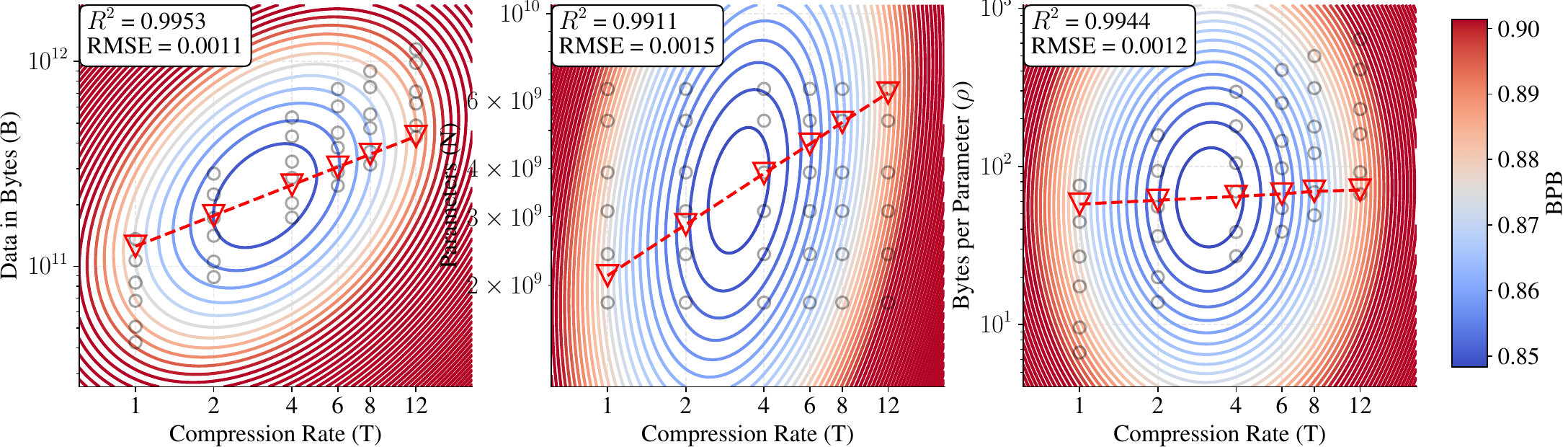}
        \caption{$C=2\times10^{21}$ Latent Tokenization}
    \end{subfigure}
    \caption{3-dimensional IsoFLOPs for latent tokenized models, as a function of \compr{} and data ($B$), parameters ($N$), or \bpp{} ratio ($\rho$).
    Training budgets are indicated in each figure's caption.
    IsoFLOPs (paraboloids) are jointly for all compression rates.
    }
    \label{fig:2d_isoflops_blt}
\end{figure}


\begin{figure}[!htb]
    \centering
    \begin{subfigure}{0.95\textwidth}
        \includegraphics[width=\linewidth]{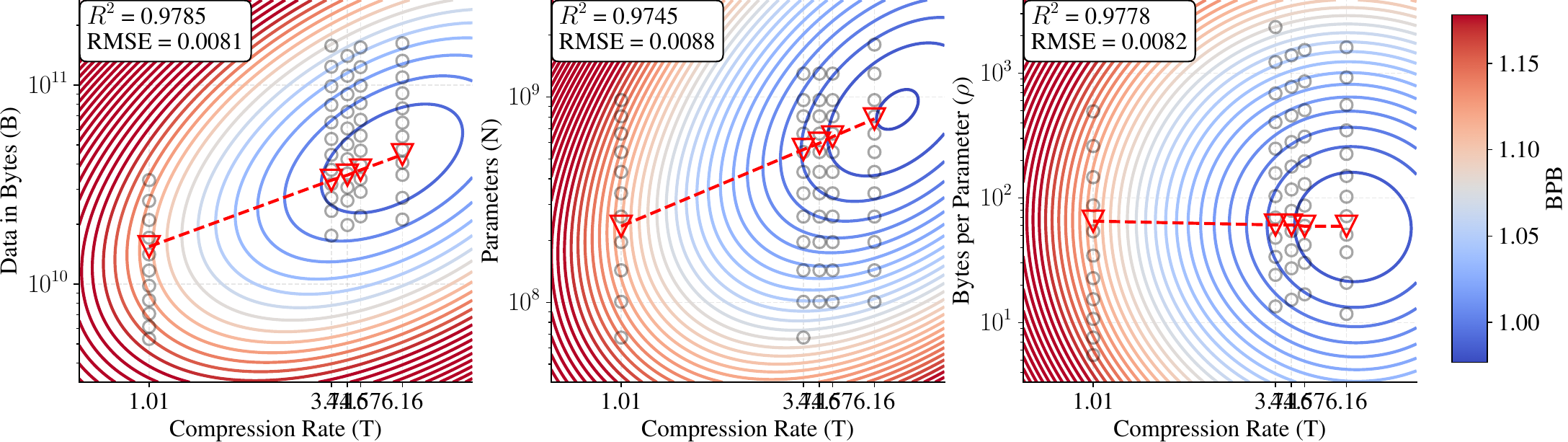}
        \caption{$C=5\times10^{19}$ Subword Tokenization}
    \end{subfigure}
    \begin{subfigure}{0.95\textwidth}
        \includegraphics[width=\linewidth]{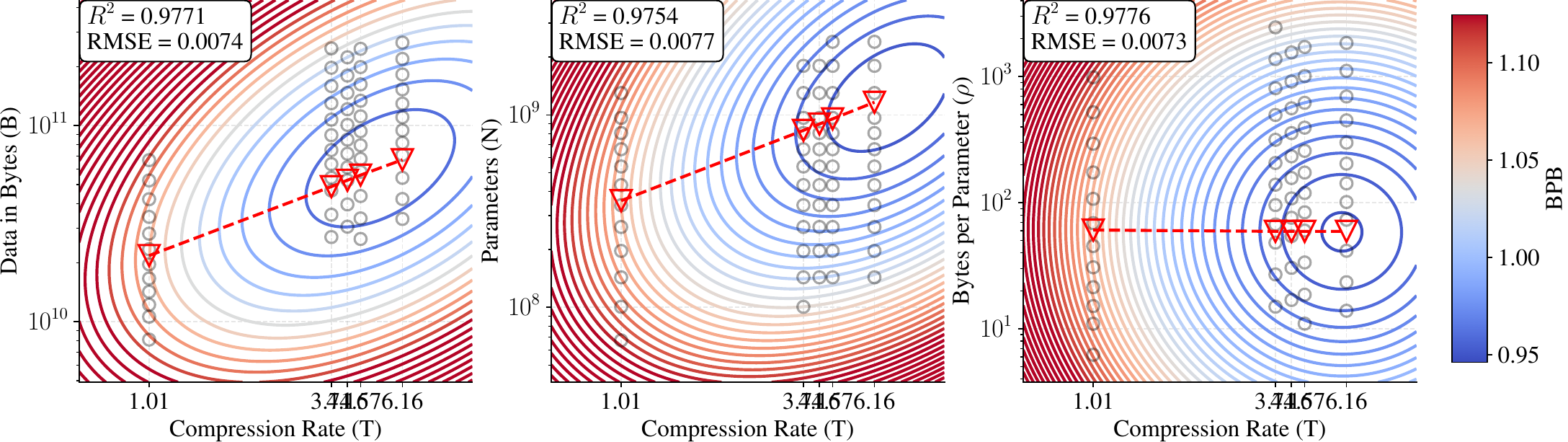}
        \caption{$C=10^{20}$ Subword Tokenization}
    \end{subfigure}
    \begin{subfigure}{0.95\textwidth}
        \includegraphics[width=\linewidth]{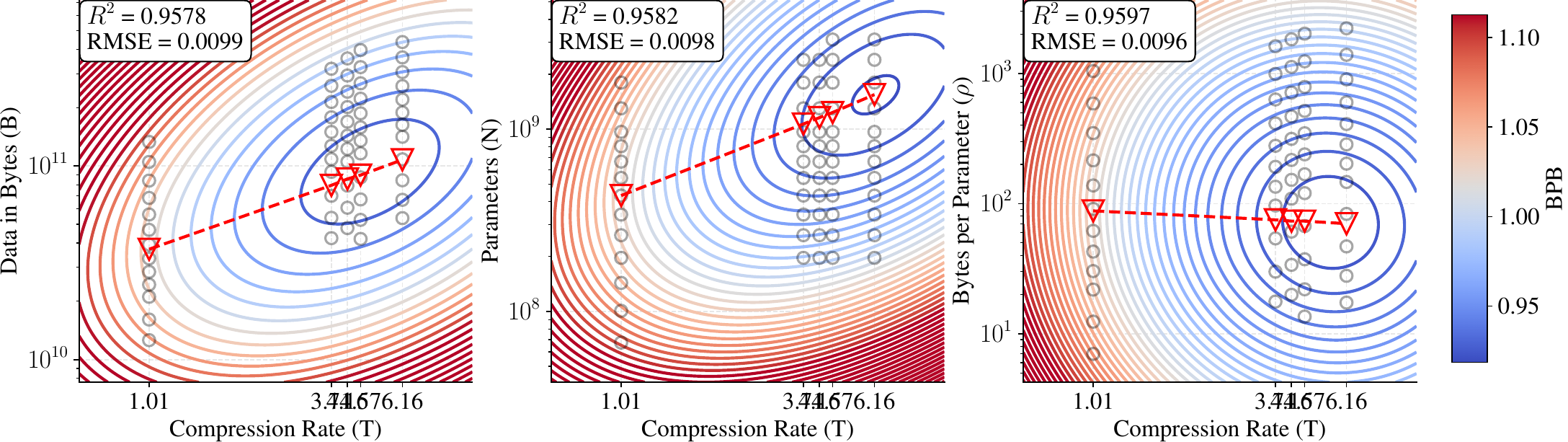}
        \caption{$C=2\times10^{20}$ Subword Tokenization}
    \end{subfigure}
    \caption{3-dimensional IsoFLOPs for subword tokenized models, as a function of \compr{} and data ($B$), parameters ($N$), or \bpp{} ratio ($\rho$).
    Training budgets are indicated in each figure's caption.
    IsoFLOPs (paraboloids) are jointly for all compression rates.
    }
    \label{fig:2d_isoflops_iso}
\end{figure}

\clearpage

\subsection{Optimal Data and Parameters across Compute Budgets}

\begin{figure}[!htb]
\centering
  \begin{subfigure}[t]{0.49\textwidth}
    \centering
    \includegraphics[width=\linewidth]{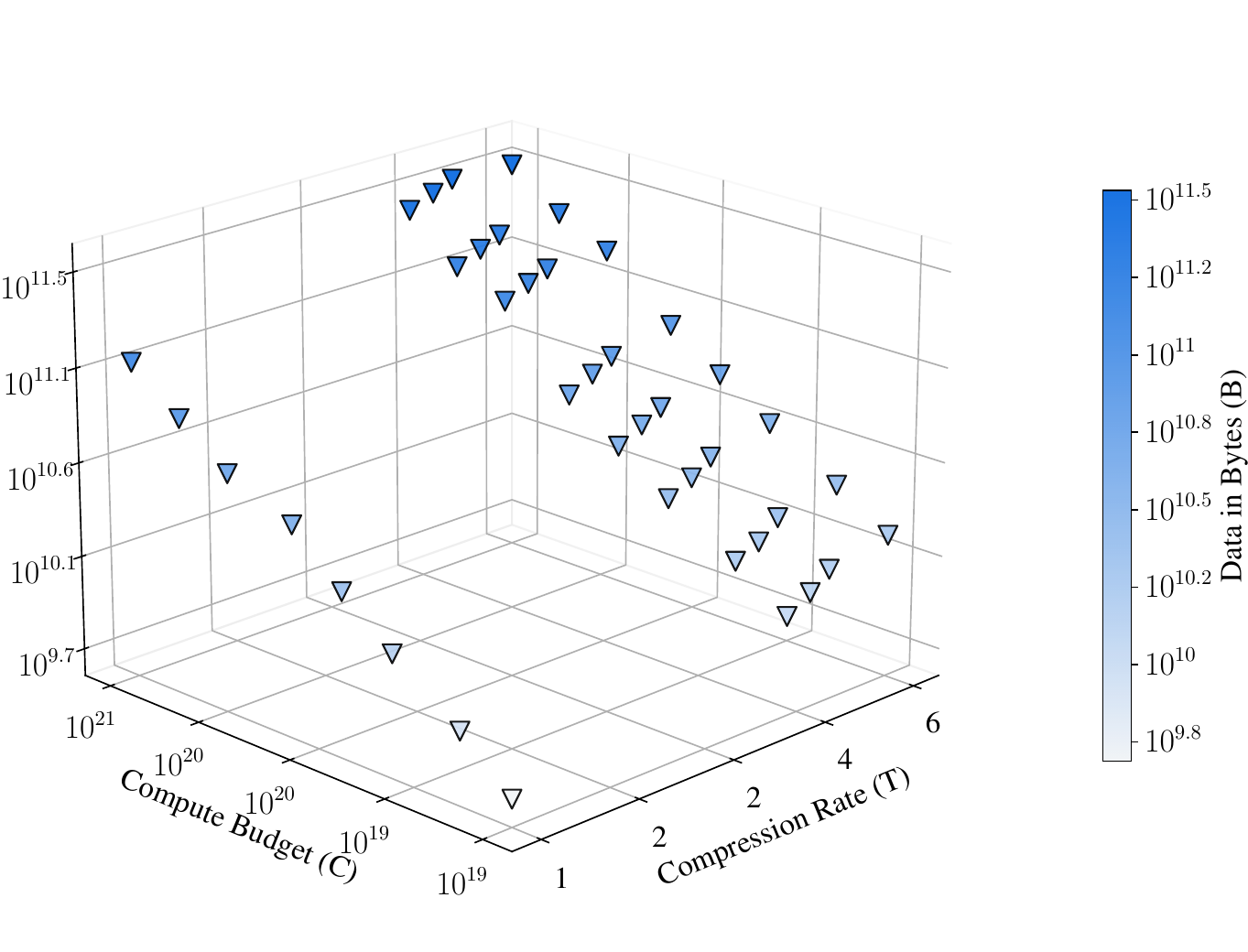}
    \caption{Amount of data}
    \label{fig:optimal_data_iso}
    \end{subfigure}
\hfill
    \begin{subfigure}[t]{0.49\textwidth}
    \centering
    \includegraphics[width=\linewidth]{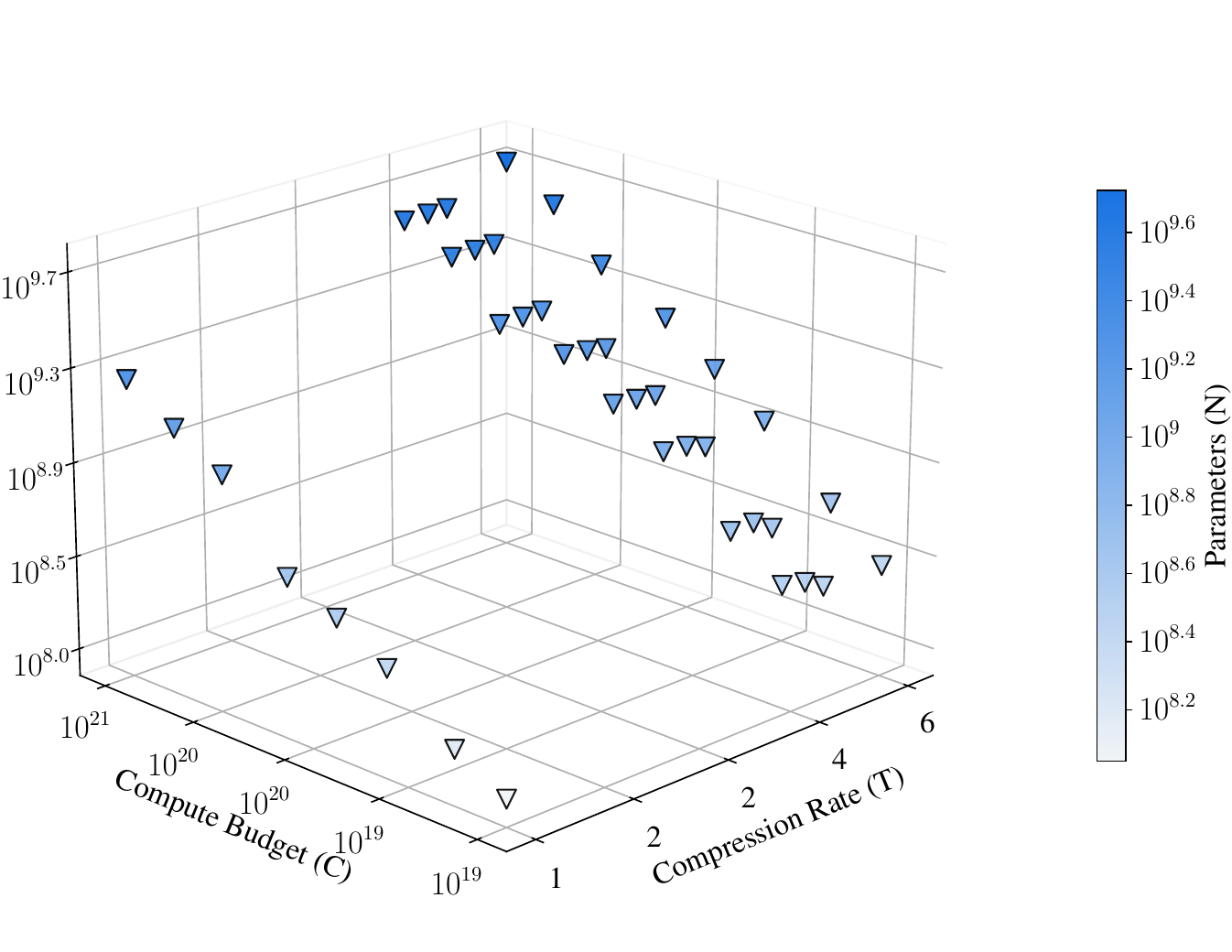}
    \caption{Model size}
    \label{fig:optimal_params_iso}
    \end{subfigure}
    \caption{Optimal data and model size configurations for each compute budget and compression rate (subword tokenized models).}
    \label{fig:optimal_data_params_iso}
\end{figure}

Figure~\ref{fig:optimal_data_params_iso} shows the optimal data in bytes $B^*$ and parameter counts $N^*$ across compressions and compute budgets for subword tokenized models.

\subsection{Loss Obtained by Optimal Configurations}

\input{tables/blt_budget_bpb_comparison}

In Tables~\ref{tab:blt_budget_bpb_comparison}~and~\ref{tab:isotropic_budget_bpb_comparison}, we present the best scores obtained by models (i.e., not derived from scaling law) respectively for latent and subword tokenized models.

\subsection{Multilingual 2D IsoFLOP}
\label{sec:xling_1d_isoflop}

\begin{figure*}[!htb]
    \centering
    \begin{subfigure}{0.48\textwidth}
        \includegraphics[width=\linewidth]{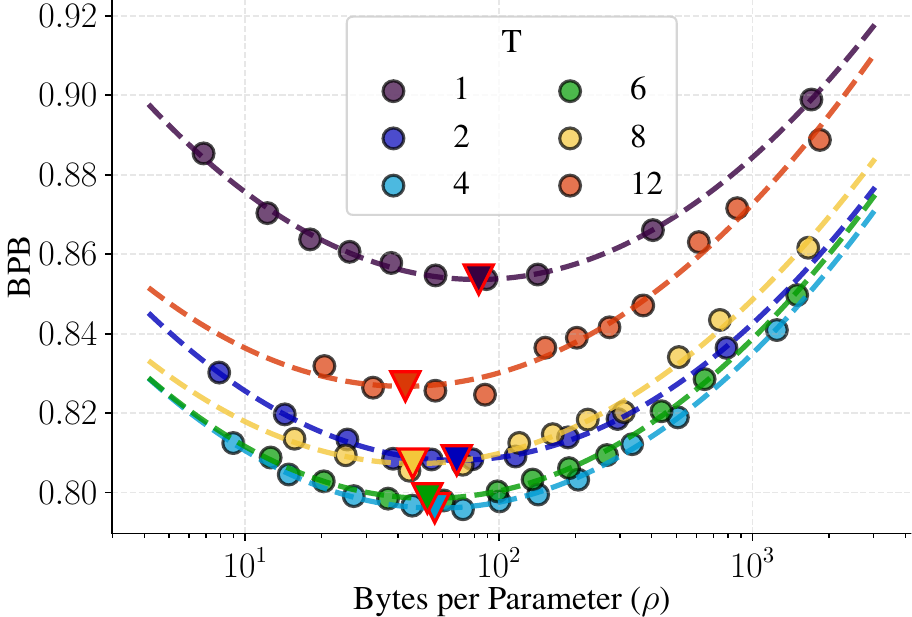}
        \caption{French (Latin)}
    \end{subfigure}
    \begin{subfigure}{0.48\textwidth}
        \includegraphics[width=\linewidth]{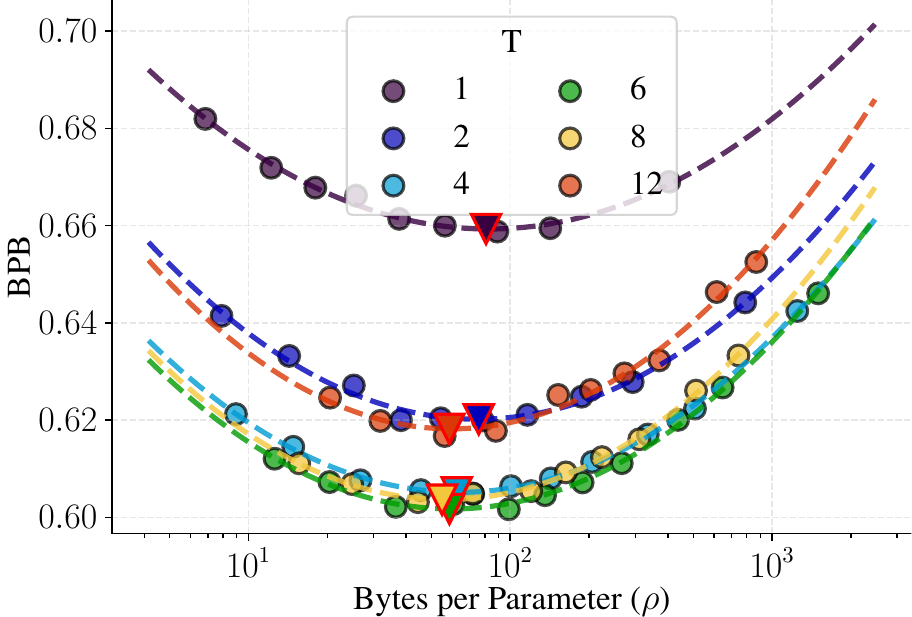}
        \caption{Vietnamese (Latin)}
    \end{subfigure}
    \begin{subfigure}{0.48\textwidth}
        \includegraphics[width=\linewidth]{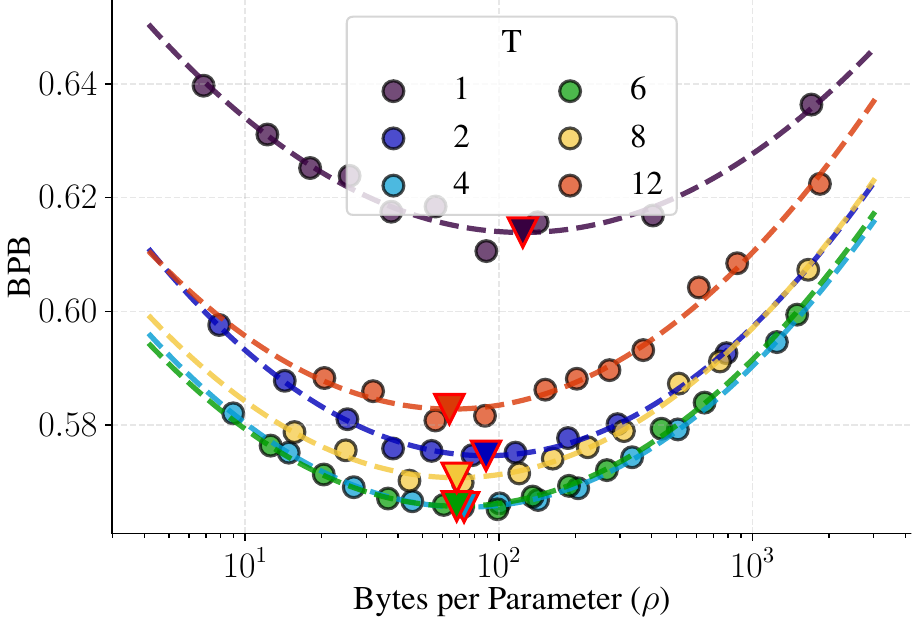}
        \caption{Arabic (Arabic)}
    \end{subfigure}
    \begin{subfigure}{0.48\textwidth}
        \includegraphics[width=\linewidth]{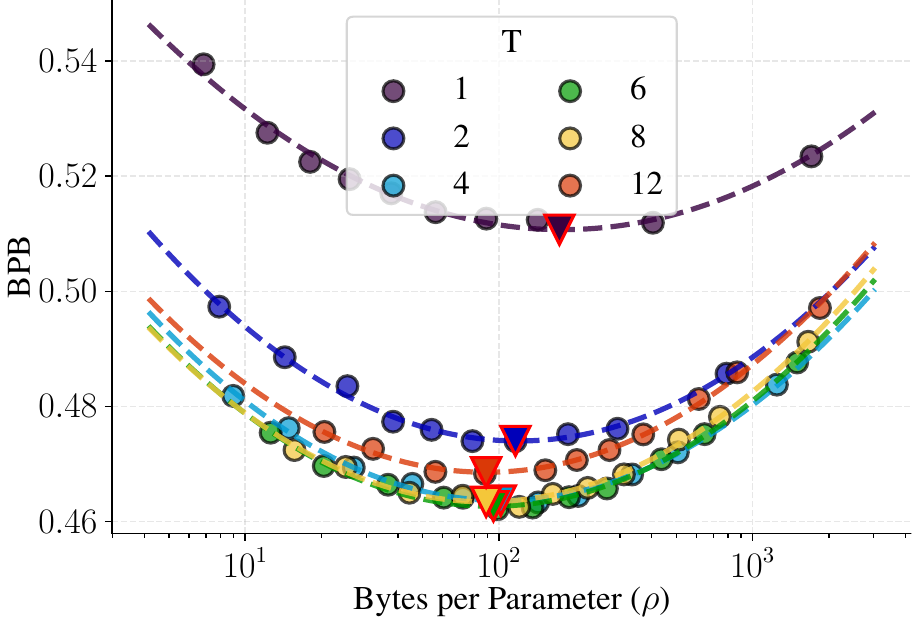}
        \caption{Russian (Cyrillic)}
    \end{subfigure}
        \begin{subfigure}{0.48\textwidth}
        \includegraphics[width=\linewidth]{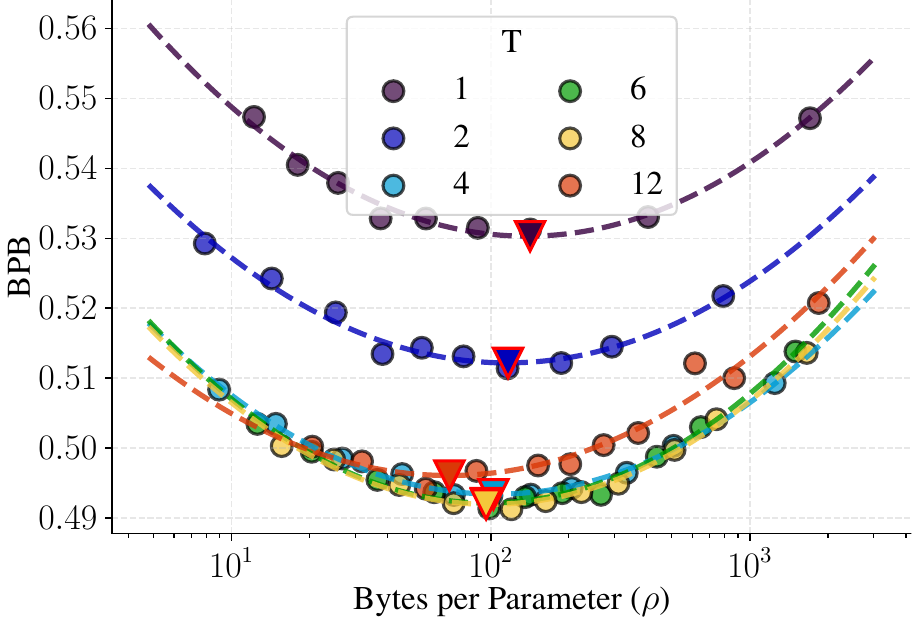}
        \caption{English x2 (Latin)}
    \end{subfigure}
    \begin{subfigure}{0.48\textwidth}
        \includegraphics[width=\linewidth]{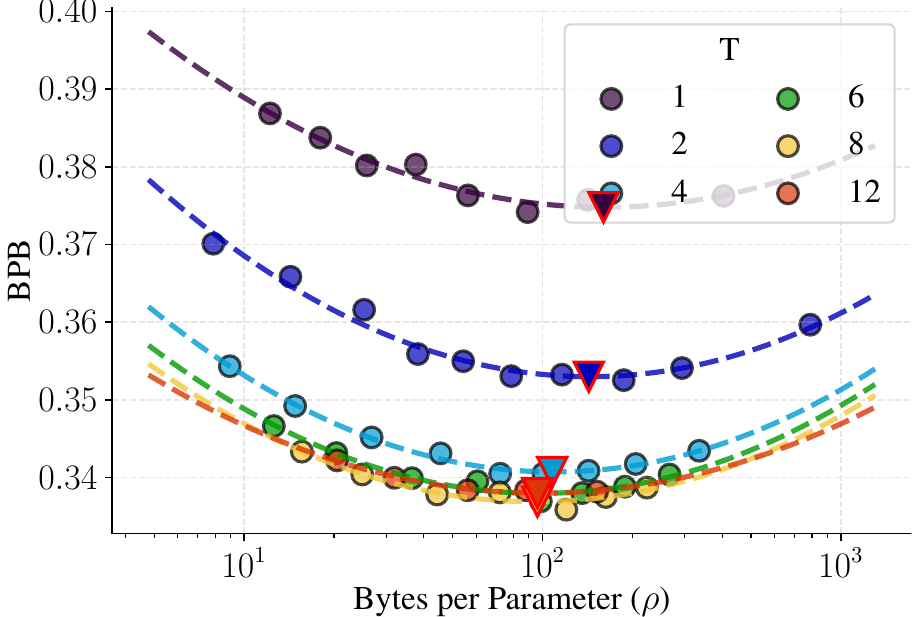}
        \caption{Hindi (Devanagari)}
    \end{subfigure}
    \caption{2D IsoFLOP fits across languages ($C=10^{20}$); all models use latent tokenization to achieve the set compression. Parabolas are fitted for each compression line to interpolate values of the loss.}
    \label{fig:1d_xlingual_1e20}
\end{figure*}

In Figure~\ref{fig:1d_xlingual_1e20}, we present 2-dimensional IsoFLOP for six considered languages.
The visualization is based on the same data as used for 3-dimensional IsoFLOP in Figure~\ref{fig:2d_xlingual_1e20}.

\clearpage

\subsection{Comparison between Character and Byte-level Models}
\label{sec:char-vs-byte}

In our analysis of subword tokenized models we focus on character-based instead of byte-based models to examine the properties of low compression.
The main difference between these models is that the former has a much larger vocabulary (148,000 vs. 256), while achieving a similar compression rate.
In our experiments, we consider character models to coerce on a similar vocabulary size as in BPE and SuperBPE.

\begin{figure}[!htb]
    \centering
    \includegraphics[width=0.75\linewidth]{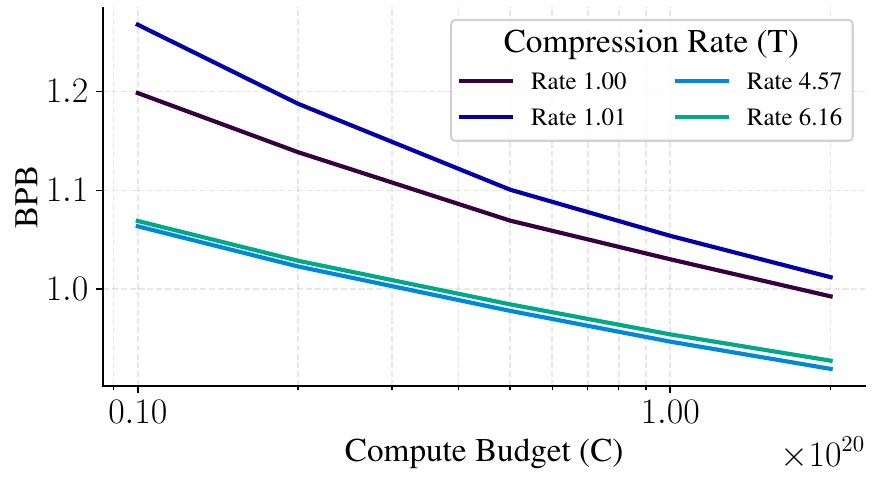}
    \caption{Comparison of optimal test losses for subword tokenized models: byte $T=1.00$; character $T=1.01$;
    BPE $T=4.57$; SuperBPE $T=6.16$.}
    \label{fig:chat_vs_byte_isotropic}
\end{figure}

We compare the loss of parameter optimal character ($T=1.01$) and byte models ($T=1.0$) in Figure~\ref{fig:chat_vs_byte_isotropic}. Notably, the gap between them is large for a small compute budget due to the relatively high cost of the embedding layer in small models.
With the increase of the training budget, the difference narrows.
This allows us to assume that character and byte tokenized models will follow similar scaling trends at larger scales.
Therefore, in the most of experiments we only consider character-based models.

\clearpage 

\subsection{AI2 Reasoning Challenge Results}
\label{sec:arc-results}

\begin{figure}[!htb]
    \centering
    \begin{subfigure}[t]{0.49\textwidth}
    \includegraphics[width=\linewidth]{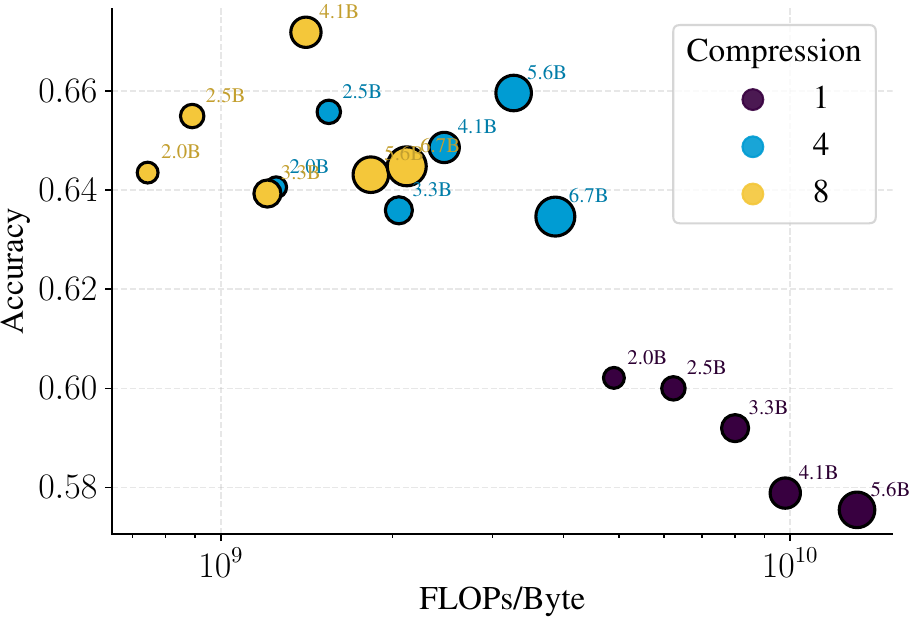}
    \caption{0-shot Accuracy on ARC-Easy}
    \end{subfigure}
\hfill
    \begin{subfigure}[t]{0.49\textwidth}
    \centering
    \includegraphics[width=\linewidth]{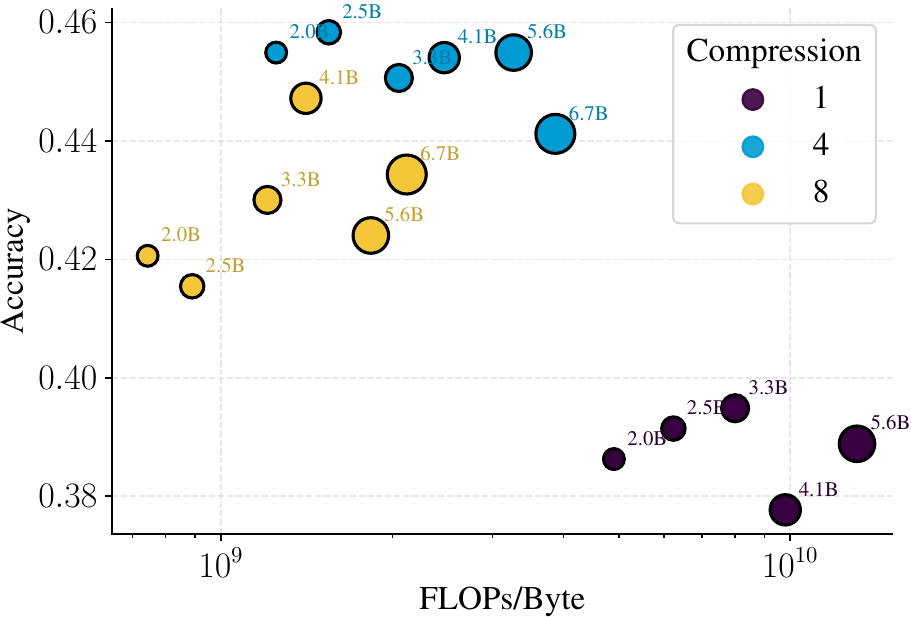}
    \caption{0-shot Accuracy on ARC-Challenge}
    \end{subfigure}
    \caption{Evaluation of the BLT models trained for $C=2\times10^{21}$ FLOPs on AI2 Reasoning Challange benchmark. The size of each point corresponds to the model parameter count. The results are plotted against inference compute cost per byte, which is dependent on model size $N$ and compression rate $T$.}
    \label{fig:endtask_arc}
\end{figure}

Figure~\ref{fig:endtask_arc} presents evaluations on multiple-choice questions from the AI2 Reasoning Challenge~\citep{clark2018arc}.
Interestingly, we observe that for the easier version of the task, models with \compr{} 8 and \compr{} 4 achieve similar scores.
The higher compression (\compr{} 8) even obtains the best score for the 4.1B-parameter model, while being cheaper to run than the corresponding \compr{} 4 model.
On the harder ``challenge'' split, we observe a different pattern: \compr{} 4 achieves higher scores than \compr{} 8.
We conclude that the choice of optimal compression can be task-dependent.
More-compressed, and thus cheaper, tokenization may be adequate for easier tasks, while harder tasks may benefit from the additional inference compute associated with lower compression.
We also note the underperformance of byte-level models, which we attribute to insufficient data seen during pre-training.

%% file: tables/BLT_scales.tex
\begin{table}[htbp]
\centering
\setlength{\tabcolsep}{4pt}
\renewcommand{\arraystretch}{1.2}
\rowcolors{6}{metabg}{white}
\begin{tabular}{ccccccccccccc}
\toprule
\multicolumn{4}{c}{\textbf{Global (Latent Module)}} & \multicolumn{4}{c}{\textbf{Local (Encoder/Decoder)}} & \multicolumn{2}{c}{\textbf{Cross-Attention}} & \textbf{Total}
\\ 
\cmidrule(lr){1-4} \cmidrule(lr){5-8} \cmidrule(lr){9-10}
\textbf{Layers} & \textbf{Heads} & \textbf{Dim} & \textbf{Params} & \textbf{Layers} & \textbf{Heads} & \textbf{Dim} & \textbf{Params} & \textbf{Heads} & \textbf{k}  & 
\textbf{Params} \\
\midrule
5  & 5  & 640  & 25M   & 2  & 10 & 640  & 10M   & 10 & 1 & 50M    \\
6  & 6  & 768  & 43M   & 2  & 10 & 640  & 10M   & 10 & 1 & 68M    \\
7  & 7  & 896  & 67M   & 2  & 10 & 640  & 10M   & 10 & 1 & 93M    \\
8  & 8  & 1024 & 101M  & 2  & 10 & 640  & 10M   & 10 & 1 & 127M   \\
9  & 9  & 1152 & 143M  & 3  & 12 & 768  & 21M   & 12 & 2 & 199M   \\
10 & 10 & 1280 & 197M  & 3  & 12 & 768  & 21M   & 12 & 2 & 253M   \\
11 & 11 & 1408 & 262M  & 3  & 12 & 768  & 21M   & 12 & 2 & 318M   \\
12 & 12 & 1536 & 340M  & 3  & 12 & 768  & 21M   & 12 & 2 & 396M   \\
13 & 13 & 1664 & 432M  & 4  & 12 & 768  & 28M   & 12 & 2 & 506M   \\
14 & 14 & 1792 & 540M  & 4  & 12 & 768  & 28M   & 12 & 2 & 613M   \\
15 & 15 & 1920 & 664M  & 4  & 12 & 768  & 28M   & 12 & 2 & 738M   \\
16 & 16 & 2048 & 805M  & 4  & 12 & 768  & 28M   & 12 & 2 & 880M   \\
17 & 17 & 2176 & 966M  & 5  & 14 & 896  & 48M   & 14 & 3 & 1.1B   \\
18 & 18 & 2304 & 1.1B  & 5  & 14 & 896  & 48M   & 14 & 3 & 1.3B   \\
19 & 19 & 2432 & 1.3B  & 5  & 14 & 896  & 48M   & 14 & 3 & 1.5B   \\
20 & 20 & 2560 & 1.6B  & 5  & 14 & 896  & 48M   & 14 & 3 & 1.7B   \\
21 & 21 & 2688 & 1.8B  & 6  & 14 & 896  & 58M   & 14 & 3 & 2.0B   \\
22 & 22 & 2816 & 2.1B  & 6  & 14 & 896  & 58M   & 14 & 3 & 2.2B   \\
23 & 23 & 2944 & 2.4B  & 6  & 14 & 896  & 58M   & 14 & 3 & 2.5B   \\
24 & 24 & 3072 & 2.7B  & 6  & 14 & 896  & 58M   & 14 & 3 & 2.9B   \\
25 & 25 & 3200 & 3.1B  & 7  & 16 & 1024 & 88M   & 16 & 4 & 3.3B   \\
26 & 26 & 3328 & 3.5B  & 7  & 16 & 1024 & 88M   & 16 & 4 & 3.7B   \\
27 & 27 & 3456 & 3.9B  & 7  & 16 & 1024 & 88M   & 16 & 4 & 4.1B   \\
28 & 28 & 3584 & 4.3B  & 7  & 16 & 1024 & 88M   & 16 & 4 & 4.6B   \\
29 & 29 & 3712 & 4.8B  & 8  & 16 & 1024 & 101M  & 16 & 4 & 5.1B   \\
30 & 30 & 3840 & 5.3B  & 8  & 16 & 1024 & 101M  & 16 & 4 & 5.6B   \\
31 & 31 & 3968 & 5.9B  & 8  & 16 & 1024 & 101M  & 16 & 4 & 6.1B   \\
32 & 32 & 4096 & 6.4B  & 8  & 16 & 1024 & 101M  & 16 & 4 & 6.7B   \\
\bottomrule
\end{tabular}
\caption{The configuration of latent tokenized models (BLT architecture) used in scaling experiments.}
\label{tab:blt_scales}
\end{table}

%% file: tables/isotropic_scales.tex
\begin{table}[htbp]
\centering
\rowcolors{6}{metabg}{white}
\setlength{\tabcolsep}{4pt}
\renewcommand{\arraystretch}{1.2}
\begin{tabular}{cccccccccccc}
\toprule
\multicolumn{4}{c}{\textbf{Global (Latent Module)}} & \multicolumn{3}{c}{\textbf{Local Parameters (Embeddings)}} & \multicolumn{3}{c}{\textbf{Total Parameters}} \\
\cmidrule(lr){1-4} \cmidrule(lr){5-7} \cmidrule(lr){8-10}
\textbf{Layers} & \textbf{Heads} & \textbf{Dim} & \textbf{Params} & \textbf{Char} & \textbf{BPE} & \textbf{SuperBPE} & \textbf{Char} & \textbf{BPE} & \textbf{SuperBPE} \\
\midrule
5  & 5  & 640  & 25M   & 96M & 82M   & 128M   & 121M   & 107M   & 153M   \\
6  & 6  & 768  & 43M   & 115M & 98M   & 154M   & 158M   & 141M   & 196M   \\
7  & 7  & 896  & 67M   & 134M & 115M  & 179M   & 202M   & 182M   & 247M   \\
8  & 8  & 1024 & 101M  & 154M & 131M  & 205M   & 254M  & 232M   & 306M   \\
9  & 9  & 1152 & 143M  & 173M & 148M  & 230M   & 316M  & 291M   & 374M   \\
10 & 10 & 1280 & 197M  & 192M & 164M  & 256M   & 389M  & 360M   & 453M   \\
11 & 11 & 1408 & 262M  & 211M & 180M  & 282M   & 473M  & 442M   & 543M   \\
12 & 12 & 1536 & 340M  & 230M & 197M  & 307M   & 570M  & 536M   & 647M   \\
13 & 13 & 1664 & 432M  & 250M & 213M  & 333M   & 682M  & 645M   & 765M   \\
14 & 14 & 1792 & 540M  & 269M & 229M  & 358M   & 808M  & 769M   & 898M   \\
15 & 15 & 1920 & 664M  & 288M & 246M  & 384M   & 952M  & 909M   & 1.0B   \\
16 & 16 & 2048 & 805M  & 307M & 262M  & 410M   & 1.1B  & 1.1B   & 1.2B   \\
17 & 17 & 2176 & 966M  & 326M & 279M  & 435M   & 1.3B  & 1.2B   & 1.4B   \\
18 & 18 & 2304 & 1.1B  & 346M & 295M  & 461M   & 1.5B  & 1.4B   & 1.6B   \\
19 & 19 & 2432 & 1.3B  & 365M & 311M  & 486M   & 1.7B  & 1.7B   & 1.8B   \\
20 & 20 & 2560 & 1.6B  & 384M & 328M  & 512M   & 2.0B  & 1.9B   & 2.1B   \\
21 & 21 & 2688 & 1.8B  & 403M & 344M  & 538M   & 2.2B  & 2.2B   & 2.4B   \\
22 & 22 & 2816 & 2.1B  & 422M & 360M  & 563M   & 2.5B  & 2.5B   & 2.7B   \\
23 & 23 & 2944 & 2.4B  & 442M & 377M  & 589M   & 2.8B  & 2.8B   & 3.0B   \\
24 & 24 & 3072 & 2.7B  & 461M & 393M  & 614M   & 3.2B  & 3.1B   & 3.3B   \\
25 & 25 & 3200 & 3.1B  & 480M & 410M  & 640M   & 3.6B  & 3.5B   & 3.7B   \\
26 & 26 & 3328 & 3.5B  & 499M & 426M  & 666M   & 4.0B  & 3.9B   & 4.1B   \\
27 & 27 & 3456 & 3.9B  & 518M & 442M  & 691M   & 4.4B  & 4.3B   & 4.6B   \\
28 & 28 & 3584 & 4.3B  & 538M & 459M  & 717M   & 4.9B  & 4.8B   & 5.0B   \\
29 & 29 & 3712 & 4.8B  & 557M & 475M  & 742M   & 5.4B  & 5.3B   & 5.5B   \\
30 & 30 & 3840 & 5.3B  & 576M & 492M  & 768M   & 5.9B  & 5.8B   & 6.1B   \\
31 & 31 & 3968 & 5.9B  & 595M & 508M  & 794M   & 6.5B  & 6.4B   & 6.7B   \\
32 & 32 & 4096 & 6.4B  & 614M & 524M  & 819M   & 7.1B  & 7.0B   & 7.3B   \\
\bottomrule
\end{tabular}
\caption{The configuration of subword tokenized models (isotropic). Parameter differences across tokenizers arise from varying vocabulary sizes $V$. For Character: $V=148,000$, for BPE: $V=128,000$, for SuperBPE $V=200,000$.}
\label{tab:isotropic_scales}
\end{table}

%% file: tables/scales_flops_share.tex
\begin{table}[htbp]
  \centering
  \setlength{\tabcolsep}{4pt}
  \renewcommand{\arraystretch}{1.1}
  \begin{tabular}{cccccccccc}
    \toprule
    & \multicolumn{6}{c}{Latent Tokenization} &
      \multicolumn{3}{c}{Subword Tokenizetion} \\
    \cmidrule(lr){2-7} \cmidrule(lr){8-10}
    \textbf{Scale} &
    \multicolumn{1}{c}{\textbf{$T{=}1$}} &
    \multicolumn{1}{c}{\textbf{$T{=}2$}} &
    \multicolumn{1}{c}{\textbf{$T{=}4$}} &
    \multicolumn{1}{c}{\textbf{$T{=}6$}} &
    \multicolumn{1}{c}{\textbf{$T{=}8$}} &
    \multicolumn{1}{c}{\textbf{$T{=}12$}} &
    \multicolumn{1}{c}{\textbf{$T{=}1.01$}} &
    \multicolumn{1}{c}{\textbf{$T{=}4.57$}} &
    \multicolumn{1}{c}{\textbf{$T{=}6.16$}} \\
    \midrule
    5  & \cellcolor{metabg!66}66\% & \cellcolor{metabg!43}43\% & \cellcolor{metabg!24}24\% & \cellcolor{metabg!17}17\% & \cellcolor{metabg!13}13\% & \cellcolor{metabg!9}9\%   & \cellcolor{metabg!34}34\% & \cellcolor{metabg!27}27\% & \cellcolor{metabg!18}18\% \\
    6  & \cellcolor{metabg!75}75\% & \cellcolor{metabg!55}55\% & \cellcolor{metabg!35}35\% & \cellcolor{metabg!25}25\% & \cellcolor{metabg!20}20\% & \cellcolor{metabg!14}14\% & \cellcolor{metabg!41}41\% & \cellcolor{metabg!34}34\% & \cellcolor{metabg!24}24\% \\
    7  & \cellcolor{metabg!82}82\% & \cellcolor{metabg!65}65\% & \cellcolor{metabg!45}45\% & \cellcolor{metabg!35}35\% & \cellcolor{metabg!28}28\% & \cellcolor{metabg!20}20\% & \cellcolor{metabg!47}47\% & \cellcolor{metabg!41}41\% & \cellcolor{metabg!30}30\% \\
    8  & \cellcolor{metabg!87}87\% & \cellcolor{metabg!73}73\% & \cellcolor{metabg!55}55\% & \cellcolor{metabg!44}44\% & \cellcolor{metabg!36}36\% & \cellcolor{metabg!27}27\% & \cellcolor{metabg!52}52\% & \cellcolor{metabg!47}47\% & \cellcolor{metabg!35}35\% \\
    9  & \cellcolor{metabg!79}79\% & \cellcolor{metabg!63}63\% & \cellcolor{metabg!44}44\% & \cellcolor{metabg!34}34\% & \cellcolor{metabg!27}27\% & \cellcolor{metabg!20}20\% & \cellcolor{metabg!57}57\% & \cellcolor{metabg!52}52\% & \cellcolor{metabg!41}41\% \\
    10 & \cellcolor{metabg!84}84\% & \cellcolor{metabg!70}70\% & \cellcolor{metabg!52}52\% & \cellcolor{metabg!41}41\% & \cellcolor{metabg!34}34\% & \cellcolor{metabg!25}25\% & \cellcolor{metabg!61}61\% & \cellcolor{metabg!57}57\% & \cellcolor{metabg!45}45\% \\
    11 & \cellcolor{metabg!87}87\% & \cellcolor{metabg!75}75\% & \cellcolor{metabg!58}58\% & \cellcolor{metabg!48}48\% & \cellcolor{metabg!41}41\% & \cellcolor{metabg!31}31\% & \cellcolor{metabg!65}65\% & \cellcolor{metabg!62}62\% & \cellcolor{metabg!50}50\% \\
    12 & \cellcolor{metabg!89}89\% & \cellcolor{metabg!79}79\% & \cellcolor{metabg!64}64\% & \cellcolor{metabg!54}54\% & \cellcolor{metabg!47}47\% & \cellcolor{metabg!37}37\% & \cellcolor{metabg!68}68\% & \cellcolor{metabg!65}65\% & \cellcolor{metabg!54}54\% \\
    13 & \cellcolor{metabg!89}89\% & \cellcolor{metabg!78}78\% & \cellcolor{metabg!63}63\% & \cellcolor{metabg!53}53\% & \cellcolor{metabg!46}46\% & \cellcolor{metabg!36}36\% & \cellcolor{metabg!71}71\% & \cellcolor{metabg!69}69\% & \cellcolor{metabg!58}58\% \\
    14 & \cellcolor{metabg!91}91\% & \cellcolor{metabg!82}82\% & \cellcolor{metabg!68}68\% & \cellcolor{metabg!58}58\% & \cellcolor{metabg!51}51\% & \cellcolor{metabg!41}41\% & \cellcolor{metabg!73}73\% & \cellcolor{metabg!72}72\% & \cellcolor{metabg!62}62\% \\
    15 & \cellcolor{metabg!92}92\% & \cellcolor{metabg!84}84\% & \cellcolor{metabg!72}72\% & \cellcolor{metabg!63}63\% & \cellcolor{metabg!56}56\% & \cellcolor{metabg!46}46\% & \cellcolor{metabg!76}76\% & \cellcolor{metabg!74}74\% & \cellcolor{metabg!65}65\% \\
    16 & \cellcolor{metabg!93}93\% & \cellcolor{metabg!87}87\% & \cellcolor{metabg!76}76\% & \cellcolor{metabg!68}68\% & \cellcolor{metabg!61}61\% & \cellcolor{metabg!51}51\% & \cellcolor{metabg!78}78\% & \cellcolor{metabg!77}77\% & \cellcolor{metabg!67}67\% \\
    17 & \cellcolor{metabg!90}90\% & \cellcolor{metabg!82}82\% & \cellcolor{metabg!69}69\% & \cellcolor{metabg!59}59\% & \cellcolor{metabg!52}52\% & \cellcolor{metabg!42}42\% & \cellcolor{metabg!80}80\% & \cellcolor{metabg!79}79\% & \cellcolor{metabg!70}70\% \\
    18 & \cellcolor{metabg!91}91\% & \cellcolor{metabg!84}84\% & \cellcolor{metabg!72}72\% & \cellcolor{metabg!63}63\% & \cellcolor{metabg!57}57\% & \cellcolor{metabg!46}46\% & \cellcolor{metabg!81}81\% & \cellcolor{metabg!81}81\% & \cellcolor{metabg!72}72\% \\
    19 & \cellcolor{metabg!93}93\% & \cellcolor{metabg!86}86\% & \cellcolor{metabg!75}75\% & \cellcolor{metabg!67}67\% & \cellcolor{metabg!60}60\% & \cellcolor{metabg!50}50\% & \cellcolor{metabg!83}83\% & \cellcolor{metabg!82}82\% & \cellcolor{metabg!74}74\% \\
    20 & \cellcolor{metabg!93}93\% & \cellcolor{metabg!88}88\% & \cellcolor{metabg!78}78\% & \cellcolor{metabg!70}70\% & \cellcolor{metabg!64}64\% & \cellcolor{metabg!54}54\% & \cellcolor{metabg!84}84\% & \cellcolor{metabg!84}84\% & \cellcolor{metabg!76}76\% \\
    21 & \cellcolor{metabg!93}93\% & \cellcolor{metabg!87}87\% & \cellcolor{metabg!77}77\% & \cellcolor{metabg!70}70\% & \cellcolor{metabg!63}63\% & \cellcolor{metabg!53}53\% & \cellcolor{metabg!85}85\% & \cellcolor{metabg!85}85\% & \cellcolor{metabg!78}78\% \\
    22 & \cellcolor{metabg!94}94\% & \cellcolor{metabg!89}89\% & \cellcolor{metabg!80}80\% & \cellcolor{metabg!72}72\% & \cellcolor{metabg!66}66\% & \cellcolor{metabg!57}57\% & \cellcolor{metabg!86}86\% & \cellcolor{metabg!86}86\% & \cellcolor{metabg!79}79\% \\
    23 & \cellcolor{metabg!95}95\% & \cellcolor{metabg!90}90\% & \cellcolor{metabg!82}82\% & \cellcolor{metabg!75}75\% & \cellcolor{metabg!69}69\% & \cellcolor{metabg!60}60\% & \cellcolor{metabg!87}87\% & \cellcolor{metabg!87}87\% & \cellcolor{metabg!81}81\% \\
    24 & \cellcolor{metabg!95}95\% & \cellcolor{metabg!91}91\% & \cellcolor{metabg!84}84\% & \cellcolor{metabg!77}77\% & \cellcolor{metabg!72}72\% & \cellcolor{metabg!63}63\% & \cellcolor{metabg!88}88\% & \cellcolor{metabg!88}88\% & \cellcolor{metabg!82}82\% \\
    25 & \cellcolor{metabg!93}93\% & \cellcolor{metabg!88}88\% & \cellcolor{metabg!79}79\% & \cellcolor{metabg!71}71\% & \cellcolor{metabg!65}65\% & \cellcolor{metabg!56}56\% & \cellcolor{metabg!89}89\% & \cellcolor{metabg!89}89\% & \cellcolor{metabg!83}83\% \\
    26 & \cellcolor{metabg!94}94\% & \cellcolor{metabg!89}89\% & \cellcolor{metabg!81}81\% & \cellcolor{metabg!74}74\% & \cellcolor{metabg!68}68\% & \cellcolor{metabg!59}59\% & \cellcolor{metabg!89}89\% & \cellcolor{metabg!89}89\% & \cellcolor{metabg!84}84\% \\
    27 & \cellcolor{metabg!95}95\% & \cellcolor{metabg!90}90\% & \cellcolor{metabg!82}82\% & \cellcolor{metabg!76}76\% & \cellcolor{metabg!70}70\% & \cellcolor{metabg!61}61\% & \cellcolor{metabg!90}90\% & \cellcolor{metabg!90}90\% & \cellcolor{metabg!85}85\% \\
    28 & \cellcolor{metabg!95}95\% & \cellcolor{metabg!91}91\% & \cellcolor{metabg!84}84\% & \cellcolor{metabg!78}78\% & \cellcolor{metabg!73}73\% & \cellcolor{metabg!64}64\% & \cellcolor{metabg!91}91\% & \cellcolor{metabg!91}91\% & \cellcolor{metabg!86}86\% \\
    29 & \cellcolor{metabg!95}95\% & \cellcolor{metabg!91}91\% & \cellcolor{metabg!83}83\% & \cellcolor{metabg!77}77\% & \cellcolor{metabg!72}72\% & \cellcolor{metabg!63}63\% & \cellcolor{metabg!91}91\% & \cellcolor{metabg!91}91\% & \cellcolor{metabg!87}87\% \\
    30 & \cellcolor{metabg!95}95\% & \cellcolor{metabg!92}92\% & \cellcolor{metabg!85}85\% & \cellcolor{metabg!79}79\% & \cellcolor{metabg!74}74\% & \cellcolor{metabg!66}66\% & \cellcolor{metabg!92}92\% & \cellcolor{metabg!92}92\% & \cellcolor{metabg!88}88\% \\
    31 & \cellcolor{metabg!96}96\% & \cellcolor{metabg!92}92\% & \cellcolor{metabg!86}86\% & \cellcolor{metabg!81}81\% & \cellcolor{metabg!76}76\% & \cellcolor{metabg!68}68\% & \cellcolor{metabg!92}92\% & \cellcolor{metabg!92}92\% & \cellcolor{metabg!88}88\% \\
    32 & \cellcolor{metabg!96}96\% & \cellcolor{metabg!93}93\% & \cellcolor{metabg!87}87\% & \cellcolor{metabg!82}82\% & \cellcolor{metabg!78}78\% & \cellcolor{metabg!70}70\% & \cellcolor{metabg!92}92\% & \cellcolor{metabg!93}93\% & \cellcolor{metabg!89}89\% \\
    \bottomrule
  \end{tabular}
  \caption{The compute cost per byte by global model as percentage of compute cost per byte of the whole model. The first column (Scale) denotes number of layers and heads of global module.
  In latent tokenization \compr{} $T \in \{1,2,4,6,8,12\}$ is set as hyperparameter, whereas in subword tokenization it is determined by the tokenizer (Character $T=1.01$, BPE $T=4.57$, SuperBPE $T=6.16$)}
  \label{tab:scales_flops_share}
\end{table}

%% file: tables/scaling_law_validation.tex
\begin{table}[htbp]
\centering
\rowcolors{2}{metabg}{white}
\setlength{\tabcolsep}{4pt}
\renewcommand{\arraystretch}{1.2}
\begin{tabular}{lccccccc}
\toprule
\textbf{Residual model} & \textbf{$E$} & \textbf{$F$} & \textbf{$T_0$} & \textbf{$\delta$} & RMSE & $R^2$ & $\bar{R}^2$ \\ \midrule
Mean of residuals (Eq.~\ref{eq:optimal_loss_irreducible}) & 0.7075 & --- & --- & --- & 0.0260 & 0.903 & 0.896 \\
Constant $T^\star$ (Eq.~\ref{eq:optimal_const_compression}) & 0.7075 & 0.0341 & 3.73 & --- & 0.0115 & 0.996 & 0.995 \\
Compute-dependent $T^\star$ (Eq.~\ref{eq:optimal_compression_rate})& 0.7075 & 0.0341 & 14.9 & 0.0302 & \textbf{0.0086}  & \textbf{0.997} & \textbf{0.996} \\ \bottomrule
\end{tabular}
\caption{
Comparison of the three considered forms for modeling $f(C,T)$ residuals in Equation~\ref{eq:optimal_loss}.
All functions were fitted using the $48$ runs with compute budgets less than or equal to $1 \times 10^{21}$ FLOPs.
To test extrapolation accuracy, Root Mean Square Error was computed for models trained at $2 \times 10^{21}$ FLOPs across 8 different compression rates.
All evaluations of extrapolation performance and goodness-of-fit (standard and adjusted for the number of variables) indicate that the model with compute-dependent \compr{} offers the best fit and extrapolation accuracy in loss estimation.
}
\label{tab:scaling_law_validation}
\end{table}

%% file: tables/tokenization_method_comparison.tex
\begin{table}[htbp]
\centering
\rowcolors{2}{metabg}{white}
\setlength{\tabcolsep}{4pt}
\renewcommand{\arraystretch}{1.2}
\begin{tabular}{lccc}
\toprule
\textbf{Tokenization Method} & \textbf{$\rho^\star$} & \textbf{$T^\star$} & \textbf{BPB} \\ \midrule
Latent (Entropy) & 62.1 & 3.71 & 0.960 \\
Latent (Fixed) & 60.0 & 3.87 & 0.973 \\
Subword & 58.8 & 5.36 & 0.947 \\ \bottomrule
\end{tabular}
\caption{Compute-optimal \bpp{} ($\rho^\star$) and \compr{} ($T^\star$) for different methods.
The values are close to each other, except for subword $T^\star$.}
\label{tab:tokenization_method_comparison}
\end{table}

%% file: tables/xlingual_mix.tex
\begin{table}[!htbp]
\centering
\rowcolors{3}{metabg}{white}
\setlength{\tabcolsep}{4pt}
\renewcommand{\arraystretch}{1.2}
\begin{tabular}{ccccccc}
\toprule
\textbf{Language} & \textbf{Parity} & \multicolumn{2}{c}{\textbf{$\rho^\star_l$}} & \multicolumn{2}{c}{\textbf{$T^\star_l$}} & \textbf{BPB} \\ \cmidrule(lr){3-4} \cmidrule(lr){5-6}
 &  & \textbf{Value} & \textbf{Ratio} & \textbf{Value} & \textbf{Ratio} & \\ \midrule
English    & 1.0 & 71.8 & 1.00 & 3.38 & 1.00 & 1.101 \\
French     & 1.2 & 72.5 & 1.01 & 3.65 & 1.08 & 0.931 \\
Vietnamese & 1.4 & 70.3 & 0.98 & 4.12 & 1.22 & 0.720 \\
Arabic     & 1.6 & 76.5 & 1.07 & 3.84 & 1.14 & 0.667 \\
Russian    & 2.0 & 77.6 & 1.08 & 5.03 & 1.49 & 0.532 \\
Hindi      & 2.6 & 68.9 & 0.96 & 6.32 & 1.87 & 0.387 \\
\bottomrule
\end{tabular}
\caption{Compute-optimal \bpp{} ($\rho^\star_l$), \compr{} ($T^\star_l$) compared to cross-lingual \pari{}.
Results for multilingual models, trained jointly on all six languages with $C=10^{20}$ FLOPs budget.
The \pari{} and compute-optimal ratios are proportions between each language and English baseline.}
\label{tab:xlingual_mix_optimals_1e20}
\end{table}

%% file: tables/blt_budget_bpb_comparison.tex
\begin{table}[tbp]
\centering
\rowcolors{4}{metabg}{white}
\setlength{\tabcolsep}{4pt}
\renewcommand{\arraystretch}{1.2}
\begin{tabular}{lcccccc}
\toprule
\textbf{Compute} & \multicolumn{6}{c}{\textbf{Latent Entropy}} \\
\cmidrule(lr){2-7}
\textbf{(FLOPs)} & 1 & 2 & 4 & 6 & 8 & 12 \\ \midrule
$1 \times 10^{19}$ & 1.1790 & 1.1178 & \textbf{1.1025} & 1.1095 & 1.1272 & 1.1598 \\
$2 \times 10^{19}$ & 1.1200 & 1.0642 & \textbf{1.0532} & 1.0587 & 1.0727 & 1.1047 \\
$5 \times 10^{19}$ & 1.0606 & 1.0080 & \textbf{0.9987} & 1.0049 & 1.0165 & 1.0422 \\
$1 \times 10^{20}$ & 1.0158 & 0.9694 & \textbf{0.9601} & 0.9631 & 0.9751 & 0.9993 \\
$2 \times 10^{20}$ & 0.9771 & 0.9359 & \textbf{0.9265} & 0.9314 & 0.9427 & 0.9650 \\
$5 \times 10^{20}$ & 0.9333 & 0.8974 & \textbf{0.8933} & 0.8990 & 0.9085 & 0.9278 \\
$1 \times 10^{21}$ & 0.9008 & 0.8722 & \textbf{0.8686} & 0.8744 & 0.8843 & 0.9041 \\
$2 \times 10^{21}$ & 0.8741 & 0.8491 & \textbf{0.8483} & 0.8543 & 0.8650 & 0.8844 \\ \midrule
\textbf{Compression:} & 1 & 2 & 4 & 6 & 8 & 12 \\ \bottomrule
\end{tabular}
\caption{Comparison of the lowest BPB obtained by latent tokenized models for specific compute budgets.}
\label{tab:blt_budget_bpb_comparison}
\end{table}

%% file: main.bbl
\begin{thebibliography}{48}
\providecommand{\natexlab}[1]{#1}
\providecommand{\url}[1]{\texttt{#1}}
\expandafter\ifx\csname urlstyle\endcsname\relax
  \providecommand{\doi}[1]{doi: #1}\else
  \providecommand{\doi}{doi: \begingroup \urlstyle{rm}\Url}\fi

\bibitem[Ahia et~al.(2023)Ahia, Kumar, Gonen, Kasai, Mortensen, Smith, and
  Tsvetkov]{ahia-etal-2023-languages}
Orevaoghene Ahia, Sachin Kumar, Hila Gonen, Jungo Kasai, David Mortensen, Noah
  Smith, and Yulia Tsvetkov.
\newblock Do all languages cost the same? tokenization in the era of commercial
  language models.
\newblock In Houda Bouamor, Juan Pino, and Kalika Bali, editors,
  \emph{Proceedings of the 2023 Conference on Empirical Methods in Natural
  Language Processing}, pages 9904--9923, Singapore, December 2023. Association
  for Computational Linguistics.
\newblock \doi{10.18653/v1/2023.emnlp-main.614}.
\newblock \url{https://aclanthology.org/2023.emnlp-main.614/}.

\bibitem[Ahia et~al.(2024)Ahia, Kumar, Gonen, Hofmann, Limisiewicz, Tsvetkov,
  and Smith]{ahia-etal-2024-magnet}
Orevaoghene Ahia, Sachin Kumar, Hila Gonen, Valentin Hofmann, Tomasz
  Limisiewicz, Yulia Tsvetkov, and Noah~A Smith.
\newblock Magnet: Improving the multilingual fairness of language models with
  adaptive gradient-based tokenization.
\newblock In A.~Globerson, L.~Mackey, D.~Belgrave, A.~Fan, U.~Paquet,
  J.~Tomczak, and C.~Zhang, editors, \emph{Advances in Neural Information
  Processing Systems}, volume~37, pages 47790--47814. Curran Associates, Inc.,
  2024.
\newblock \doi{10.52202/079017-1514}.
\newblock
  \url{https://proceedings.neurips.cc/paper_files/paper/2024/file/5572bc595de865c1450868fd5391e9c5-Paper-Conference.pdf}.

\bibitem[Clark et~al.(2018)Clark, Cowhey, Etzioni, Khot, Sabharwal, Schoenick,
  and Tafjord]{clark2018arc}
Peter Clark, Isaac Cowhey, Oren Etzioni, Tushar Khot, Ashish Sabharwal, Carissa
  Schoenick, and Oyvind Tafjord.
\newblock Think you have solved question answering? try arc, the ai2 reasoning
  challenge.
\newblock \emph{arXiv:1803.05457v1}, 2018.

\bibitem[Gall{\'e}(2019)]{galle-2019-investigating}
Matthias Gall{\'e}.
\newblock Investigating the effectiveness of {BPE}: The power of shorter
  sequences.
\newblock In Kentaro Inui, Jing Jiang, Vincent Ng, and Xiaojun Wan, editors,
  \emph{Proceedings of the 2019 Conference on Empirical Methods in Natural
  Language Processing and the 9th International Joint Conference on Natural
  Language Processing (EMNLP-IJCNLP)}, pages 1375--1381, Hong Kong, China,
  November 2019. Association for Computational Linguistics.
\newblock \doi{10.18653/v1/D19-1141}.
\newblock \url{https://aclanthology.org/D19-1141/}.

\bibitem[Goldman et~al.(2024)Goldman, Caciularu, Eyal, Cao, Szpektor, and
  Tsarfaty]{goldman-etal-2024-unpacking}
Omer Goldman, Avi Caciularu, Matan Eyal, Kris Cao, Idan Szpektor, and Reut
  Tsarfaty.
\newblock Unpacking tokenization: Evaluating text compression and its
  correlation with model performance.
\newblock In Lun-Wei Ku, Andre Martins, and Vivek Srikumar, editors,
  \emph{Findings of the Association for Computational Linguistics: ACL 2024},
  pages 2274--2286, Bangkok, Thailand, August 2024. Association for
  Computational Linguistics.
\newblock \doi{10.18653/v1/2024.findings-acl.134}.
\newblock \url{https://aclanthology.org/2024.findings-acl.134/}.

\bibitem[Goyal et~al.(2021)Goyal, Gao, Chaudhary, Chen, Wenzek, Ju, Krishnan,
  Ranzato, Guzm{\'a}n, and Fan]{goyal2021flores}
Naman Goyal, Cynthia Gao, Vishrav Chaudhary, Peng-Jen Chen, Guillaume Wenzek,
  Da~Ju, Sanjana Krishnan, Marc'Aurelio Ranzato, Francisco~(Paco) Guzm{\'a}n,
  and Angela Fan.
\newblock The flores-101 evaluation benchmark for low-resource and multilingual
  machine translation.
\newblock \emph{Transactions of the Association for Computational Linguistics},
  10:\penalty0 522--538, 2021.
\newblock \url{https://https://arxiv.org/abs/2106.03193}.

\bibitem[He et~al.(2025)He, Benhaim, Patra, Vaddamanu, Ahuja, Chopra,
  Chaudhary, Zhao, and Song]{he-etal-2025-scaling}
Yifei He, Alon Benhaim, Barun Patra, Praneetha Vaddamanu, Sanchit Ahuja, Parul
  Chopra, Vishrav Chaudhary, Han Zhao, and Xia Song.
\newblock Scaling laws for multilingual language models.
\newblock In Wanxiang Che, Joyce Nabende, Ekaterina Shutova, and Mohammad~Taher
  Pilehvar, editors, \emph{Findings of the Association for Computational
  Linguistics: ACL 2025}, pages 4257--4273, Vienna, Austria, July 2025.
  Association for Computational Linguistics.
\newblock ISBN 979-8-89176-256-5.
\newblock \doi{10.18653/v1/2025.findings-acl.221}.
\newblock \url{https://aclanthology.org/2025.findings-acl.221/}.

\bibitem[Hoffmann et~al.(2022)Hoffmann, Borgeaud, Mensch, Buchatskaya, Cai,
  Rutherford, de~Las~Casas, Hendricks, Welbl, Clark, Hennigan, Noland,
  Millican, van~den Driessche, Damoc, Guy, Osindero, Simonyan, Elsen, Rae,
  Vinyals, and Sifre]{hoffmann2022training}
Jordan Hoffmann, Sebastian Borgeaud, Arthur Mensch, Elena Buchatskaya, Trevor
  Cai, Eliza Rutherford, Diego de~Las~Casas, Lisa~Anne Hendricks, Johannes
  Welbl, Aidan Clark, Tom Hennigan, Eric Noland, Katie Millican, George van~den
  Driessche, Bogdan Damoc, Aurelia Guy, Simon Osindero, Karen Simonyan, Erich
  Elsen, Jack~W. Rae, Oriol Vinyals, and Laurent Sifre.
\newblock Training compute-optimal large language models.
\newblock \emph{arXiv preprint arXiv:2203.15556}, 2022.
\newblock \url{https://arxiv.org/abs/2203.15556}.

\bibitem[Hwang et~al.(2025)Hwang, Wang, and
  Gu]{hwang2025dynamicchunkingendtoendhierarchical}
Sukjun Hwang, Brandon Wang, and Albert Gu.
\newblock Dynamic chunking for end-to-end hierarchical sequence modeling, 2025.
\newblock \url{https://arxiv.org/abs/2507.07955}.

\bibitem[Kaplan et~al.(2020)Kaplan, McCandlish, Henighan, Brown, Chess, Child,
  Gray, Radford, Wu, and Amodei]{kaplan2020scaling}
Jared Kaplan, Sam McCandlish, Tom Henighan, Tom~B. Brown, Benjamin Chess, Rewon
  Child, Scott Gray, Alec Radford, Jeffrey Wu, and Dario Amodei.
\newblock Scaling laws for neural language models.
\newblock \emph{arXiv preprint arXiv:2001.08361}, 2020.
\newblock \url{https://arxiv.org/abs/2001.08361}.

\bibitem[Kudo(2018)]{kudo-2018-subword}
Taku Kudo.
\newblock Subword regularization: Improving neural network translation models
  with multiple subword candidates.
\newblock In Iryna Gurevych and Yusuke Miyao, editors, \emph{Proceedings of the
  56th Annual Meeting of the Association for Computational Linguistics (Volume
  1: Long Papers)}, pages 66--75, Melbourne, Australia, July 2018. Association
  for Computational Linguistics.
\newblock \doi{10.18653/v1/P18-1007}.
\newblock \url{https://aclanthology.org/P18-1007/}.

\bibitem[Li et~al.(2024)Li, Fang, Smez, Albalak, Mehta, Openshaw, Haber,
  Wortsman, Keh, Gadre, Taori, Tian, Jitsev, Ilharco, Smola, Farhadi, Shankar,
  Schmidt, Carmon, and Beaumont]{li2024datacomp}
Jeffrey Li, Alex Fang, Georgios Smez, Alon Albalak, Kaber Mehta, Etash
  Openshaw, Louis Haber, Mitchell Wortsman, Sedrick Keh, Samir~Yitzhak Gadre,
  Rohan Taori, Shuran Tian, Jenia Jitsev, Gabriel Ilharco, Alexander Smola, Ali
  Farhadi, Vaishaal Shankar, Ludwig Schmidt, Yair Carmon, and Romain Beaumont.
\newblock Datacomp-lm: In search of the next generation of training sets for
  language models.
\newblock \emph{arXiv preprint arXiv:2406.11794}, 2024.
\newblock \url{https://arxiv.org/abs/2406.11794}.

\bibitem[Li et~al.(2025)Li, Kudugunta, and Zettlemoyer]{li2025misfitting}
Margaret Li, Sneha Kudugunta, and Luke Zettlemoyer.
\newblock (mis)fitting scaling laws: A survey of scaling law fitting techniques
  in deep learning.
\newblock In \emph{The Thirteenth International Conference on Learning
  Representations}, 2025.
\newblock \url{https://openreview.net/forum?id=xI71dsS3o4}.

\bibitem[Limisiewicz et~al.(2023)Limisiewicz, Balhar, and
  Mare{\v{c}}ek]{limisiewicz-etal-2023-tokenization}
Tomasz Limisiewicz, Ji{\v{r}}{\'i} Balhar, and David Mare{\v{c}}ek.
\newblock Tokenization impacts multilingual language modeling: Assessing
  vocabulary allocation and overlap across languages.
\newblock In Anna Rogers, Jordan Boyd-Graber, and Naoaki Okazaki, editors,
  \emph{Findings of the Association for Computational Linguistics: ACL 2023},
  pages 5661--5681, Toronto, Canada, July 2023. Association for Computational
  Linguistics.
\newblock \doi{10.18653/v1/2023.findings-acl.350}.
\newblock \url{https://aclanthology.org/2023.findings-acl.350/}.

\bibitem[Limisiewicz et~al.(2024)Limisiewicz, Blevins, Gonen, Ahia, and
  Zettlemoyer]{limisiewicz-etal-2024-myte}
Tomasz Limisiewicz, Terra Blevins, Hila Gonen, Orevaoghene Ahia, and Luke
  Zettlemoyer.
\newblock {MYTE}: Morphology-driven byte encoding for better and fairer
  multilingual language modeling.
\newblock In Lun-Wei Ku, Andre Martins, and Vivek Srikumar, editors,
  \emph{Proceedings of the 62nd Annual Meeting of the Association for
  Computational Linguistics (Volume 1: Long Papers)}, pages 15059--15076,
  Bangkok, Thailand, August 2024. Association for Computational Linguistics.
\newblock \doi{10.18653/v1/2024.acl-long.804}.
\newblock \url{https://aclanthology.org/2024.acl-long.804/}.

\bibitem[Liu et~al.(2025)Liu, Hayase, Hofmann, Oh, Smith, and
  Choi]{liu2025superbpespacetravellanguage}
Alisa Liu, Jonathan Hayase, Valentin Hofmann, Sewoong Oh, Noah~A. Smith, and
  Yejin Choi.
\newblock Superbpe: Space travel for language models, 2025.
\newblock \url{https://arxiv.org/abs/2503.13423}.

\bibitem[Liu and Nocedal(1989)]{liu1989bfgs}
Dong~C. Liu and Jorge Nocedal.
\newblock On the limited memory bfgs method for large scale optimization.
\newblock \emph{Math. Program.}, 45\penalty0 (1–3):\penalty0 503–528,
  August 1989.
\newblock ISSN 0025-5610.

\bibitem[{Llama Team, AI @ Meta}(2024)]{llama3herd2024}
{Llama Team, AI @ Meta}.
\newblock The llama 3 herd of models.
\newblock \emph{arXiv preprint arXiv:2407.21783}, 2024.
\newblock \url{https://arxiv.org/abs/2407.21783}.

\bibitem[Longpre et~al.(2026)Longpre, Kudugunta, Muennighoff, Hsu, Caswell,
  Pentland, Arik, Lee, and Ebrahimi]{longpre2026atlas}
Shayne Longpre, Sneha Kudugunta, Niklas Muennighoff, I-Hung Hsu, Isaac Caswell,
  Alex Pentland, Sercan Arik, Chen-Yu Lee, and Sayna Ebrahimi.
\newblock Atlas: Adaptive transfer scaling laws for multilingual pretraining,
  finetuning, and decoding the curse of multilinguality, 2026.
\newblock \url{https://arxiv.org/abs/2510.22037}.

\bibitem[Loshchilov and Hutter(2019)]{loshchilov2019decoupled}
Ilya Loshchilov and Frank Hutter.
\newblock Decoupled weight decay regularization.
\newblock 2019.
\newblock \url{https://arxiv.org/abs/1711.05101}.

\bibitem[Ludziejewski et~al.(2024)Ludziejewski, Krajewski, Adamczewski,
  Pi\'{o}ro, Krutul, Antoniak, Ciebiera, Kr\'{o}l, Odrzyg\'{o}\'{z}d\'{z},
  Sankowski, Cygan, and Jaszczur]{ludziejewski2024moe}
Jan Ludziejewski, Jakub Krajewski, Kamil Adamczewski, Maciej Pi\'{o}ro,
  Micha{\l} Krutul, Szymon Antoniak, Kamil Ciebiera, Krystian Kr\'{o}l, Tomasz
  Odrzyg\'{o}\'{z}d\'{z}, Piotr Sankowski, Marek Cygan, and Sebastian Jaszczur.
\newblock Scaling laws for fine-grained mixture of experts.
\newblock In Ruslan Salakhutdinov, Zico Kolter, Katherine Heller, Adrian
  Weller, Nuria Oliver, Jonathan Scarlett, and Felix Berkenkamp, editors,
  \emph{Proceedings of the 41st International Conference on Machine Learning},
  volume 235 of \emph{Proceedings of Machine Learning Research}, pages
  33270--33288. PMLR, 21--27 Jul 2024.
\newblock \url{https://proceedings.mlr.press/v235/ludziejewski24a.html}.

\bibitem[Martins et~al.(2025)Martins, Fernandes, Alves, Guerreiro, Rei, Alves,
  Pombal, Farajian, Faysse, Klimaszewski, Colombo, Haddow, de~Souza, Birch, and
  Martins]{martins2024eurollm}
Pedro~Henrique Martins, Patrick Fernandes, Jo\~{a}o Alves, Nuno~M. Guerreiro,
  Ricardo Rei, Duarte~M. Alves, Jos\'{e} Pombal, Amin Farajian, Manuel Faysse,
  Mateusz Klimaszewski, Pierre Colombo, Barry Haddow, Jos\'{e}~G.C. de~Souza,
  Alexandra Birch, and Andr\'{e}~F.T. Martins.
\newblock Eurollm: Multilingual language models for europe.
\newblock \emph{Procedia Comput. Sci.}, 255\penalty0 (C):\penalty0 53–62,
  January 2025.
\newblock ISSN 1877-0509.
\newblock \doi{10.1016/j.procs.2025.02.260}.
\newblock \url{https://doi.org/10.1016/j.procs.2025.02.260}.

\bibitem[Nawrot et~al.(2023)Nawrot, Chorowski, Lancucki, and
  Ponti]{nawrot-etal-2023-efficient}
Piotr Nawrot, Jan Chorowski, Adrian Lancucki, and Edoardo~Maria Ponti.
\newblock Efficient transformers with dynamic token pooling.
\newblock In Anna Rogers, Jordan Boyd-Graber, and Naoaki Okazaki, editors,
  \emph{Proceedings of the 61st Annual Meeting of the Association for
  Computational Linguistics (Volume 1: Long Papers)}, pages 6403--6417,
  Toronto, Canada, July 2023. Association for Computational Linguistics.
\newblock \doi{10.18653/v1/2023.acl-long.353}.
\newblock \url{https://aclanthology.org/2023.acl-long.353/}.

\bibitem[Neitemeier et~al.(2025)Neitemeier, Deiseroth, Eichenberg, and
  Balles]{neitemeier2025hierarchical}
Pit Neitemeier, Bj{\"o}rn Deiseroth, Constantin Eichenberg, and Lukas Balles.
\newblock Hierarchical autoregressive transformers: Combining byte- and
  word-level processing for robust, adaptable language models.
\newblock In \emph{The Thirteenth International Conference on Learning
  Representations}, 2025.
\newblock \url{https://openreview.net/forum?id=tU074jg2vS}.

\bibitem[Owodunni et~al.(2025)Owodunni, Ahia, and
  Kumar]{owodunni2025flexitokens}
Abraham~Toluase Owodunni, Orevaoghene Ahia, and Sachin Kumar.
\newblock Flexitokens: Flexible tokenization for evolving language models.
\newblock \emph{arXiv preprint arXiv:2507.12720}, 2025.

\bibitem[Pagnoni et~al.(2025)Pagnoni, Pasunuru, Rodriguez, Nguyen, Muller, Li,
  Zhou, Yu, Weston, Zettlemoyer, Ghosh, Lewis, Holtzman, and
  Iyer]{pagnoni-etal-2025-byte}
Artidoro Pagnoni, Ramakanth Pasunuru, Pedro Rodriguez, John Nguyen, Benjamin
  Muller, Margaret Li, Chunting Zhou, Lili Yu, Jason~E Weston, Luke
  Zettlemoyer, Gargi Ghosh, Mike Lewis, Ari Holtzman, and Srini Iyer.
\newblock Byte latent transformer: Patches scale better than tokens.
\newblock In \emph{Proceedings of the 63rd Annual Meeting of the Association
  for Computational Linguistics (Volume 1: Long Papers)}, pages 9238--9258,
  Vienna, Austria, July 2025. Association for Computational Linguistics.
\newblock \doi{10.18653/v1/2025.acl-long.453}.
\newblock \url{https://aclanthology.org/2025.acl-long.453/}.

\bibitem[Pearce and Song(2024)]{Pearce2024reconciling}
Tim Pearce and Jinyeop Song.
\newblock Reconciling kaplan and chinchilla scaling laws.
\newblock In \emph{TMLR}, 2024.

\bibitem[Penedo et~al.(2025)Penedo, Kydlíček, Sabolčec, Messmer, Foroutan,
  Kargaran, Raffel, Jaggi, Werra, and Wolf]{penedo2025fineweb}
Guilherme Penedo, Hynek Kydlíček, Vinko Sabolčec, Bettina Messmer, Negar
  Foroutan, Amir~Hossein Kargaran, Colin Raffel, Martin Jaggi, Leandro~Von
  Werra, and Thomas Wolf.
\newblock Fineweb2: One pipeline to scale them all -- adapting pre-training
  data processing to every language, 2025.
\newblock \url{https://arxiv.org/abs/2506.20920}.

\bibitem[Petrov et~al.(2023)Petrov, La~Malfa, H.~S.~Torr, and
  Bibi]{petrov2023token_unfairness}
Aleksandar Petrov, Emanuele La~Malfa, Philip H.~S.~Torr, and Adel Bibi.
\newblock Language model tokenizers introduce unfairness between languages.
\newblock In \emph{Advances in Neural Information Processing Systems}, 2023.
\newblock \url{https://arxiv.org/abs/2305.15425}.

\bibitem[Porian et~al.(2024)Porian, Wortsman, Jitsev, Schmidt, and
  Carmon]{porian2024resolving}
Tomer Porian, Mitchell Wortsman, Jenia Jitsev, Ludwig Schmidt, and Yair Carmon.
\newblock Resolving discrepancies in compute-optimal scaling of language
  models.
\newblock In A.~Globerson, L.~Mackey, D.~Belgrave, A.~Fan, U.~Paquet,
  J.~Tomczak, and C.~Zhang, editors, \emph{Advances in Neural Information
  Processing Systems}, volume~37, pages 100535--100570. Curran Associates,
  Inc., 2024.
\newblock \doi{10.52202/079017-3189}.
\newblock
  \url{https://proceedings.neurips.cc/paper_files/paper/2024/file/b6341525cd84f3be0ef203e4d7cd8556-Paper-Conference.pdf}.

\bibitem[Provilkov et~al.(2020)Provilkov, Emelianenko, and
  Voita]{provilkov-etal-2020-bpe}
Ivan Provilkov, Dmitrii Emelianenko, and Elena Voita.
\newblock {BPE}-dropout: Simple and effective subword regularization.
\newblock In Dan Jurafsky, Joyce Chai, Natalie Schluter, and Joel Tetreault,
  editors, \emph{Proceedings of the 58th Annual Meeting of the Association for
  Computational Linguistics}, pages 1882--1892, Online, July 2020. Association
  for Computational Linguistics.
\newblock \doi{10.18653/v1/2020.acl-main.170}.
\newblock \url{https://aclanthology.org/2020.acl-main.170/}.

\bibitem[Raffel et~al.(2020)Raffel, Shazeer, Roberts, Lee, Narang, Matena,
  Zhou, Li, and Liu]{raffel2020exploring}
Colin Raffel, Noam Shazeer, Adam Roberts, Katherine Lee, Sharan Narang, Michael
  Matena, Yanqi Zhou, Wei Li, and Peter~J. Liu.
\newblock Exploring the limits of transfer learning with a unified text-to-text
  transformer.
\newblock \emph{Journal of Machine Learning Research}, 21\penalty0
  (140):\penalty0 1--67, 2020.
\newblock \url{http://jmlr.org/papers/v21/20-074.html}.

\bibitem[Rust et~al.(2021)Rust, Pfeiffer, Vuli{\'c}, Ruder, and
  Gurevych]{rust-etal-2021-good}
Phillip Rust, Jonas Pfeiffer, Ivan Vuli{\'c}, Sebastian Ruder, and Iryna
  Gurevych.
\newblock How good is your tokenizer? on the monolingual performance of
  multilingual language models.
\newblock In Chengqing Zong, Fei Xia, Wenjie Li, and Roberto Navigli, editors,
  \emph{Proceedings of the 59th Annual Meeting of the Association for
  Computational Linguistics and the 11th International Joint Conference on
  Natural Language Processing (Volume 1: Long Papers)}, pages 3118--3135,
  Online, August 2021. Association for Computational Linguistics.
\newblock \doi{10.18653/v1/2021.acl-long.243}.
\newblock \url{https://aclanthology.org/2021.acl-long.243/}.

\bibitem[Schmidt et~al.(2024)Schmidt, Reddy, Zhang, Alameddine, Uzan, Pinter,
  and Tanner]{schmidt-etal-2024-tokenization}
Craig~W Schmidt, Varshini Reddy, Haoran Zhang, Alec Alameddine, Omri Uzan,
  Yuval Pinter, and Chris Tanner.
\newblock Tokenization is more than compression.
\newblock In Yaser Al-Onaizan, Mohit Bansal, and Yun-Nung Chen, editors,
  \emph{Proceedings of the 2024 Conference on Empirical Methods in Natural
  Language Processing}, pages 678--702, Miami, Florida, USA, November 2024.
  Association for Computational Linguistics.
\newblock \doi{10.18653/v1/2024.emnlp-main.40}.
\newblock \url{https://aclanthology.org/2024.emnlp-main.40/}.

\bibitem[Sennrich et~al.(2016)Sennrich, Haddow, and
  Birch]{sennrich-etal-2015-neural}
Rico Sennrich, Barry Haddow, and Alexandra Birch.
\newblock Neural machine translation of rare words with subword units.
\newblock In \emph{Proceedings of the 54th Annual Meeting of the Association
  for Computational Linguistics (Volume 1: Long Papers)}, pages 1715--1725,
  Berlin, Germany, August 2016. Association for Computational Linguistics.
\newblock \doi{10.18653/v1/P16-1162}.
\newblock \url{https://aclanthology.org/P16-1162/}.

\bibitem[Slagle(2024)]{slagle2024spacebyte}
Kevin Slagle.
\newblock Spacebyte: Towards deleting tokenization from large language
  modeling.
\newblock In \emph{The Thirty-eighth Annual Conference on Neural Information
  Processing Systems}, 2024.
\newblock \url{https://openreview.net/forum?id=KEe4IUp20I}.

\bibitem[Tao et~al.(2024)Tao, Liu, Dou, Muennighoff, Wan, Luo, Lin, and
  Wong]{tao2024scaling}
Chaofan Tao, Qian Liu, Longxu Dou, Niklas Muennighoff, Zhongwei Wan, Ping Luo,
  Min Lin, and Ngai Wong.
\newblock Scaling laws with vocabulary: Larger models deserve larger
  vocabularies.
\newblock In \emph{The Thirty-eighth Annual Conference on Neural Information
  Processing Systems}, 2024.
\newblock \url{https://openreview.net/forum?id=sKCKPr8cRL}.

\bibitem[team et~al.(2022)team, Costa-juss{\`a}, Cross, cCelebi, Elbayad,
  Heafield, Heffernan, Kalbassi, Lam, Licht, Maillard, Sun, Wang, Wenzek,
  Youngblood, Akula, Barrault, Gonzalez, Hansanti, Hoffman, Jarrett, Sadagopan,
  Rowe, Spruit, Tran, Andrews, Ayan, Bhosale, Edunov, Fan, Gao, Goswami,
  Guzm{\'a}n, Koehn, Mourachko, Ropers, Saleem, Schwenk, and Wang]{nllb2022}
Nllb team, Marta~Ruiz Costa-juss{\`a}, James Cross, Onur cCelebi, Maha Elbayad,
  Kenneth Heafield, Kevin Heffernan, Elahe Kalbassi, Janice Lam, Daniel Licht,
  Jean Maillard, Anna Sun, Skyler Wang, Guillaume Wenzek, Alison Youngblood,
  Bapi Akula, Lo{\"i}c Barrault, Gabriel~Mejia Gonzalez, Prangthip Hansanti,
  John Hoffman, Semarley Jarrett, Kaushik~Ram Sadagopan, Dirk Rowe, Shannon~L.
  Spruit, C.~Tran, Pierre~Yves Andrews, Necip~Fazil Ayan, Shruti Bhosale,
  Sergey Edunov, Angela Fan, Cynthia Gao, Vedanuj Goswami, Francisco~(Paco)
  Guzm{\'a}n, Philipp Koehn, Alexandre Mourachko, Christophe Ropers, Safiyyah
  Saleem, Holger Schwenk, and Jeff Wang.
\newblock No language left behind: Scaling human-centered machine translation.
\newblock \emph{ArXiv}, abs/2207.04672, 2022.
\newblock \url{https://arxiv.org/abs/2207.04672}.

\bibitem[Team(2025)]{qwen3technicalreport}
Qwen Team.
\newblock Qwen3 technical report, 2025.
\newblock \url{https://arxiv.org/abs/2505.09388}.

\bibitem[van~den Oord et~al.(2017)van~den Oord, Vinyals, and
  Kavukcuoglu]{oord2017vqvae}
Aaron van~den Oord, Oriol Vinyals, and Koray Kavukcuoglu.
\newblock Neural discrete representation learning.
\newblock In \emph{Proceedings of the 31st International Conference on Neural
  Information Processing Systems}, NIPS'17, page 6309–6318, Red Hook, NY,
  USA, 2017. Curran Associates Inc.
\newblock ISBN 9781510860964.

\bibitem[Vaswani et~al.(2017)Vaswani, Shazeer, Parmar, Uszkoreit, Jones, Gomez,
  Kaiser, and Polosukhin]{vaswani2017attention}
Ashish Vaswani, Noam Shazeer, Niki Parmar, Jakob Uszkoreit, Llion Jones,
  Aidan~N. Gomez, Lukasz Kaiser, and Illia Polosukhin.
\newblock Attention is all you need.
\newblock In \emph{Advances in Neural Information Processing Systems},
  volume~30. Curran Associates, Inc., 2017.
\newblock \url{https://papers.nips.cc/paper/7181-attention-is-all-you-need}.

\bibitem[Videau et~al.(2025)Videau, Idrissi, Leite, Schoenauer, Teytaud, and
  Lopez-Paz]{videau2025bytesideaslanguagemodeling}
Mathurin Videau, Badr~Youbi Idrissi, Alessandro Leite, Marc Schoenauer, Olivier
  Teytaud, and David Lopez-Paz.
\newblock From bytes to ideas: Language modeling with autoregressive u-nets.
\newblock \emph{arXiv preprint arXiv:2506.14761}, 2025.

\bibitem[Wang et~al.(2024)Wang, Gangavarapu, Yan, and
  Rush]{wang2024mambabytetokenfreeselectivestate}
Junxiong Wang, Tushaar Gangavarapu, Jing~Nathan Yan, and Alexander~M. Rush.
\newblock Mambabyte: Token-free selective state space model, 2024.
\newblock \url{https://arxiv.org/abs/2401.13660}.

\bibitem[Xue et~al.(2022)Xue, Barua, Constant, Al-Rfou, Narang, Kale, Roberts,
  and Raffel]{xue-etal-2022-byt5}
Linting Xue, Aditya Barua, Noah Constant, Rami Al-Rfou, Sharan Narang, Mihir
  Kale, Adam Roberts, and Colin Raffel.
\newblock {B}y{T}5: Towards a token-free future with pre-trained byte-to-byte
  models.
\newblock \emph{Transactions of the Association for Computational Linguistics},
  10:\penalty0 291--306, 2022.
\newblock \doi{10.1162/tacl_a_00461}.
\newblock \url{https://aclanthology.org/2022.tacl-1.17/}.

\bibitem[Yang et~al.(2025)Yang, Guo, Jing, Zhang, Liu, Hao, Li, Zhao, Liu, Lv,
  and Dai]{yang2025scalinglawscode}
Jian Yang, Shawn Guo, Lin Jing, Wei Zhang, Aishan Liu, Chuan Hao, Zhoujun Li,
  Wayne~Xin Zhao, Xianglong Liu, Weifeng Lv, and Bryan Dai.
\newblock Scaling laws for code: Every programming language matters, 2025.
\newblock \url{https://arxiv.org/abs/2512.13472}.

\bibitem[Yu et~al.(2024)Yu, Weber, Deng, Shen, Cremers, and Chen]{yu2024image}
Qihang Yu, Mark Weber, Xueqing Deng, Xiaohui Shen, Daniel Cremers, and
  Liang-Chieh Chen.
\newblock An image is worth 32 tokens for reconstruction and generation.
\newblock In \emph{The Thirty-eighth Annual Conference on Neural Information
  Processing Systems}, 2024.
\newblock \url{https://openreview.net/forum?id=tOXoQPRzPL}.

\bibitem[Zellers et~al.(2019)Zellers, Holtzman, Bisk, Farhadi, and
  Choi]{zellers2019hellaswag}
Rowan Zellers, Ari Holtzman, Yonatan Bisk, Ali Farhadi, and Yejin Choi.
\newblock Hellaswag: Can a machine really finish your sentence?
\newblock In \emph{Proceedings of the 57th Annual Meeting of the Association
  for Computational Linguistics}, 2019.

\bibitem[Zhu et~al.(1997)Zhu, Byrd, Lu, and Nocedal]{zhu1997bfgsb}
Ciyou Zhu, Richard~H Byrd, Peihuang Lu, and Jorge Nocedal.
\newblock Algorithm 778: L-bfgs-b: Fortran subroutines for large-scale
  bound-constrained optimization.
\newblock \emph{ACM Transactions on Mathematical Software (TOMS)}, 23\penalty0
  (4):\penalty0 550--560, 1997.

\end{thebibliography}
